\documentclass[msom,nonblindrev]{informs3noheader}

\usepackage{algorithm}
\usepackage{algpseudocode}
\usepackage{tikz}
\usepackage{amsmath,amssymb}
\usepackage{bbm}
\usepackage{endnotes}
\usepackage{mathtools}
\usepackage{xcolor}
\usepackage{comment}
\usepackage{caption}
\usepackage{subcaption}
\usepackage{hyperref}
\usepackage{booktabs}

\usepackage{natbib}
\bibpunct[, ]{(}{)}{,}{a}{}{,}%
\def\bibfont{\small}%

\EquationsNumberedThrough    

\TheoremsNumberedThrough

\ECRepeatTheorems

\begin{document}

\RUNAUTHOR{Bastani et al.}

\RUNTITLE{Inference-Aware Policy Optimization}

\TITLE{Beating the Winner's Curse via \\Inference-Aware Policy Optimization}

\ARTICLEAUTHORS{
\AUTHOR{Hamsa Bastani}
\AFF{Operations, Information, and Decisions Department,
The Wharton School, \EMAIL{hamsab@wharton.upenn.edu}}

\AUTHOR{Osbert Bastani}
\AFF{Department of Computer and Information Science,
University of Pennsylvania, \EMAIL{obastani@seas.upenn.edu}}

\AUTHOR{Bryce McLaughlin}
\AFF{Wharton Healthcare Analytics Lab,
Wharton AI \& Analytics Initiative, \EMAIL{brycemcl@wharton.upenn.edu}}

} 

\ABSTRACT{
There has been a surge of recent interest in automatically learning policies to target treatment decisions based on rich individual covariates. In addition, practitioners want confidence that the learned policy has better performance than the incumbent policy according to downstream policy evaluation. However, due to the \emph{winner's curse}---an issue where the policy optimization procedure exploits prediction errors rather than finding actual improvements---predicted performance improvements are often not substantiated by downstream policy evaluation. To address this challenge, we propose a novel strategy called \emph{inference-aware policy optimization}, which modifies policy optimization to account for how the policy will be evaluated downstream. Specifically, it optimizes not only for the estimated objective value, but also for the chances that the estimate of the policy's improvement passes a significance test during downstream policy evaluation. We mathematically characterize the Pareto frontier of policies according to the tradeoff of these two goals. Based on our characterization, we design a policy optimization algorithm that estimates the Pareto frontier using machine learning models; then, the decision-maker can select the policy that optimizes their desired tradeoff, after which policy evaluation can be performed on the test set as usual. Finally, we perform simulations to illustrate the effectiveness of our methodology.
\footnote{We gratefully acknowledge Katherine Milkman, Dena Gromet, and the Behavior Change for Good Initiative for providing access to the megastudy data used in Section~\ref{sec:vaccine} of our paper. Furthermore, we would like to thank Ron Berman, Dean Eckles, Vishal Gupta, Dean Knox, Daniel Rock, Neha Sharma,  Hummy Song, Alp Sungu, Sophie Yu, and members of the Wharton AI \& Analytics Initiative for their helpful discussion and feedback.
}%
}

\maketitle

\section{Introduction}

Due to rapid improvements in the capabilities of machine learning over the past two decades, there has been a surge of recent interest in systems that automatically make decisions based on the predictions of machine learning models~\citep{dudik2011doubly,bottou2013counterfactual}. These methods have been deployed across a wide variety of application domains, including refugee matching programs~\citep{bansak_improving_2018,ahani_placement_2021} and personalized healthcare treatment decisions~\citep{boutilier_improving_2022, lekwijit2024evaluating}.

However, existing methodologies for policy learning face a major challenge known as the \emph{winner's curse}. This issue arises when training a model to approximate the true objective and then optimizing against this model. Intuitively, even if the model is highly accurate on the training data, the optimization procedure can systematically exploit its prediction errors to obtain a policy whose predicted value is substantially higher than its true value. This failure has previously been observed across a variety of contexts, including common value auctions~\citep{capen1971competitive,thaler1988anomalies} and managerial decision-making (where it has also been called \emph{post-decision surprise}~\citep{harrison1984decision} or the \emph{optimizer's curse}~\citep{smith2006optimizer}); it is also an instance of regression to the mean~\citep{harrison1984decision}. It is greatly exacerbated by modern policy optimization pipelines, which consider complex policies mapping rich, individual covariates to actions.

Much of the early work on policy evaluation assumes that a policy is fixed prior to evaluation~\citep{rosenbaum_central_1983, rosenbaum1984reducing,bang2005doubly}. Due to increasing interest in policy optimization, recent work has proposed methodologies for correcting the optimistic bias arising from the winner's curse~\citep{gupta2024debiasing,andrews_inference_2024,zrnic_flexible_2024,xu2025winner}. However, they only correct for the evaluation error---if the learned policy is ineffective, they cannot help compute a more effective policy. Furthermore, practitioners often want evidence that the learned policy actually increases performance---e.g., by obtaining a statistically significant estimate of the learned policy's performance improvement over the incumbent policy (which we call the \emph{observational policy}). Thus, a critical remaining question is how to identify a policy that improves performance according to a rigorous downstream evaluation methodology. Addressing this challenge requires rethinking the entire policy optimization pipeline---our key insight is that policy optimization should account for how the policy will be evaluated downstream, which we call \emph{inference-aware policy optimization}.

We formalize the inference-aware policy optimization problem and mathematically characterize the space of solutions to this problem. We focus on a simple policy evaluation methodology using inverse propensity weighting (IPW), which evaluates a learned policy by reweighting test outcomes to appear as if they were collected under the learned policy~\citep{rosenbaum_central_1983}; we restrict the policy optimization algorithm to only see training data and test covariates, which suffices to avoid the winner's curse during evaluation. Our goal is to design a policy optimization algorithm that aims to achieve statistically significant performance improvement. Specifically, we aim to compute policies that trade off (1) the expected objective value according to the downstream IPW estimate, and (2) the expected $z$-score of the IPW estimate used to assess whether the learned policy is statistically significantly better than the observational policy. Intuitively, if the learned and observational policies are quite different, then the IPW estimator will downweight a large part of the test dataset when evaluating the proposed policy~\citep{li_unbiased_2011}, thereby effectively reducing sample size and failing to establish a statistically significant performance improvement. Thus, to be conservative, we want to keep the learned policy closer to the random policy used to collect data, avoiding large propensity weights that can result in high variance.

In this setting, we obtain a precise mathematical characterization of the Pareto frontier of optimal solutions---namely, the tradeoff between the expected objective value and $z$-score. Specifically, we create a superset of this frontier by considering a family of convex optimization problems that minimize the variance of the IPW estimator for a fixed expected improvement $\lambda$. By analyzing the Lagrangian of this problem, we obtain a parametric representation of this family which we prune to obtain the Pareto frontier of optimal policies.

Using this characterization, we show that the policy with the largest expected $z$-score is close to the observational policy, making a small but confident improvement. Specifically, it differs from the observational policy for all subjects where an expected improvement can be made, with larger differences when the counterfactual means of treatments differ more and their related variances are small. The step size reaches the point where some treatment option is eliminated for some unit.
Moving along the frontier from this initial solution, the tradeoff between expected objective and statistical significance becomes strict---i.e., any strict increase in expected objective requires a strict decrease in statistical significance. Furthermore, we provide a simple closed form for the largest expected $z$-score, effectively allowing for power calculations that determine whether a test dataset has enough useful variation to evaluate any policy's improvement as significant.

Next, we show how our theoretical results can be used to derive our algorithm for inference-aware policy optimization (IAPO). The main barrier is that the solution to our convex program is specialized to assigning treatments to our test units instead of being a generalizable policy that can be applied to any unit. However, we note that the policy assigns treatments based purely on the mean and standard deviation of the predicted counterfactual outcomes; thus, we can straightforwardly devise a generalizable policy from our solution. A full policy optimization and evaluation pipeline consists of the following steps: (1) train a predictive model on the training dataset, (2) compute the Pareto frontier using our characterization and select a policy along this frontier, and (3) evaluate the efficacy of this policy using IPW. We implement this pipeline in multiple simulation studies, demonstrating that it can effectively extract policies that deliver confident improvements in cases where na\"{i}ve policy optimization approaches fail.

\subsection{Related Work}

\paragraph{Policy evaluation.}

Our approach builds on the causal model introduced by \cite{rubin_estimating_1974}. Specifically, the key challenge with policy evaluation is the need to estimate counterfactual outcomes---i.e., for each treatment unit (e.g., an individual), we only observe the outcome for the treatment that is assigned to that unit by the observational policy (called the \emph{factual outcome}), not the outcomes for other treatments (called \emph{counterfactual outcomes}). Then, the policy evaluation problem is to estimate the average treatment effect (ATE) of one treatment compared to another (usually the control). This problem does not directly consider targeted interventions, where the treatment is targeted based on unit covariates instead of being constant across all units. However, when a specific candidate targeting policy is given, then the problem of policy evaluation can be reduced to this one by considering the policy to be the intervention, and evaluating the efficacy of that policy.

We focus on the setting where there are no unobserved confounders; for instance, this is the case in applications where data is collected from a known observational policy such as random assignment. In this setting, there are two strategies that can be applied for policy evaluation. The first (called the \emph{direct method} in \cite{dudik2011doubly}) is to train a predictive model to estimate counterfactual outcomes, and use it to evaluate the candidate policy. This approach assumes that the predictive model is accurate, which is typically not true in practice. An alternative strategy is inverse propensity weighting~\citep{rosenbaum_central_1983}, which uses importance weights to correct for the shift from the observational policy to the candidate policy. This approach requires that the inverse propensity weights are known (e.g., this is the case when the observational policy is random assignment); otherwise, they must be estimated. One approach to further improve on importance weighting is doubly robust estimation~\citep{robins1994estimation}, which subtracts a prediction of the treatment outcome from the realized outcome to normalize it, resulting in the augmented IPW (AIPW) estimator. When the predicted outcomes are accurate, this strategy can significantly reduce variance compared to the IPW estimator. Doubly robust estimation is often used in settings where weights must be estimated, but it can reduce variance even when they are known. While we do not explicitly include the doubly robust correction in our main derivation, it can be straightforwardly incorporated into our framework.

\paragraph{Policy learning.}

The problem of learning an optimal treatment policy has long been studied in the multi-armed bandit~\citep{lai1985asymptotically} and reinforcement learning~\citep{sutton1998reinforcement,murphy2003optimal} literatures. While a majority of work in this space focuses on learning in the sequential setting, which seeks to learn through data collected online from the current policy, there has been recent interest in \emph{offline learning}~\citep{dudik2011doubly,bottou2013counterfactual} (also called \emph{batch learning}~\citep{ernst2005tree}), where the goal is to learn from historical data (the literature also distinguishes \emph{off-policy learning}, which seeks to learn from a combination of historical and online data~\citep{watkins1992q,precup2001off}; see \cite{levine2020offline} for a discussion).

The standard approach to policy learning is to construct an objective based on a policy evaluation technique, and then select the policy that optimizes this objective. Using the direct method for policy evaluation results in a \emph{plug-in} approach~\citep{qian_performance_2011}, whereas using IPW results in a \emph{weight-based} approach~\citep{bottou2013counterfactual,kitagawa_who_2018}; doubly robust approaches combining the two can also be used~\citep{dudik2011doubly,zhang2012robust,zhou_residual_2017}. There has also been work on using self-normalization to improve out-of-sample performance \citep{NIPS2015_39027dfa,kuzborskij_confident_2021}, improving robustness to confounders \citep{Kallus_Confounding}, and handling multiple observational policies~\citep{zhan2024policy}.

However, these strategies all suffer from the winner's curse---even if policy evaluation is unbiased for a single candidate policy, the objective value of the learned policy is optimistically biased. These approaches avoid this issue by using sample splitting---i.e., optimize the objective on an optimization set, and then perform policy evaluation (e.g., using IPW) on a held-out test set (the policy optimization procedure can use the test covariates; only treatment assignments and outcomes need to be held out). More sophisticated approaches have recently been proposed to avoid the reduction in sample complexity from sample splitting~\citep{gupta2024debiasing,andrews_inference_2024,zrnic_flexible_2024,xu2025winner}. However, all of these approaches can only correct for optimistic bias in policy evaluation; they cannot select a policy that is more likely to achieve statistically significant policy improvement, and often fail to do so.

There are two existing strategies that can be viewed as performing policy optimization to improve statistical significance. The first is treatment pooling, which tries to pool treatment options to reduce the size of the policy class in consideration \citep{banerjee_selecting_2025,Ma_adaptive_fusion,ladhania_personalized_2023,saito_potec:_2024}; a related approach is identifying promising subgroups to reduce the size of the policy space~\citep{spiess_finding_2023}. This procedure can be viewed as reducing variance in exchange for a small amount of bias; however, it does so in an ad hoc way and can only be applied to small policy classes.

The second is regularization, where the policy optimization objective includes some kind of term to penalize variance in the policy evaluation objective. For instance, \cite{kallus_balanced_2018} estimates policy performance using weights that minimize the conditional mean squared error of the policy's performance estimates. Similarly, the Policy Optimizer for Exponential Model (POEM)~\citep{swaminathan_batch_2015} and its self-normalized variant (Norm-POEM)~\citep{NIPS2015_39027dfa} penalize the variance of importance weighted samples. There has also been work along these lines in reinforcement learning, including conservative $Q$-learning~\citep{kumar2020conservative} and KL-regularized policy optimization~\citep{nachum2019algaedice}. However, these approaches do not connect variance regularization to statistical significance; furthermore, their optimization objectives are non-convex and are not amenable to theoretical analysis. Finally, in concurrent work, \cite{chernozhukov_policy_2025} connects variance-regularized policy optimization to statistical inference. However, they do not provide any insights into the structure of the Pareto frontier; instead, they provide valid inference after selecting a policy along the frontier. In contrast, we obtain a closed-form solution to the Pareto frontier, and use it to obtain interesting insights on its structure.

\subsection{Contributions}

Our contributions are three-fold. (1) We formalize the inference-aware policy optimization problem, where the goal is to obtain a policy with statistically significant policy improvement. (2) We perform a theoretical analysis on the tradeoff between expected performance and statistically significant policy improvement---specifically, we compute the entire Pareto frontier of policies trading off expected improvement and statistical significance of improvement. We find a number of interesting insights (summarized below). (3) We propose a novel policy optimization algorithm based on this analysis; our policy optimization algorithm is the first one designed to enable practitioners to obtain statistically significant policy improvements. By selecting a desired policy along the Pareto frontier, practitioners can obtain the best policy at a desired level of confidence.

Now, we summarize some of the key insights from our theoretical analysis:
\begin{itemize}
\item \textbf{Structure of the Pareto frontier:} The Pareto frontier ranges between two policies, one maximizing statistical significance (specifically, the $z$-score) and one maximizing expected outcomes. Between the two, there is a strict tradeoff between $z$-score and expected outcomes.
\item \textbf{Policies on the frontier:} Notably, policies along the frontier are stochastic. Rather than assign units deterministically to treatments (as is the case with many existing methodologies), they assign treatment probabilities. These probabilities start from that of the observational policy, and move away until they bind to zero or one (this movement may be non-monotonic when there are multiple treatments). Intuitively, if a unit has high variance outcomes, then keeping its treatment probability close to that of the observational policy removes variance by effectively dropping it from the sample. This finding shows that policy learning algorithms must learn stochastic policies to improve statistical power.
\item \textbf{Statistical power:} The statistical power of the optimal $z$-score monotonically increases with the number of available units. However, the behavior of the $z$-score for the expectation-maximizing policy is very different. There can be a substantial gap between this $z$-score and the optimal one. Worse, it can scale to zero as the number of units increases, meaning increasing sample size can actually worsen significance. This finding requires somewhat pathological conditions, but even under reasonable conditions, it can scale non-monotonically with the number of units.
\end{itemize}

\section{General Framework}

Our goal is to characterize the Pareto frontier of policies that jointly maximize the expected value and expected significance of an inverse propensity weighting (IPW) estimate of the policy improvement over the observational policy used to collect an observational dataset.
We begin by formalizing this problem (Section~\ref{sec:problem}), and establishing the convex optimization framework we use to solve it (Section~\ref{sec:convex}). In Section~\ref{sec:binary}, we examine the special case of a single treatment (i.e., treatment vs. control group), enabling us to derive closed-form expressions for the Pareto frontier and obtain qualitative insights. Then, in Section~\ref{sec:multi}, we analyze the general case of multiple treatments; while we can no longer obtain a closed-form representation of the frontier, we nevertheless show that our key insights about its structure continue to hold.

\subsection{Problem Formulation}
\label{sec:problem}

Policy evaluation combines the Rubin causal model with inverse propensity weighting (IPW) to provide an unbiased estimate of an (unobserved) policy's performance using data collected by an observational policy. Under the Rubin causal model, we cannot observe the counterfactual outcomes units would realize under treatments other than the one they were assigned by the observational policy, inhibiting our ability to estimate the performance of other policies. IPW forms an unbiased estimate of any policies performance by reweighting the outcomes of the observational policy so they appear as if generated by a different policy. As these estimates can be quite noisy, confidently believing any policy delivers an improvement over the observational policy requires passing a significance test. Thus, when proposing policies, practitioners must find high performing policies which they can show are significantly better than the observational policy.  

Following the Rubin causal model~\citep{rubin_estimating_1974}, we consider a set of units $n \in [N]$ that each receive a treatment $t \in \{0,\ldots,K\}$, yielding a random real-valued outcome $Y_{n,t}$. We assume that each of these outcomes are generated independently upon treatment assignment without interference between units. We let $\mu_{n,t} = \mathbb{E}\left[Y_{n,t}\right]$ denote the mean and $\sigma^2_{n,t} = \mathrm{Var}(Y_{n,t})$ denote the variance of these counterfactual outcomes. Each unit is assigned a treatment $T^o_n$ according to a known observational policy $\pi^o_n$ generating an observed outcome $Y_n^o = Y_{n,T_n^o}$. Each observational policy $\pi^o_n\in\Delta^K$ is an element of the standard $K$-simplex, and treatment assignments are independent random variables $T^o_n \sim \mathrm{Categorical}(\pi_{n}^o)$. We assume each treatment is assigned to each unit with positive weight (i.e. positivity). We refer to the collection $\mathcal{O} = \{(\pi^o_1,T^o_1,Y^o_1),\ldots,(\pi^o_N,T^o_N,Y^o_N)\}$ as the observational dataset and let $\pi^o_{n,t}$ be the probability unit $n$ is assigned treatment $t$.

We can use $\mathcal{O}$ to compare the performance of an alternate policy $\pi\in (\Delta^K)^N$ to $\pi^o$ using inverse propensity weighting (IPW)~\citep{rosenbaum_central_1983}. Specifically, the difference in expected performance between $\pi$ and $\pi^o$ is given by the sample average treatment effect
$$\tau(\pi) = \frac{1}{N}\sum_{n=1}^N \sum_{t=0}^K \mu_{n,t}(\pi_{n,t} - \pi_{n,t}^o).$$
Applying IPW to $\mathcal{O}$ yields an unbiased estimate of $\tau(\pi)$:
\begin{align*}
\hat{\tau}(\pi) &= \frac{1}{N}\sum_{n=1}^N Y_n^o(W_n - 1), & &\text{where} &  W_n& = \frac{\pi_{n,T_n^o}}{\pi^o_{n,T_n^o}}
\end{align*}
is the inverse propensity weight. As $\hat{\tau}(\pi)$ is only an estimate of the improvement of $\pi$ over $\pi^o$, a $z$-test needs to be applied to gain confidence that $\tau(\pi)>0$; we use a $z$-test as we assume knowledge of the variances $\sigma^2$ (otherwise a $t$-test would be used), and we expect to use large sample sizes where the central limit theorem guarantees $\hat{\tau}(\pi)$ is normally distributed. The $z$-score for $\hat{\tau}(\pi)$ is
\begin{align*}
Z(\pi)&= \frac{\hat{\tau}(\pi)}{s(\pi)}, & &\text{where} & s(\pi)&=\sqrt{\mathrm{Var}(\hat{\tau}(\pi))}
\end{align*}
is the standard error of $\hat{\tau}(\pi)$.\footnote{Typically, the standard error is unknown, so we instead use a sample approximation $s(\pi)\approx\hat{s}(\pi)$ in conjunction with the $t$-score. For simplicity, we assume the standard error is known, so we can use the $z$-score.}
The policy is accepted if the $z$-score exceeds some level $Z_{\text{min}}$.

Our goal is to construct a policy which improves upon the observational policy as much as possible while passing a significance test on the observational dataset. To do so, we have access to the counterfactual outcome distributions of $Y_{n,t}$ and the observational policy $\pi^o$, but not the observed treatments or outcomes. Note that for all $\pi$, $Z(\pi)$ should be approximately normal with $\mathrm{Var}(Z(\pi)) = 1$, meaning that to maximize our probability of achieving statistical significance, it suffices to maximize the expected $z$-score. Thus, we formalize our goal as computing the Pareto frontier $\Pi_{\mathrm{Opt}}$ for a multi-objective optimization problem, where the objectives are (1) the expected policy improvement $\tau(\pi)$, and (2) the expected $z$-score $\mathbb{E}[Z(\pi)] = \frac{\tau(\pi)}{s(\pi)}$ for the IPW estimator used in conjunction with the observational dataset $\mathcal{O}$. Mathematically, our goal is to compute
\begin{align*}
\Pi_{\text{Opt}} = \bigg\{ \pi \in (\Delta^K)^N\biggm\vert &\not\exists \ \pi'\ \in (\Delta^K)^N  \\ &\text{ s.t. } \tau(\pi') \geq\tau(\pi)\wedge\frac{\tau(\pi')}{s(\pi')} \geq \frac{\tau(\pi)}{s(\pi)}\wedge(\tau(\pi'),s(\pi')) \neq (\tau(\pi),s(\pi))\bigg\},
\end{align*}

i.e., the Pareto frontier $\Pi_{\text{Opt}}$ consists of exactly the policies $\pi$ that cannot be strictly improved upon in terms of both expected improvement and $z$-score.

\subsection{Convex Relaxation Framework}
\label{sec:convex}

This frontier is difficult to characterize directly as one of the objectives (the expected $z$-score) is not convex. We start by showing that a superset of $\Pi_{\mathrm{Opt}}$ is given by a family of convex optimization problems $\Pi_{\Lambda}$. Then, we decompose the first order condition for this family of problems in order to characterize $\Pi_{\Lambda}$, which allows us to characterize $\Pi_{\mathrm{Opt}}$.  Proofs for all claims are given in Appendix~\ref{apx:proofs}.
\begin{repeatproposition}
\label{prop:relax}
We have $\Pi_{\mathrm{Opt}} \subseteq \Pi_{\Lambda}$, where $\tau_{\text{max}} = \max_{\pi} \tau(\pi)$ and
\begin{align*}
\Pi_{\Lambda} = \bigcup_{\lambda \in [0,\tau_{\text{max}}]} \operatorname*{\arg\min}_{\{\pi \in (\Delta^K)^N|\tau(\pi) = \lambda\}} s^2(\pi).
\end{align*}
\end{repeatproposition}
Intuitively, any policy on the frontier $\Pi_{\text{Opt}}$ also minimizes the standard error $s(\pi)$ for some expected improvement $\tau(\pi)$; if this were not this case, then there would exist another policy with the same expected improvement but a higher expected $z$-score, removing it from $\Pi_{\text{Opt}}$. As the standard error is positive, this also holds for minimizing $s^2(\pi)$. We give complete proofs in Appendix~\ref{apx:proofs}.

Proposition~\ref{prop:relax} offers a path to solving for $\Pi_{\text{Opt}}$ by first solving for $\Pi_\Lambda$ then removing extraneous solutions to form the Pareto frontier. This is particularly useful as each optimization problem in $\Pi_{\Lambda}$ is convex in terms of the policy $\pi$.
\begin{repeatproposition}
\label{prop:convex}
$s^2(\pi)$ is convex and $\tau(\pi)$ is affine in $\pi$. Moreover,
\begin{align*}
s^2(\pi) &= \frac{1}{N^2} \sum_{n=1}^N\sum_{t=0}^K\left[ \frac{\sigma^2_{n,t}\left(\pi_{n,t} - \pi^o_{n,t}\right)^2}{\pi^o_{n,t}} + \pi^o_{n,t}\left(\frac{\mu_{n,t}\left(\pi_{n,t}-\pi^o_{n,t}\right)}{\pi^o_{n,t}} - \sum_{v=0}^K \mu_{n,v}(\pi_{n,v} - \pi_{n,v}^o)\right)^2\right]
\end{align*}
\end{repeatproposition}
To characterize $\Pi_\Lambda$ we need to solve this convex problem for all values of $\lambda \in [0,\tau_{\text{max}}]$. The first order condition of this optimization problem does not contain $\lambda$, instead giving a form for $\pi_{n,t}$ in terms of dual variables whose optimal values are induced by $\lambda$. The Lagrangian associated with all problems forming $\Pi_{\Lambda}$ is,
$$\mathcal{L}(\pi,\zeta,\beta,\kappa) = s^2(\pi) + \zeta\left(\lambda - \tau(\pi)\right) +\sum_{n=1}^N\beta_n \left(1-\sum_{t=0}^K \pi_{n,t}\right) - \sum_{n=1}^N\sum_{t=0}^K\kappa_{n,t}\pi_{n,t},$$ where $\kappa > 0$ is the only dual constraint. To write its first order condition to respect $\pi_{n,t}$ we first need the partial derivatives of $s^2(\pi)$ and $\tau(\pi)$,
\begin{repeatproposition}
\label{prop:FOCs}
We have
\begin{align*}
\frac{\partial s^2(\pi)}{\partial \pi_{n,t}} &= \frac{2}{N^2} \left[\frac{\mu_{n,t}^2 + \sigma_{n,t}^2}{\pi^o_{n,t}} \left(\pi_{n,t} - \pi^o_{n,t}\right) - \mu_{n,t} \sum_{v=0}^K \mu_{n,v} \left(\pi_{n,v} - \pi^o_{n,v}\right)\right] \\
\frac{\partial \tau(\pi)}{\partial \pi_{n,t}} &= \frac{1}{N} \mu_{n,t}.
\end{align*}
\end{repeatproposition}
Thus, the first order condition with respect to $\pi_{n,t}^*$ for an optimal solution $(\pi^*,\zeta^*,\beta^*,\kappa^*)$ is
$$ \frac{2}{N^2} \left[\frac{\mu_{n,t}^2 + \sigma_{n,t}^2}{\pi^o_{n,t}} \left(\pi_{n,t}^* - \pi^o_{n,t}\right) - \mu_{n,t} \sum_{v=0}^K \mu_{n,v} \left(\pi_{n,v}^* - \pi^o_{n,v}\right)\right] + \frac{\zeta^*}{N}\mu_{n,t} - \beta_{n}^* - \kappa_{n,t}^* = 0.$$
We can eliminate $\beta^*_n$ using the primal constraint $\sum_{t=0}^K \pi_{n,t}^* = 1$ and then reorganize this system of equations to solve for $\pi^*$ in terms of $\zeta^*$ and $\kappa^*$,
\begin{repeatproposition}
\label{prop:form}
We have
\begin{align*}
\pi^*_{n,t} &=\pi^o_{n,t} \left(1+\frac{Q_n^{-1}\left(\zeta^* + N\sum_{v=0}^K \frac{\pi^o_{n,v}}{\mu_{n,v}^2 + \sigma_{n,v}^2} \kappa^*_{n,v} \left(\mu_{n,v} - \tilde{\mu}_n\right)\right) \left(\mu_{n,t} - \tilde{\mu}_n\right) + N\left(\kappa^*_{n,t} - \tilde{\kappa}^*_n \right)}{2\left(\mu_{n,t}^2 + \sigma_{n,t}^2\right) }N\right)
\end{align*}
where
\begin{align*}
\tilde{\mu}_n &= \frac{\sum_{v=0}^K \frac{\pi^o_{n,v}}{\mu^2_{n,v} + \sigma^2_{n,v}} \mu_{n,v}}{\sum_{v=0}^K \frac{\pi^o_{n,v}}{\mu^2_{n,v} + \sigma^2_{n,v}}}
&
\tilde{\kappa}^*_n&= \frac{\sum_{v=0}^K \frac{\pi^o_{n,v}}{\mu^2_{n,v} + \sigma^2_{n,v}} \kappa^*_{n,v}}{\sum_{v=0}^K \frac{\pi^o_{n,v}}{\mu^2_{n,v} + \sigma^2_{n,v}}}
\end{align*}
and
\begin{align*}
Q_n & = 1 - \sum_{v=0}^K \frac{\pi_{n,v}^o}{\mu_{n,v}^2 + \sigma_{n,v}^2} \mu_{n,v} \left(\mu_{n,v} - \tilde{\mu}_n\right) > 0.
\end{align*}
\end{repeatproposition}
To convert Proposition~\ref{prop:form} into a characterization of $\Pi_{\Lambda}$, we solve for the values of all dual variables except $\zeta^*$, leaving a one-dimensional parameterization of $\Pi_{\Lambda}$ in terms of $\zeta^*$.

\section{Single Treatment Setting}
\label{sec:binary}

For intuition, we first consider for the case of comparing a single treatment to a control (i.e., $K=1$). We begin by solving our convex relaxation to obtain a superset of the Pareto frontier (Section~\ref{sec:binarysuperset}). Then, we characterize the structure of the Pareto frontier based on this result (Section~\ref{sec:binarystructure}). Finally, we discuss implications of our characterization for policy learning methodologies (Section~\ref{sec:binarydiscussion}). Section~\ref{sec:multi} generalizes our characterization results to the multi-treatment setting.

\subsection{Superset of the Pareto Frontier}
\label{sec:binarysuperset}

We begin by characterizing the Pareto frontier $\Pi_{\Lambda}$. The first order condition given in Proposition~\ref{prop:form} is independent across units, thus, for any $\zeta^*$ we can calculate $\pi_n^*(\zeta^*)$ in terms of $\pi_{n}^o$, $\mu_n$, and $\sigma^2_n$. This provides insight into how the dual variable determines optimal solution propensities in terms of the relative quality of the treatment for each unit. We characterize the policies $\pi \in \Pi_{\Lambda}$ in terms of $\pi_{n,1}$, the propensity to give the treatment to unit $n$, noting that $\pi_{n,0} = 1-\pi_{n,1}$. For convenience, define $\tau(\zeta)=\tau(\pi^*(\zeta))$, $s^2(\zeta)=s^2(\pi^*(\zeta))$, and $Z(\zeta)=\mathbb{E}[Z(\pi^*(\zeta))]=\tau(\zeta)/s(\zeta)$. A proof for Theorem~\ref{thm:binary} is given in Appendix~\ref{apx:proofs}.
\begin{repeattheorem}
\label{thm:binary}
For the single-treatment setting (i.e., $K=1$), we have $\Pi_{\Lambda}=\bigcup_{\zeta\ge0}\{\pi^*(\zeta)\}$. Also,
\begin{align}
\label{eqn:binary}
\pi^*_{n,t}(\zeta)=\pi_{n,t}^o+\psi_t\alpha_n
\qquad\qquad
\tau(\zeta)=\frac{1}{N}\sum_{n=1}^N\tau_n\alpha_n
\qquad\qquad
s^2(\zeta)=\frac{1}{N^2}\sum_{n=1}^N\eta_n\alpha_n^2,
\end{align}
where
\begin{align*}
\psi_t=\begin{cases}
1&\text{if }t=1 \\
-1&\text{if }t=0
\end{cases}
\qquad\qquad
\alpha_n=\begin{cases}
\min\left\{\dfrac{N\zeta\tau_n}{2\eta_n},1-\pi_{n,1}^o\right\}&\text{if }\tau_n>0 \\
0&\text{if }\tau_n=0 \\
\max\left\{\dfrac{N\zeta\tau_n}{2\eta_n},-\pi_{n,1}^o\right\}&\text{if }\tau_n<0,
\end{cases}
\end{align*}
and where
\begin{align*}
\tau_n=\mu_{n,1}-\mu_{n,0}
\qquad\qquad\qquad
\eta_n=\sum_{t=0}^1\frac{\sigma^2_{n,t}}{\pi^o_{n,t}}
+\left(\sum_{t=0}^1\sqrt{\frac{1-\pi^o_{n,t}}{\pi^o_{n,t}}}\mu_{n,t}\right)^2.
\end{align*}
\end{repeattheorem}
Here, $\tau_n$ is the treatment effect for unit $n$, and $\eta_n$ is the variance of its IPW estimate:\footnote{The second term of $\eta_{n,t}$ involving $\mu_{n,t}$ may seem strange; it arises because the IPW estimator is not invariant to additive constants to the outcome variable. Indeed, normalizing outcomes can reduce variance of IPW, resulting in the augmented IPW (AIPW) estimator~\citep{robins1994estimation}.}
\begin{align*}
\tau_n &= \mathbb{E}\left[\left(\mathbbm{1}(T_{n}^o = 1) - \mathbbm{1}(T_{n}^o = 0) \right)\frac{Y_{n}^o}{\pi^o_{n,T_{n}^o}}\right]  \\
\eta_n &= \text{Var}\left[\left(\mathbbm{1}(T_{n}^o = 1) - \mathbbm{1}(T_{n}^o = 0)\right)\frac{Y_{n}^o}{\pi^o_{n,T_{n}^o}}\right]
\end{align*}
Furthermore, $\Pi_{\Lambda}$ is parameterized by the optimal value of the dual variable $\zeta$. It starts from the observational policy $\pi^*(0)=\pi^o$. As $\zeta$ increases, the treatment probability increases or decreases (depending on the sign of $\tau_n$) continuously at a rate of $N\tau_n/(2\eta_n)$ until it binds to either zero or one; the binding order is determined by the magnitude of this rate. This rate prioritizes units with higher expected treatment effect and lower estimation variance; notably, it is independent of the other units except through the total number of units $N$. The binding value optimizes the expected treatment effect; also, letting $\zeta_{\text{max}}$ denote the smallest $\zeta$ for which which the treatment probabilities are bound for all units, then $\pi^*(\zeta)$ is the expectation maximizing policy for all $\zeta\ge\zeta_{\text{max}}$.

\subsection{Structure of the Pareto Frontier}
\label{sec:binarystructure}

\begin{figure}[t]
\centering
\includegraphics[width=0.6\textwidth]{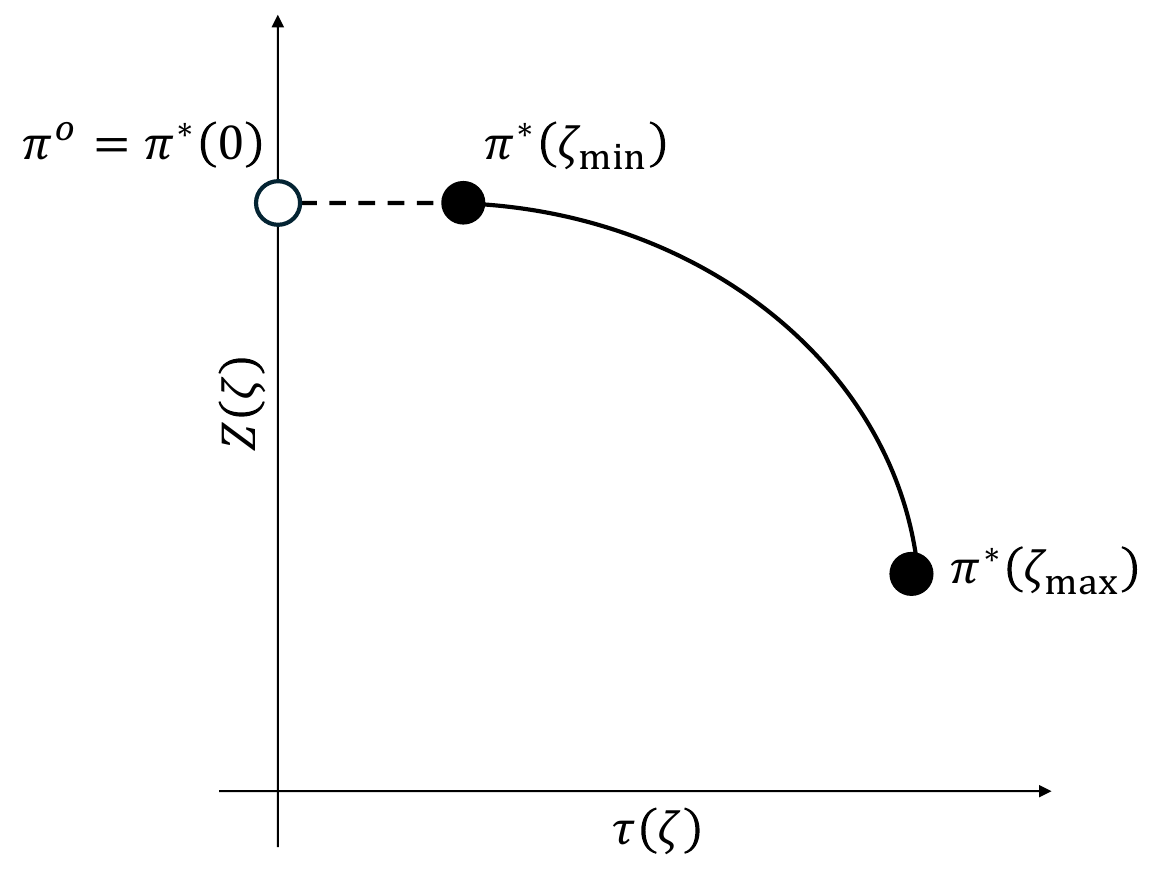}
\caption{Structure of the Pareto frontier.}
\label{fig:structure}
\end{figure}

Theorem~\ref{thm:binary} provides detailed insights into the structure of the Pareto frontier. First, we define
\begin{align*}
\mathcal{N}_u(\zeta)=\left\{n\in[N]\biggm\vert\alpha_n=\frac{N\zeta\tau_n}{\eta_n}\right\},
\end{align*}
i.e., $\mathcal{N}_u(\zeta)$ is the set of units for which $\pi_{n,t}^*(\zeta)$ is not bound to zero or one (but including the boundary case); we denote its complement by $\mathcal{N}_b(\zeta)=[N]\setminus\mathcal{N}_u(\zeta)$.

\begin{repeatproposition}
\label{prop:binarytau}
The treatment effect $\tau(\zeta)$ is monotonically increasing  in $\zeta$, and this increase is strict for all $\zeta>0$ such that $|\mathcal{N}_u(\zeta)|<N$.
\end{repeatproposition}
\begin{proof}{Proof.}
This result follows immediately from Eq.~(\ref{eqn:binary}). \qed
\end{proof}
The behavior of the $z$-score is only slightly more complicated (though its proof is less trivial):
\begin{repeatproposition}
\label{prop:binaryzscore}
The $z$-score $Z(\zeta)$ is monotonically decreasing in $\zeta$, and this decrease is strict for all $\zeta>0$ such that $1\le|\mathcal{N}_u(\zeta)|<N$; furthermore, it is constant for $\zeta>0$ such that $|\mathcal{N}_u(\zeta)|=0$.
\end{repeatproposition}
Before we provide a proof, we first describe the implications of Proposition~\ref{prop:binarytau} \&~\ref{prop:binaryzscore} for the structure of the Pareto frontier, which is illustrated in Figure~\ref{fig:structure}. First, note that the observational policy is $\pi^o=\pi^*(0)$; thus, we have $\tau(0)=0$ (i.e., observational policy has zero improvement over itself) and $Z(0)$ is undefined since $s^2(0)=0$ (intuitively, we can never ``prove'' that the observational policy is better than itself). Next, define
\begin{align*}
\zeta_{\text{min}}=\inf\{\zeta\ge0\mid\mathcal{N}_b(\zeta)\neq\varnothing\}
\qquad\qquad\qquad
\zeta_{\text{max}}&=\inf\{\zeta \geq 0\mid\mathcal{N}_u(\zeta)=\varnothing\}.
\end{align*}
Intuitively, $\zeta_{\text{min}}$ is the smallest value of $\zeta$ for which some unit has treatment probability bound to zero or one; analogously, $\zeta_{\text{max}}$ is the smallest value at which all units become bound to zero or one (so $\pi^*(\zeta_{\text{max}})\in\operatorname*{\arg\max}_{\pi}\tau(\pi)$ is the maximizer of the expected treatment effect).

By Proposition~\ref{prop:binaryzscore}, the $z$-score is constant for $\zeta\in(0,\zeta_{\text{min}}]$, and by Proposition~\ref{prop:binarytau}, the expected treatment effect is strictly increasing along this interval. Thus, $\zeta_{\text{min}}$ is a strictly better choice than the remainder of the interval, so we prune $(0,\zeta_{\text{min}})$ from the frontier. Now, Propositions~\ref{prop:binarytau} \&~\ref{prop:binaryzscore} say that on $[\zeta_{\text{min}},\zeta_{\text{max}}]$, the $z$-score is strictly decreasing and the expected treatment effect is strictly increasing; thus, all policies $\pi^*(\zeta)$ for $\zeta_{\text{min}}\le\zeta\le\zeta_{\text{max}}$ are part of the frontier. Finally, the policy becomes constant for $\zeta\ge\zeta_{\text{max}}$, so we also prune $(\zeta_{\text{max}},\infty)$ from the frontier. Thus:
\begin{repeatcorollary}
\label{cor:pareto}
The Pareto frontier is $\Pi_{\text{opt}}=\{\pi^*(\zeta)\mid\zeta\in[\zeta_{\text{min}},\zeta_{\text{max}}]\}$.
\end{repeatcorollary}
\begin{proof}{Proof of Proposition~\ref{prop:binaryzscore}.}
Note that
\begin{align*}
Z'(\zeta)=\frac{\tau'(\zeta)s(\zeta)-s'(\zeta)\tau(\zeta)}{s^2(\zeta)}.
\end{align*}
Thus, it suffices to show that  $\frac{s'(\zeta)}{s(\zeta)}-\frac{\tau'(\zeta)}{\tau(\zeta)}\ge0$, and that this inequality is strict when $|\mathcal{N}_b(\zeta)|>0$. Note that $(s^2)'(\zeta)=2s(\zeta)s'(\zeta)$, so in fact it suffices to show that
\begin{align*}
\Delta(\zeta)\coloneqq\frac{(s^2)'(\zeta)}{2s^2(\zeta)}-\frac{\tau'(\zeta)}{\tau(\zeta)}\ge0.
\end{align*}
By our definition of $\mathcal{N}_u(\zeta)$ and $\mathcal{N}_b(\zeta)$, we have
\begin{align*}
\tau(\zeta)&=\frac{1}{N}\sum_{n\in\mathcal{N}_b(\zeta)}\tau_n\alpha_n+\frac{\zeta}{2}\sum_{n\in\mathcal{N}_u(\zeta)}\frac{\tau_n^2}{\eta_n} \\
s^2(\zeta)&=\frac{1}{N^2}\sum_{n\in\mathcal{N}_b(\zeta)}\eta_n\alpha_n^2+\frac{\zeta^2}{4}\sum_{n\in\mathcal{N}_u(\zeta)}\frac{\tau_n^2}{\eta_n}.
\end{align*}
Defining $A(\zeta)=N^{-1}\sum_{n\in\mathcal{N}_b(\zeta)}\tau_n\alpha_n$, $B(\zeta)=N^{-2}\sum_{n\in\mathcal{N}_b(\zeta)}\eta_n\alpha_n^2$, and $D(\zeta)=\sum_{n\in\mathcal{N}_u(\zeta)}(\tau_n^2/\eta_n)$, then
\begin{align*}
\tau(\zeta)=A(\zeta)+\frac{\zeta}{2}D(\zeta),
\qquad
s^2(\zeta)=B(\zeta)+\frac{\zeta^2}{4}D(\zeta).
\end{align*}
Since $A(\zeta)$, $B(\zeta)$, and $D(\zeta)$ are piecewise constant, we have
\begin{align*}
\Delta(\zeta)
&=\frac{D(\zeta)}{(4B(\zeta)/\zeta)+\zeta D(\zeta)}-\frac{D(\zeta)}{2A(\zeta)+\zeta D(\zeta)}.
\end{align*}
Thus, it suffices to show that $2B(\zeta)/\zeta\ge A(\zeta)$. Note that by definition, for any $n\in\mathcal{N}_b$, we have
\begin{align*}
|\alpha_n|>\frac{N\zeta|\tau_n|}{2\eta_n}
\end{align*}
As a consequence, we have
\begin{align*}
\frac{2B(\zeta)}{\zeta}
=\frac{1}{N}\sum_{n\in\mathcal{N}_b(\zeta)}\frac{2\eta_n\alpha_n^2}{N\zeta}
\ge\frac{1}{N}\sum_{n\in\mathcal{N}_b(\zeta)}|\tau_n|\cdot|\alpha_n|
=A(\zeta),
\end{align*}
where the inequality is strict if $\mathcal{N}_b(\zeta)\neq\varnothing$; here, we have used the fact that $|\tau_n|\cdot|\alpha_n|=\tau_n\alpha_n$

So far, we have ignored the fact that $Z(\zeta)$ is not everywhere differentiable. However, it is everywhere continuous and differentiable except potentially on a finite set of points. As a consequence, if $Z'(\zeta)<0$ on an open interval $(\zeta_{\text{min}},\zeta_{\text{max}})$ except at points where $Z'$ is not defined, then it follows that $Z$ is strictly monotone everywhere on that interval.

It remains to show that $Z(\zeta)$ is constant when $\mathcal{N}_b(\zeta)=\varnothing$. In this case, we have
\begin{align*}
Z(\zeta)
=\frac{\tau(\zeta)}{s(\zeta)}
=\frac{\frac{\zeta}{2}\sum_{n=1}^N\frac{\tau_n^2}{\eta_n}}{\sqrt{\frac{\zeta^2}{4}\sum_{n=1}^N\frac{\tau_n^2}{\eta_n}}}
=\sqrt{\sum_{n=1}^N\frac{\tau_n^2}{\eta_n}},
\end{align*}
which does not depend on $\zeta$. The claim follows. \qed
\end{proof}

\subsection{Discussion}
\label{sec:binarydiscussion}

Here, we discuss several implications of our results for policy learning algorithms targeting statistically significant performance improvements. While our discussion is based on the characterization in a specialized setting, we show that our key insights extend to the general setting in Section~\ref{sec:multi}; thus, our discussion considers the general case.

\paragraph{Comparing policy learning methodologies.}

Prior work has proposed \emph{ad hoc} techniques for policy learning targeting statistical significance. They can be grouped into three strategies:
\begin{itemize}
\item \textbf{Variance regularization:} These approaches select policies that minimize some combination of the mean and variance of the treatment effect. In the idealized setting where all $\mu_{n,t}$ and $\sigma_{n,t}$ are known, variance regularization computes a policy along our Pareto frontier.
\item \textbf{Treatment aggregation:}  These approaches merges similar treatments, which mitigates the winner's curse since they select among fewer candidate policies. Note that this strategy can only be applied when multiple treatments are available.
\item \textbf{Unit aggregation:} These approaches aggregate similar units; with the same goal of reducing the number of candidate policies as treatment aggregation.
\end{itemize}
To compare these approaches, we introduce the following standard definitions.
\begin{definition}
\rm
$\pi$ is \emph{deterministic} if $\forall n\in[N],t\in\mathcal{T}$, $\pi_{n,t}\in\{0,1\}$; otherwise, it is \emph{stochastic}.
\end{definition}
\begin{definition}
$\pi$ is \emph{untargeted} if $\forall t\in\mathcal{T}$, $\pi_{n,t}=\pi_t$ is independent of $n$; otherwise, it is \emph{targeted}.
\end{definition}
One of our key insights is that policies along the Pareto frontier are largely stochastic.
\begin{repeatcorollary}
All policies $\pi\in\Pi_{\text{opt}}$ are stochastic except $\pi^*=\pi^*(\zeta_{\text{max}})$.
\end{repeatcorollary}
Now, we consider the effect of treatment aggregation on the policy space. Treatment aggregation typically considers a large number of treatment options, but does not consider targeting---i.e., we have a finite policy class $\Pi=\{\pi^t\}_{t\in\mathcal{T}}$, where $\pi^t_{n,t'}=\mathbbm{1}(t'=t)$\footnote{Targeting can be in principle encoded by having the ``treatment'' vary depending on the unit, but this is not typical.}. In other words, policy $\pi^t$ deterministically assigns treatment $t$ to all units. Now, merging treatments $t_1,...,t_H\in\mathcal{T}$ is mathematically achieved by creating a new policy $\tilde{\pi}$ that uniformly randomly selects among these treatments---i.e.,
\begin{align*}
\tilde{\pi}_{n,t}=\frac{1}{H}\sum_{h=1}^H\pi_{n,t}^{t_h}.
\end{align*}
Then, $\pi^{t_1},...,\pi^{t_H}$ are removed from $\Pi$ and $\tilde{\pi}$ is added to $\Pi$. Note that $\tilde{\pi}$ is stochastic, which is consistent with our insight that Pareto optimal policies are stochastic. However, in our analysis in Section~\ref{sec:multi}, we show that the policy $\pi^*(\zeta_{\text{min}})$ is achieved by increasing $\zeta$ until $\pi_{n,t}^*(\zeta)$ binds to zero for \emph{any} unit $n\in[N]$ and treatment $t\in\mathcal{T}$; thus, $\pi^*(\zeta_{\text{min}})$ typically randomizes over all treatments. It may be possible to improve treatment aggregation by randomly  selecting among all treatments in a non-uniform way rather than merging a few high-value treatments.

Unit aggregation approaches also assume a finite policy class $\Pi$, where in this case each policy $\pi\in\Pi$ is a targeted policy; then, it identifies similar units $n_1,...,n_H\in[N]$ and imposes the constraint that $\pi_{n_h}=\pi_{n_{h'}}$ for all $h,h'\in[H]$. A typical strategy is to simply remove policies that do not satisfy this constraint from $\Pi$; thus, if $\Pi$ initially only contains deterministic policies, then unit aggregation never introduces any stochasticity into $\Pi$. As a consequence, we might expect these approaches to be more effective if $\Pi$ initially contains stochastic policies.

\paragraph{Statistical power.}

So far, we have been assuming the observational dataset is fixed; however, a common question in practice is how to collect the sample $\mathcal{O}$ to achieve a target level of significance. To address this question, we need to understand $Z(\zeta_{\text{min}})$, which is the best possible $z$-score that can be achieved along the Pareto frontier. In fact, we have an explicit formula for $Z(\zeta_{\text{min}})$:
\begin{repeatcorollary}
\label{cor:binaryzbest}
The best possible $z$-score is achieved by $\pi^*(\zeta_{\text{min}})$, and equals
\begin{align*}
Z(\zeta_{\text{min}})=\sqrt{\sum_{n=1}^NZ_n},
\qquad\text{where}\qquad
Z_n=\frac{\tau_n^2}{\eta_n}.
\end{align*}
\end{repeatcorollary}
This result follows from the proof of Proposition~\ref{prop:binaryzscore}. Given assumptions about $\mu_{n,0}$, $\mu_{n,1}$, and $\sigma_n$, it can be used to calculate the number of samples needed for a target level of statistical significance. It also quantifies the value of each unit toward statistical significance, which might help identify promising test subjects when recruitment is costly \citep{kouvelis_clinical_2017,tian_optimal_2023}. For instance, suppose we further assume the following: (1) $\pi^o$ is an RCT policy (i.e., $\pi^o_{n,t}=1/2$ for all $n\in[N]$ and $t\in\mathcal{T}$), (2) the outcome variances are independent of treatment (i.e., $\sigma_{n,t}=\sigma_n$ for some $\sigma_n$), and (3) outcomes are centered (i.e., $\mu_{n,1}=-\mu_{n,0}=\mu_n$ for some $\mu_n$; for instance, the augmented IPW estimator aims to center outcomes). Then, we have
\begin{align*}
Z_n=\frac{\mu_n^2}{\sigma_n^2},
\end{align*}
which is exactly the signal-to-noise ratio (SNR) for unit $n$. For instance, note that
\begin{align*}
\frac{\partial}{\partial Z_n}Z(\zeta_{\text{min}})=\frac{1}{2Z(\zeta_{\text{min}})}.
\end{align*}
From this result, we observe that the marginal value of increasing the SNR $Z_n$ of unit $n$ only depends on the current $z$-score $Z(\zeta_{\text{min}})$. Intuitively, the marginal value decreases proportional to $Z(\zeta_{\text{min}})$. Thus, it is always better to allocate resources to recruiting units with higher SNR, but the overall efficacy of increasing SNR has diminishing returns.

Finally, we can compare $\pi^*(\zeta_{\text{min}})$ to $\pi^*(\zeta_{\text{max}})$, which is given by
\begin{align*}
\pi^*_{n,t}(\zeta_{\text{max}})&=\begin{cases}
1&\text{if }\tau_n>0 \\
\pi^o_{n,t}&\text{if }\tau_n=0 \\
0&\text{if }\tau_n<0.
\end{cases}
\end{align*}
While $Z(\zeta_{\text{max}})$ can be complicated in general, it simplifies when $\pi^o$ is uniform.
\begin{repeatcorollary}
Suppose that $\pi_{n,t}^o=1/2$ for all $n\in[N]$ and $t\in\mathcal{T}$. Then, we have
\begin{align*}
Z(\zeta_{\text{max}})=\frac{\sum_{n=1}^N|\tau_n|}{\sqrt{\sum_{n=1}^n\eta_n}}.
\end{align*}
\end{repeatcorollary}
This result follows from the proof of Proposition~\ref{prop:binaryzscore}. Under conditions (1), (2), and (3) above,
\begin{align*}
Z(\zeta_{\text{max}})=\frac{\sum_{n=1}^N|\mu_n|}{\sqrt{\sum_{n=1}^N\sigma_n^2}}.
\end{align*}
In particular, in this case, it is no longer true that $Z(\zeta_{\text{max}})$ can be expressed as in terms of the SNRs of individual units. In general, $Z(\zeta_{\text{min}})\ge Z(\zeta_{\text{max}})$; this inequality follows from Proposition~\ref{prop:binaryzscore}, but it also follows directly from Sedrakyan's inequality~\citep{sedrakyan2018useful}. It is tight when $\mu_n=\mu$ and $\sigma_n=\sigma$ are constant; indeed, in this case, it is easy to see that $\pi^*(\zeta_{\text{min}})=\pi^*(\zeta_{\text{max}})$. To understand when there is a gap, consider the following:
\begin{itemize}
\item \textbf{Arbitrarily large gap:} Consider two units with $\mu_1=\mu_2=1$, $\sigma_1=1$, and $\sigma_2=\sigma$; then
\begin{align*}
Z(\zeta_{\text{min}})=\sqrt{1+\frac{1}{\sigma^2}}\to\infty
\qquad\qquad
Z(\zeta_{\text{max}})=\frac{2}{\sqrt{1+\sigma^2}}\to2
\end{align*}
as $\sigma\to0$. Intuitively, $\pi^*(\zeta_{\text{min}})$ ``drops'' the first unit from the analysis (i.e., taking $\pi_{1,t}^*(\zeta_{\text{min}})\to\pi_{1,t}^o$ while $\pi_{2,t}^*(\zeta_{\text{min}})\to \mathbbm{1}(t=1)$), so as $\sigma\to 0$, it achieves complete confidence in policy improvement. In contrast, $\pi^*(\zeta_{\text{max}})$ always treats both units (i.e., taking $\pi_{n,t}^*(\zeta_{\text{max}})=\mathbbm{1}(t=1)$; thus, it always retains the variance from treating the first unit, bounding its confidence.
\item \textbf{Asymptotic gap:} There is a gap between the two even as $N\to\infty$. We can easily show a constant multiplicative gap; consider two unit types, where there are $N$ units of the first type (with $\mu_n=1$ and $\sigma_n=1$) and $N$ of the second type (with $\mu_n=1$ and $\sigma_n=\sigma$, for $\sigma\ll1$). Then, we have
\begin{align*}
Z(\zeta_{\text{min}})
=\sqrt{N}\cdot\sqrt{1+\frac{1}{\sigma^2}}
\ge\sqrt{N}\cdot\frac{1}{\sigma}
\qquad\qquad
Z(\zeta_{\text{max}})
=\sqrt{N}\cdot\frac{2}{\sqrt{1+\sigma^2}}
\le\sqrt{N}\cdot 2
\end{align*}
We can also show a non-constant gap, though it requires the variance of each unit to grow with $n$; specifically, letting $\mu_n=1$ for all $n$ and $\sigma_n=\sqrt{n}$, then we have
\begin{align*}
Z(\zeta_{\text{min}})=\sqrt{\sum_{n=1}^N\frac{1}{n}}\ge\sqrt{\log N}
\qquad\qquad
Z(\zeta_{\text{max}})=\frac{N}{\sqrt{\sum_{n=1}^Nn}}
=\frac{\sqrt{2}N}{\sqrt{N(N+1)}}\le \sqrt{2}.
\end{align*}
Intuitively, this scenario involves adding higher and higher variance units to the sample, which gradually improves $Z(\zeta_{\text{min}})$ but not $Z(\zeta_{\text{max}})$.
\item \textbf{Non-monotonicity of $Z(\zeta_{\text{max}})$:}
Consider two unit types, where there are $N$ units of the first type (with $\mu_n=\mu\ll1$ and $\sigma_n=1$) and $M\gg1$ of the second (with $\mu_n=1$ and $\sigma_n=1$). Then
\begin{align*}
Z(\zeta_{\text{min}})=\sqrt{N\mu+M}
\qquad\qquad
Z(\zeta_{\text{max}})=\frac{N\mu+M}{\sqrt{N+M}}.
\end{align*}
While $Z(\zeta_{\text{min}})$ is monotonically increasing in $N$ (in general, it is monotonically increasing as units are added), this is not true for $Z(\zeta_{\text{max}})$ for sufficiently small $\mu$ and large $M$. Intuitively, $\pi^*(\zeta_{\text{max}})$ initially performs well due to the $M$ ``good'' units, so adding a small number of ``bad'' units actually reduces its $z$-score by adding variance. Eventually, adding sufficiently many bad units overcomes this variance to improve the $z$-score. If we choose $\sigma_n$ to increase with $n$, $Z(\zeta_{\text{max}})$ can be made to monotonically decrease in $n$; specifically, letting $\mu_n=1$ and $\sigma_n=n$ for all $n$, then we have
\begin{align*}
Z(\zeta_{\text{min}})=\sqrt{\sum_{n=1}^N\frac{1}{n^2}}
\qquad\qquad
Z(\zeta_{\text{max}})=\frac{N}{\sqrt{\sum_{n=1}^Nn^2}}=\frac{N}{\sqrt{N(N+1)(2N+1)/6}}.
\end{align*}
As $N\to\infty$, then $Z(\zeta_{\text{max}})\to0$ and $Z(\zeta_{\text{min}})\to\sqrt{\pi^2/6}$ (here, $\pi$ refers to the number, not the policy). That is, if we add units with successively higher variance by a sufficient increment, this variance actually decreases the $z$-score of $\pi^*(\zeta_{\text{max}})$.
\end{itemize}

\section{Multiple Treatment Setting}
\label{sec:multi}

Next, we consider the general case of multiple treatments. Mirroring our analysis in the single-treatment analysis, we begin by computing the solutions to our optimization problem, which form a superset of the Pareto frontier (Section~\ref{sec:multisuperset}). Then, we establish the structure of the Pareto frontier, which is essentially identical to the structure in Figure~\ref{fig:structure} for the single-treatment setting (Section~\ref{sec:multistructure}). Finally, we provide results on the structure of policies along the Pareto frontier (Section~\ref{sec:multipolicy}) and the best possible statistical power (Section~\ref{sec:multipower}).  Proofs for all claims are given in Appendix~\ref{apx:proofs}.

\subsection{Superset of the Pareto Frontier}
\label{sec:multisuperset}

The lack of a well defined treatment effect when comparing multiple treatments complicates the multiarm characterization of $\Pi_{\Lambda}$ ($K>1$) relative to the single-treatment setting. The policies in $\Pi_{\Lambda}$ can still be parameterized by the dual variable which uniquely generates them $\zeta$ and still form a path of policies from $\pi^o$ to the expectation maximizing policy. However, for each unit $n$, the rate of change for each non-zero propensity $\pi^*_{n,t}(\zeta)$ is impacted by which of the unit's treatments receive zero propensity. These inactive treatments do not impact the weighted average of treatments which act as a control for each treatment's relative impact, causing $\pi^*_{n,t}(\zeta)$ to change in a piece-wise linear manner as $\zeta$ increases. To overcome this issue, our general characterization of $\Pi_{\Lambda}$ first determines the value of $\zeta$ at which each propensity $\pi^*_{n,t}(\zeta)$ first hits $0$, and then determines the non-zero propensities in terms of the set of zero propensity treatments. Doing so allows us to first identify all of the knot points in the piecewise linear evolution of $\pi^*(\zeta)$, and then describe $\pi^*(\zeta)$ in terms of its locally linear form. First, we have: \begin{repeatproposition}
\label{prop:zeros}
Let $(\pi',\zeta',\kappa')$ and $(\pi'',\zeta'',\kappa'')$ both represent policies in $\Pi_{\Lambda}$ with their accompanying dual variables. Furthermore, for any policy $\pi^\cdot\in \Pi_{\Lambda}$ let
$$\omega_n^\cdot = \zeta^\cdot + N\sum_{t=0}^K \frac{\pi_{n,t}^o}{\mu_{n,t}^2 + \sigma_{n,t}^2} \kappa_{n,t}^\cdot \left(\mu_{n,t} - \tilde{\mu}_n\right).$$ If $\omega_n'' \geq \omega_n' > 0$ and $\pi'_{n,t} = 0$ for some unit $n$ and treatment $t$, then $\pi''_{n,t} = 0$.
\end{repeatproposition}
Proposition~\ref{prop:zeros} ensures that for each $\omega^*$ which generates a solution, there exists a unique set $\mathcal{Z}_n(\omega^*) = \{t \mid \pi^*_{n,t}(\zeta^*) = 0\}$ of treatments which receive no assignments for unit $n$, and that $\mathcal{Z}_n(\omega')\subseteq\mathcal{Z}_n(\omega'')$ if $\omega'' \geq \omega' > 0$ both generate policies in $\Pi_{\Lambda}$. This provides enough structure to uniquely solve for $\kappa^*$ in Proposition~\ref{prop:form} for any $\omega^* \in \mathbb{R}_+$, generating the full solution to $\Pi_{\Lambda}$ (as $\omega^* < 0$ induces $\lambda < 0$), which can still be uniquely parameterized in terms of $\zeta^* \geq 0$.
\begin{repeattheorem}
\label{thm:full_char}
For each unit $n$, set  $\omega_n^0 = 0$ and $\hat{\mathcal{Z}}_n = \{ \}$, and apply this procedure to form $\zeta_{n,t}$:
\begin{itemize}
\item[1)] If $\mu_{n,v} = \mu_{n,u} \ \forall \ v,u \not\in \hat{\mathcal{Z}}_n$, set $\zeta_{n,t} = \infty \ \forall \ t \not\in \hat{\mathcal{Z}}_n$, then all $\zeta_{n,t}$ have been formed. Otherwise move to step 2).
\item[2)] For all $t \not \in \hat{\mathcal{Z}}_n$, form
$$\hat{\omega}_{n,t} = \frac{2Q_n\left(\left(\mu_{n,t}^2 + \sigma_{n,t}^2\right)\sum_{v\not\in \hat{\mathcal{Z}}_n} \frac{\pi^o_{n,v}}{\mu_{n,v}^2 + \sigma_{n,v}^2} + \sum_{v\in \hat{\mathcal{Z}}_n} \pi^o_{n,v}\right)}{N\sum_{v\not \in \hat{\mathcal{Z}}_n} \frac{\pi^o_{n,v}}{\mu_{n,v}^2 + \sigma_{n,v}^2} \left(\mu_{n,v}-\mu_{n,t}\right)},$$
where $Q_n$ is as defined in Proposition~\ref{prop:form}
\item[3)] Update $\omega^0_n \leftarrow \min\left(\left\{\hat{\omega}_{n,t}\Bigm\vert t\not\in\hat{\mathcal{Z}}_n,\hat{\omega}_{n,t} > \omega^0_n\right\}\right)$. If $\hat{\omega}_{n,t} = \omega_n^0$, then set,
\begin{align*}
\zeta_{n,t} &= \left[1 - \sum_{v\not\in\hat{\mathcal{Z}}_n}\frac{\pi^o_{n,v}}{\mu_{n,v}^2+\sigma_{n,v}^2} \left(\mu_{n,v} - \frac{\sum_{j\not\in\hat{\mathcal{Z}}_n}\frac{\pi^o_{n,j}}{\mu_{n,j}^2+\sigma_{n,j}^2}\mu_{n,j}}{\sum_{j\not\in\hat{\mathcal{Z}}_n}\frac{\pi^o_{n,j}}{\mu_{n,j}^2+\sigma_{n,j}^2}}\right)^2\right] \frac{\omega_n^0}{Q_n}\\
&\hspace{10pt}+ \frac{2}{N} \sum_{v\in\hat{\mathcal{Z}}_n} \pi_{n,v}^o \left(\mu_{n,v} - \frac{\sum_{j\not\in\hat{\mathcal{Z}}_n}\frac{\pi^o_{n,j}}{\mu_{n,j}^2+\sigma_{n,j}^2} \mu_{n,j}}{\sum_{j\not\in\hat{\mathcal{Z}}_n}\frac{\pi^o_{n,j}}{\mu_{n,j}^2+\sigma_{n,j}^2}}\right)
\end{align*}
\item[4)] Update $\hat{\mathcal{Z}}_n \leftarrow \hat{\mathcal{Z}}_n \cup \left\{t\not \in \hat{\mathcal{Z}}_n\middle| \hat{\omega}_{n,t} = \omega^0_n\right\}$. Return to step 1).
\end{itemize}

Let  $\mathcal{Z}_n(\zeta) = \{t |\zeta_{n,t} \leq \zeta \}.$

For all $K \geq 1$, we have $\Pi_{\Lambda} = \bigcup_{\zeta \geq 0}\left\{\pi^*(\zeta) \right\},$ where $\pi^*_{n,t}(\zeta) = 0 \ \forall \ t \in \mathcal{Z}_n(\zeta)$ while
\begin{align*}
\pi^*_{n,t}(\zeta)=& \pi^o_{n,t} + \left(\frac{\frac{\pi_{n,t}^o}{\mu_{n,t}^2+\sigma_{n,t}^2}}{\sum_{v\not\in\mathcal{Z}_n(\zeta)} \frac{\pi_{n,v}^o}{\mu_{n,v}^2+\sigma_{n,v}^2}}\right)\left(\sum_{v\in\mathcal{Z}_n(\zeta)} \pi^o_{n,v} \right)  \\&+ \left(\mu_{n,t} - \frac{\sum_{v\not\in\mathcal{Z}_n(\zeta)}\frac{\pi^o_{n,v}}{\mu_{n,v}^2+\sigma_{n,v}^2} \mu_{n,v}}{\sum_{v\not\in\mathcal{Z}_n(\zeta)}\frac{\pi^o_{n,v}}{\mu_{n,v}^2+\sigma_{n,v}^2}}\right)\frac{N\omega_n^*(\zeta)\pi^o_{n,t}}{2Q_n\left(\mu_{n,t}^2 + \sigma_{n,t}^2\right)}
\end{align*}
where,
\begin{align*}
\omega_n^*(\zeta)&= \frac{\zeta - \frac{2}{N}\left( \sum_{t\in\mathcal{Z}_n(\zeta)} \pi_{n,t}^o \left(\mu_{n,t} - \frac{\sum_{v\not\in\mathcal{Z}_n(\zeta)}\frac{\pi^o_{n,v}}{\mu_{n,v}^2+\sigma_{n,v}^2} \mu_{n,v}}{\sum_{v\not\in\mathcal{Z}_n(\zeta)}\frac{\pi^o_{n,v}}{\mu_{n,v}^2+\sigma_{n,v}^2}}\right)\right)}{1 - \sum_{t\not\in\mathcal{Z}_n(\zeta)}\frac{\pi^o_{n,t}}{\mu_{n,t}^2+\sigma_{n,t}^2} \left(\mu_{n,t} - \frac{\sum_{v\not\in\mathcal{Z}_n(\zeta)}\frac{\pi^o_{n,v}}{\mu_{n,v}^2+\sigma_{n,v}^2}\mu_{n,v}}{\sum_{v\not\in\mathcal{Z}_n(\zeta)}\frac{\pi^o_{n,v}}{\mu_{n,v}^2+\sigma_{n,v}^2}}\right)^2} Q_n
\end{align*}
for all $t \not \in \mathcal{Z}_n(\zeta)$.
\end{repeattheorem}
To compute $\mathcal{Z}_n(\zeta)$ for each unit, we iteratively compute which treatment's propensity will first hit zero for each unit, remove that treatment and continue. An added benefit to the computation and analysis of $\pi^*(\zeta)$ is that each unit's treatment propensities can be calculated independently. We see that as $\zeta$ increases, $\pi^*(\zeta)$ moves from the observational policy to the expectation maximizing policy by iteratively eliminating suboptimal treatment arms for each unit.

The non-zero propensities of $\pi^*(\zeta)$ are determined by a term which redistributes zero propensity treatments and a linear term similar to the single-treatment characterization. This term replaces $\tau_n$ with a multi-treatment version which compares the expected outcome of unit $n$ in response to treatment $t$ to a weighted average of the expected outcome of the other active treatments. The variance of the conditional treatment effect is then replaced by $Q_n^{-1} \frac{\pi^o_{n,t}}{\mu_{n,t}^2 + \sigma^2_{n,t}}$. The treatment specific term here, which also determines how zero propensity treatments are redistributed as well as the weighted mean each treatment is measured against, is once again related to IPW's evaluation:
$$\frac{\mu_{n,t}^2 + \sigma_{n,t}^2}{\pi^o_{n,t}} = \mathbb{E}\left[\left(\mathbbm{1}\left[T_n^o = t\right] \frac{Y_n^o}{\pi^o_{n,T_n^o}}\right)^2\right] = \mathrm{Var}\left(\mathbbm{1}\left[T_n^o = t\right] \frac{Y_n^o}{\pi^o_{n,T_n^o}}\right) + \mathbb{E}\left[\mathbbm{1}\left[T_n^o = t\right] \frac{Y_n^o}{\pi^o_{n,T_n^o}}\right]^2.$$
Thus, as $\zeta$ increases, $\pi^*(\zeta)$ evolves similarly to the single-treatment case; it compares each treatment against the other active treatments, and then changes each unit's policy to increase the policies impact while introducing minimal variance to an estimate of the policy's improvement.

\subsection{Structure of the Pareto Frontier}
\label{sec:multistructure}

As in the single-treatment setting, the expected improvement is strictly increasing as $\pi^*(\zeta)$ moves from the observational policy toward the expectation-maximizing policy, while the expected $z$-score is constant for small $\zeta$, and then strictly decreasing after any single unit has a treatment reach zero propensity.
Define $\tau(\zeta)=\tau(\pi^*(\zeta))$, $s^2(\zeta)=s^2(\pi^*(\zeta))$, and $Z(\zeta)=\mathbb{E}[Z(\pi^*(\zeta))]=\frac{\tau(\zeta)}{s^2(\zeta)}$; also,
\begin{align*}
\zeta_{\text{max}} &= \operatorname*{\arg\min}_{\zeta \geq 0} \{\zeta\mid\mathcal{Z}_n(\zeta) = \lim_{\zeta\to \infty} \mathcal{Z}_n(\zeta) \ \forall\ n \in [N] \} \\
\zeta_{\text{min}} &= \operatorname*{\arg\min}_{\zeta \geq 0}\{\zeta\mid\min_{n\in[N],t\in\{1,\ldots,K\}}\pi_{n,t}^*(\zeta) = 0\}.
\end{align*}
In other words, $\zeta_{\text{max}}$ is the largest $\zeta$ for which a treatment propensity first reaches $0$. For $\zeta\geq \zeta_{\text{max}}$ we know $\pi^*(\zeta)$ recovers the expectation maximizing policy; up until that point, $\pi^*(\zeta)$ increases the variance of its policy evaluation to improve its expectation as $\zeta$ increases. Similarly, $\zeta_{\text{min}}$ is the smallest $\zeta$ for which some unit is never assigned one of the treatments.
\begin{repeatproposition} 
\label{prop:tau_deriv}
$\tau(\zeta)$ is strictly increasing for $\zeta\in(0,\zeta_{\text{max}})$. 
\end{repeatproposition}
\begin{repeatproposition}
\label{prop:z_deriv}
$Z(\zeta)$ is constant in $\zeta$ for $\zeta\in(0,\zeta_{\text{min}})$ and
strictly decreasing for $\zeta \in (\zeta_{\text{min}},\zeta_{\text{max}})$.
\end{repeatproposition}
\begin{repeattheorem}
\label{thm:ParetoEquivalance} 
The Pareto frontier is $\Pi_{\mathrm{Opt}} = \{\pi^*(\zeta)\mid\zeta \in [\zeta_{\text{min}},\zeta_{\text{max}}]\}$.
\end{repeattheorem}
This result mirrors Corollary~\ref{cor:pareto} in the single-treatment case. The main difference is that only a single unit-treatment pair needs to bind to zero at $\zeta_{\text{min}}$; thus, all units may have stochastic treatments.

\subsection{Structure of Policies on the Pareto Frontier}
\label{sec:multipolicy}

Next, we analyze how policies on the Pareto frontier assign treatments that trade off statistically significance vs. expected improvement. At a high level, as $\zeta$ increases, for each unit, policies along the frontier increase the propensity assigned to treatments with expected value above some unit-specific threshold, while decreasing the propensity assigned to the remaining treatments. In addition, a treatment's propensity will remain closer to that of the observational policy if that treatment's counterfactual outcome is often far from $0$. Thus policies on the frontier react conservatively to variance, choosing to adjust the propensities of treatments leading to certain outcomes sooner than uncertain ones.

Our first result shows that policies along the frontier decrease treatment propensities below the observational policy in order of their expected impact.
\begin{repeatproposition}
\label{prop:order_0}
For unit $n$, if $t$ and $t'$ are treatments with $\mu_{n,t} > \mu_{n,t'}$ and $\frac{\pi_{n,t'}^*(\zeta)}{\pi_{n,t'}^o} > 1$, then 
$$\frac{\pi_{n,t}^*(\zeta)}{\pi_{n,t}^o} > 1.$$
\end{repeatproposition}
This result follows a simple logic: for any policy that does not obey this ordering, we can select a pair of offending treatments and contract their propensities towards the observational policy to both increase expected performance and decrease the standard error of the IPW estimator. As such we can understand policies on the frontier as setting a reservation value for each unit; increasing the frequency of treatments in the subgroup that outperform this reservation value while decreasing the frequency of treatments which underperform the reservation value.

However, we also show that policies along the frontier adjust treatments whose counterfactual outcomes are far from 0 more conservatively, meaning a treatment whose average impact is worse than an alternate may be assigned relatively more. 
\begin{repeatproposition}
\label{prop:order_breaks}
For unit $n$, if $t$ and $t'$ are treatments with $\mu_{n,t} > \mu_{n,t'}$ and $\frac{\pi^*_{n,t'}(\zeta)}{\pi^o_{n,t'}} > \frac{\pi^*_{n,t}(\zeta)}{\pi^o_{n,t}}$, then
\begin{itemize}
\item[(1)] $\frac{\pi^*_{n,t'}(\zeta)}{\pi^o_{n,t'}}> 1 \implies \underbrace{\mathbb{E}[Y_{n,t}^2]}_{\mu_{n,t}^2 + \sigma_{n,t}^2} > \underbrace{\mathbb{E}[Y_{n,t'}^2]}_{\mu_{n,t'}^2 + \sigma_{n,t'}^2}$
\item[(2)] $\frac{\pi^*_{n,t'}(\zeta)}{\pi^o_{n,t'}} < 1 \implies \underbrace{\mathbb{E}[Y_{n,t}^2]}_{\mu_{n,t}^2 + \sigma_{n,t}^2} < \underbrace{\mathbb{E}[Y_{n,t'}^2]}_{\mu_{n,t'}^2 + \sigma_{n,t'}^2}$
\end{itemize}
\end{repeatproposition}
Note that under Proposition~\ref{prop:order_0}, the conditions of Proposition~\ref{prop:order_breaks} can only occur if the frontier policy increases both treatments' propensities above that of the observational policy, or decreases both below that of the observational policy. Thus condition (1) says that if the frontier policy increases both treatments, it may increase the lower mean treatment relatively more if its average outcome is closer to zero, while condition (2) says that if the frontier policy decreases both treatments, it may decrease the lower mean treatment relatively less if its average outcome is closer to zero. Re-ranking treatments in this way reduces the variance of the IPW estimator as the estimator scales observed outcomes in the test set with a magnitude that increases with the difference between the frontier policy and the observational policy. If outcomes are far from zero, this scaling can introduce excessive variance that destroys the potential of the IPW estimator to achieve significance, leading policies on the Pareto frontier to adjust treatments with counterfactual outcomes close to zero more than other treatments.

\subsection{Statistical Power}
\label{sec:multipower}

Finally, we compute the maximum expected z-score any policy can achieve, which is the analog of Corollary~\ref{cor:binaryzbest} in the general setting.
\begin{repeatproposition}
\label{prop:bestZ}
Let
\begin{align*}
\pi_{n,t}^\dagger &= \frac{\mathbb{E}\left[\mathbbm{1}[T_n^o = t]\mathbb{E}[Y_{nT_n^o}^2\mid T_n^o]^{-1}\right]}{\mathbb{E}\left[\mathbb{E}\left[Y_{nT_n^o}^2\mid T_n^o\right]^{-1}\right]} = \frac{\frac{\pi_{n,t}^o}{\mu_{n,t}^2+\sigma_{n,t}^2}}{\sum_{v=0}^K \frac{\pi_{n,v}^o}{\mu_{n,v}^2+\sigma_{n,v}^2}},
&
T_n^\dagger&\sim \mathrm{Categorical}(\pi_n^\dagger),
\end{align*}
 then
\begin{align*}
\max_{\pi \in (\Delta^K)^N} \mathbb{E}[Z(\pi)]
=\sqrt{\sum_{n=1}^N \frac{\mathrm{Var}(\mu_{n,T^\dagger_n})}{\mathbb{E}[\sigma^2_{n,T^\dagger_n}] + \mathbb{E}[\mu_{n,T^\dagger_n}]^2}}.
\end{align*}
\end{repeatproposition}
Intuitively, each unit's contribution increases with the treatment effect for that unit and decreases with the variance in its treatment effect and with the magnitude of the mean outcomes.

\section{Integration to a Policy Optimization Pipeline}

Practitioners face two key challenges before they can use Theorem~\ref{thm:full_char} to propose a new policy. First, they do not know the true counterfactual outcome distribution and thus must form estimates of the counterfactual means $\hat{\mu}$ and variances $\hat{\sigma}^2$ to estimate $\hat{\Pi}_\Lambda$. Second, to avoid multiple testing concerns and streamline their decision process, practitioners need to select a single policy from $\hat{\Pi}_\Lambda$ to propose and evaluate on $\mathcal{O}$. We suggest practitioners use supervised machine learning algorithms on available training data to form counterfactual estimates and offer a method for selecting $\zeta$ to generate $\hat{\pi}^*(\zeta)$ in terms of the desired minimum $z$-score $Z_{\text{min}}$, and the desired improvement $\lambda$.

\subsection{Estimating the Frontier Using Machine Learning}

Forming $\Pi_{\Lambda}$ via Theorem~\ref{thm:full_char} requires knowledge about the observational dataset $\mathcal{O}$, namely the observational policy $\pi^o$ as well as the mean $\mu$ and variance $\sigma^2$ of the counterfactual outcomes. While practitioners often know $\pi^o$, the policy which generated $\mathcal{O}$, $\mu$ and $\sigma^2$ are fundamentally unknown when optimizing our policy. We suggest first learning estimates of the mean $\hat{\mu}$ and variance $\hat{\sigma}^2$ of the counterfactual outcomes in $\mathcal{O}$ using the practitioner's choice of machine learning algorithm and an available training dataset $\mathcal{T}$. Our approach can also be applied when $\pi^o$ is unknown by additionally estimating $\hat{\pi}^o$ from the training dataset.

To form the estimates $\hat{\mu}$ and $\hat{\sigma}^2$ we assume access to a training dataset $\mathcal{T} = \{(X_1,T_1,Y_1),\ldots,(X_M,T_M,Y_M)\}$ where each unit's treatment and outcome are accompanied by a set of individual covariates $X_m$. We assume that these covariates are similarly available in the observational dataset, expanding its definition to $\mathcal{O} = \{(\pi^o_1,X^o_1,T^o_1,Y^o_1),\ldots,(\pi^o_N,X^o_n,T^o_N,Y^o_N)\}$. The policy which generates the treatments for $\mathcal{T}$ need not be the same as $\pi^o$ unless the observational policy needs to be estimated from $\mathcal{T}$. However treatment assignment in $\mathcal{T}$ should follow standard causal inference assumptions such as positivity and selection on observables.

This training dataset can be used to form counterfactual outcome estimates $\hat{\mu}(x,t)$ and $\hat{\sigma}^2(x,t)$ in terms of a unit's covariates and any treatment. These estimates can be formed by running any supervised machine learning algorithm or regression $f$ on $\mathcal{T}$. If counterfactual variance estimates are difficult to extract from $f$ directly, they can be estimated from the squared residuals of $\mathcal{T}$, $\{(X_1,T_1,(Y_1 - \hat{\mu}(X_1,T_1))^2),\ldots,(X_M,T_M,(Y_M - \hat{\mu}(X_M,T_M))^2)\}$. In our simulations, we choose $f$ to be a set of $K$ honest regression forests, one for each treatment \citep{athey_generalized_2019}. This approach allows for $f$ to form unbiased estimates non-parametrically when the covariate space is rich, without allowing cross-treatment pooling. Moreover, bag-of-bootstrap approaches allow forests to provide variance estimates directly, eliminating the need for prediction of a separate residual model \citep{Lu_varPred}.

Once the counterfactual mean $\hat{\mu}$ and variance $\hat{\sigma}^2$ estimators are formed, we plug them into Theorem~\ref{thm:full_char} with $\mu_{n,t} \leftarrow \hat{\mu}(X^o_n,t)$ and $\sigma^2_{n,t} \leftarrow \hat{\sigma}^2(X^o_n,t)$ to generate an estimate of $\hat{\Pi}_\Lambda$, which contains the Pareto frontier for the counterfactual outcome distribution described by the estimators $\hat{\mu}$ and $\hat{\sigma}^2$. It almost certainly underperforms the true frontier of policies related to $\mu$ and $\sigma^2$. We note that as long as $\hat{\mu}$ and $\hat{\sigma}^2$ are reasonably close to the true parameters $\mu$ and $\sigma$, $\hat{\Pi}_\Lambda$ will extend from the observational policy in a similar direction as $\Pi_{\Lambda}$, ensuring similar policy suggestions for low values of $\zeta$. 

The policies trained and evaluated by this pipeline depend on both the training dataset $\mathcal{T}$ and the observed test covariates $X^o$. A relevant concern is whether the IPW estimator still provides an unbiased estimate of a policies performance given this information. Under the assumptions of the Rubin Causal model we show that the IPW estimator is still unbiased for policies which depend on a training set and test covariates, $\pi(\mathcal{T},X^o)$.
\begin{repeatproposition}
\label{prop:unbiased_est}
$$\mathbb{E}\left[\hat{\tau}(\pi(\mathcal{T},X^o))|\mathcal{T},X^o\right] = \tau(\pi(\mathcal{T},X^o))$$
\end{repeatproposition}
Thus, while the learned policy itself may vary if the training dataset or test covariates change, IPW still provides an unbiased estimate of the learned policy on the observational dataset.

\subsection{Selecting a Policy from the Optimal Frontier}
\label{sec:select}

In addition to generating an estimated frontier, practitioners need to select a policy from the frontier to implement. This problem can be reduced to choosing the value of $\zeta$ which generates $\hat{\pi}^*(\zeta) \in \hat{\Pi}_{\Lambda}$. This can be chosen via guess and check or some form of cross-validation, but this still requires some general idea of $\zeta$'s magnitude and may lead to inaccurate evaluation due to concerns with multiple hypothesis testing \citep{shaffer_multiple_1995}. Instead, we propose to set $\zeta$ using the economic intuition its has as a dual variable.

In the optimization problems defining $\Pi_{\Lambda}$, $\zeta$ is the dual variable associated with the constraint $\tau(\pi) = \lambda$. Thus, $\zeta$ represents the marginal decrease that the squared standard error $s^2(\pi)$ can attain if $\lambda$ is decreased. By selecting $\zeta$, we set a limit on how much any policy change can increase the squared standard error in terms of its expected improvement. All expected policy improvements which can be made while introducing less standard error then this limit are included in $\pi^*(\zeta)$. This intuition offers a conservative way to set $\zeta$ for a practitioner hoping to achieve an improvement of $\lambda$ with a $z$-score of at least $Z_{\text{min}}$. From the proof of Proposition~\ref{prop:z_deriv}, we have that if $\tau(\pi^*(\zeta)) = \lambda$, then $s^2(\pi^*(\zeta)) \leq \frac{\zeta}{2} \lambda$ due to the limit set by the dual variable $\zeta$ and the concavity of $\tau(\pi^*(\zeta))$ in $\zeta$. As a result, we have
$$\mathbb{E}[Z(\pi^*(\zeta))] = \frac{\tau(\pi^*(\zeta))}{s(\pi^*(\zeta))} \geq \sqrt{\frac{2\lambda}{\zeta}}.$$
Thus, we can ensure $\mathbb{E}[Z(\pi^*(\zeta))] \geq Z_{\text{min}}$ by setting
$$\zeta < \frac{2\lambda}{Z_{\text{min}}^2}.$$
Lower $\zeta$ will raise the probability of achieving a significant result, but lower the expected improvement.
While the limit on $s^2(\pi^*(\zeta))$ is not tight if $\pi^*(\zeta) \in \Pi_{\text{Opt}}$, we recommend practitioners use this bound as our lack of knowledge with respect to the true counterfactual outcome distribution only allows construction of the inferior $\hat{\pi}^*(\zeta)$. Using this value of $\zeta$ ensures that $\hat{\pi}^*(\zeta)$ only attempts to make improvements that introduce allowable variance for the given significance level, limiting the concern that noise will drown out significant policy gains. 

\section{Simulations}

We perform simulations exploring our method's ability to extract policy improvements when na\"{i}ve techniques fail. We apply our method to data from a megastudy encouraging flu vaccination using text message reminders \citep{milkman_680000-person_2022} and find a personalized policy that increases the vaccination rate by 4\% while the na\"{i}ve approach fails to extract a significantly improved policy (Section~\ref{sec:vaccine}). Then, we simulate data based on the refugee relocation study of \cite{bansak_improving_2018}, showing our method can be effectively utilized in settings with additional policy constraints (Section~\ref{sec:refugee_sim}). In Appendix~\ref{sec:stylized} we provide additional simulations in a stylized environment in to explore how the accuracy of the prediction model given to our method impacts the performance of learned policies and compare our method to other policy optimization techniques.

\subsection{Case Study: Vaccine Nudge Megastudy}
\label{sec:vaccine}

Behavioral health interventions, such as text message reminders, offer a nearly costless mechanism for practitioners to improve health outcomes such as vaccination rates \citep{chapman_opting_2010,milkman_using_2011,regan_randomized_2017}---many people intend to get vaccinated, but fail to follow through \citep{sheeran_intentionbehavior_2002}. Effective reminders can motivate individuals to act on their intentions, improving vaccination rates with minimal cost. We demonstrate that IAPO can be used to learn provably effective policies for targeting these kinds of nudges. We use data from \cite{milkman_680000-person_2022}, who performed a megastudy (i.e., a large-scale randomized control trial) testing the efficacy of 22 randomly chosen text messages and a control condition at encouraging 680,000 patients at a large US retail pharmacy to receive the flu vaccine during the 2019-2020 flu season.

The data includes the vaccination outcome of each patient (we use vaccinations before October 31, 2019) and their treatment assignment. In addition, it includes patient covariates related to their locations and health records, including (i) age, (ii) gender, (iii) zip code demographics, (iv) past vaccinations, and (v) hypertension. These covariates offer the opportunity to target the choice of text messages---different patients may respond to different motivations, offering the potential for large gains from a personalized treatment. We split the study subjects into training and test samples, train a set of honest random forests to predict each treatment's counterfactual outcomes, use the forests to estimate the Pareto frontier of policies via IAPO, then estimate their performance on the test samples using IPW. After removing samples with missing covariates we maintain 474,667 samples for training and 203,429 samples for testing. In total 25.6\% of the patients in our (combined training and test) sample received a vaccination by our cutoff.

\begin{figure}
\centering
\begin{subfigure}[b]{0.49\textwidth}
\centering
\includegraphics[width=\textwidth]{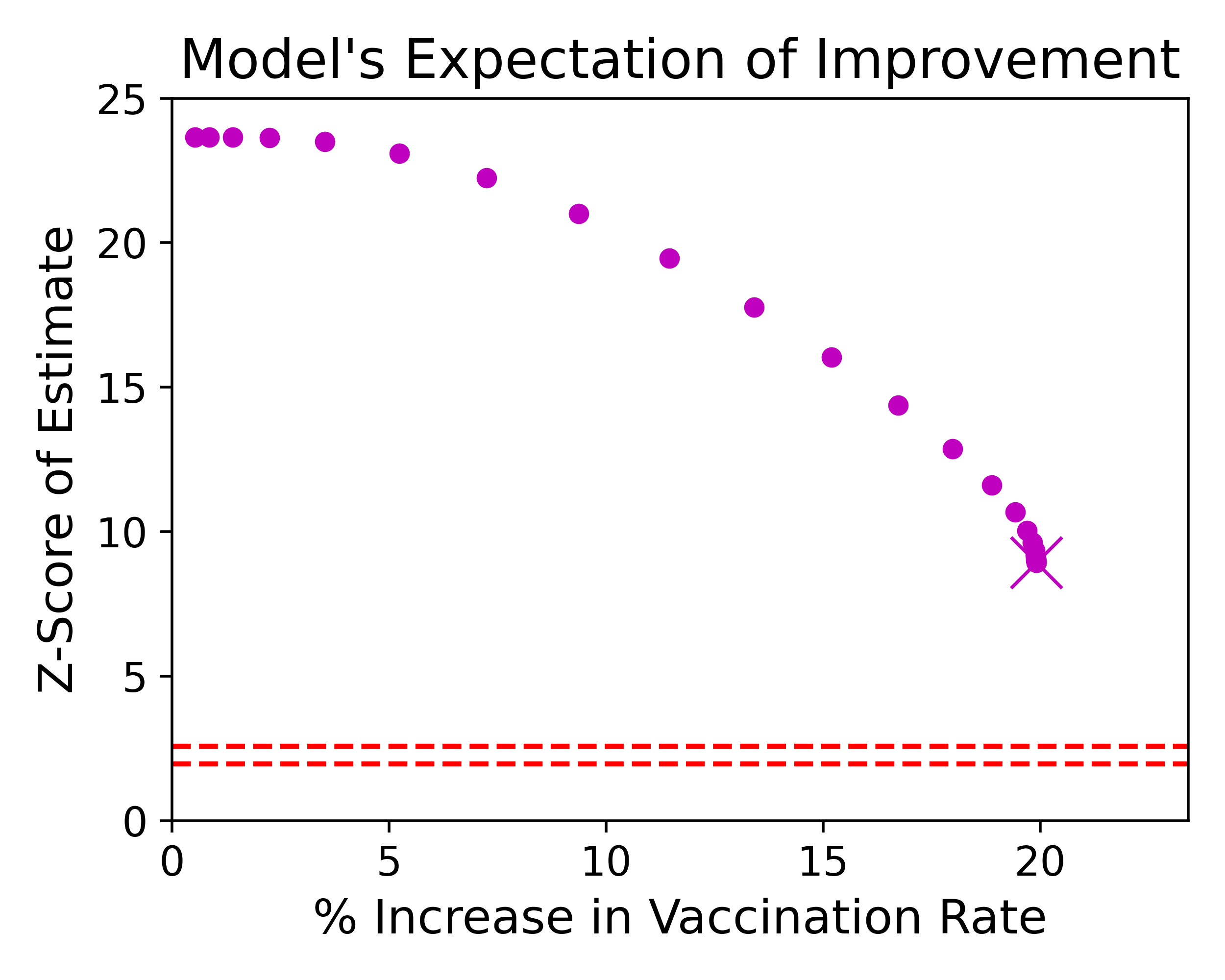}
\label{fig:Wal_full}
\end{subfigure}
\hfill
\begin{subfigure}[b]{0.49\textwidth}
\centering
\includegraphics[width=\textwidth]{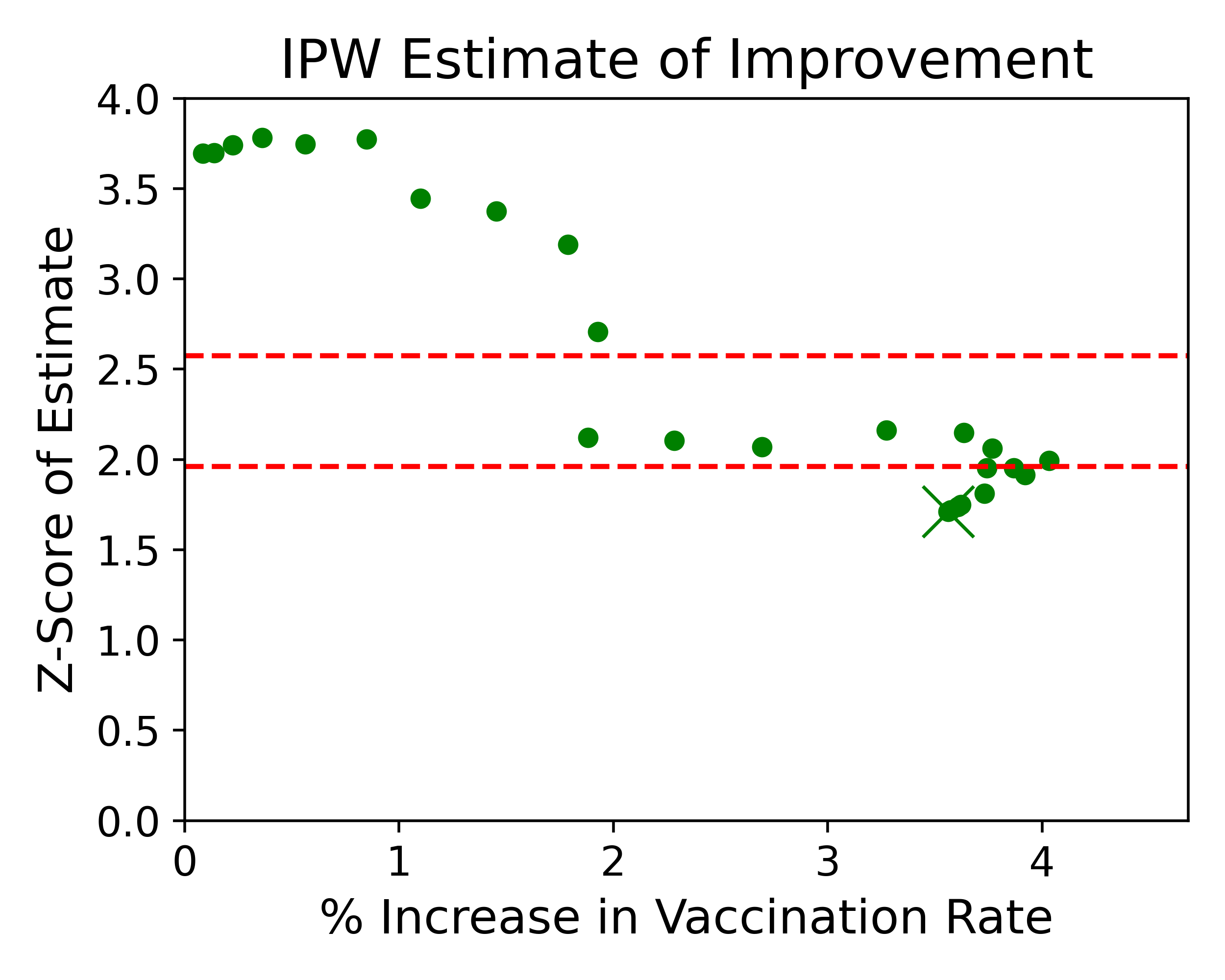}
\label{fig:Wal_zoom}
\end{subfigure} \\
\includegraphics[width = 0.8\textwidth]{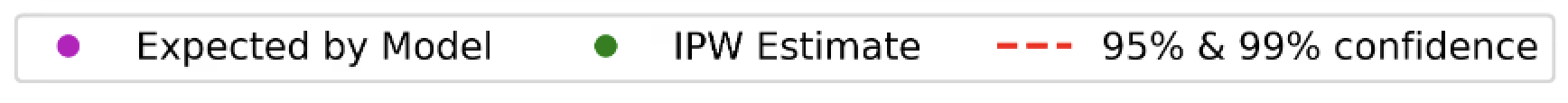}
\caption{The model believes policies on the estimated frontier can all achieve highly significant improvements; however a deterministic assignment of treatments following optimization of the model (marked by an `x') does not achieve a confident improvement. IAPO still manages to extract significant policy improvements.}
\label{fig:Wal_case}
\end{figure}

Figure~\ref{fig:Wal_case} shows both the performance improvement the honest random forest model believes that policies on the estimated frontier can achieve (left) and the evaluation of those policies by IPW (right). It believes that na\"{i}vely choosing the ``winning'' treatment for each patient should yield statistically significant 20\% increase in vaccination rate. However, when properly evaluated using IPW, this policy yields an improvement of 3.5\%, and is not statistically significantly better. In contrast, IAPO learns a policy with a similar 3.5\% increase in vaccination rate, which is statistically significantly better than the baseline policy at the 95\% confidence level. Note that IAPO also suffers from the winner's curse, since it is highly optimistic about performance along the frontier; however, it provides the regularization needed to obtain improvements that hold up under rigorous evaluation.

Table~\ref{tab:wal_policies} describes how policies learned by IAPO achieve significant improvements when na\"{i}ve optimization of our model fails. We focus on two policies from the frontier; 95\% confident refers to a policy which achieves a similar improvement estimate to the na\"{i}ve approach, but passes a significance test with 95\% confidence while 99\% confident refers to a policy which achieves half the improvement but passes a significance test with 99\% confidence. All policies essentially eliminate the control treatment, but are highly personalized, with the most frequently used treatment given to less than 9\% of patients. About half of the units are assigned a deterministic treatment by the 95\% confident policy, while the 99\% confident policy use stochastic treatments assignments for all units. This increases the overlap of both policies compared to the na\"{i}ve policy, helping them achieve lower standard errors on their improvement estimates and allowing them to pass significance tests at higher confidence levels. The 95\% confident policy shows that the na\"{i}ve approach can be improved while only making half of units receive nondeterministic treatments while the 99\% confident policy shows that much larger gains in significance require much greater overlap with the observational policy, assigning no treatments deterministically.

\begin{table}[]
\centering
\begin{tabular}{ccccc}
\toprule
Policy & Frequency (Min , Max) & Active (Min, Mean, Max) & Deterministic & Overlap \\
\midrule
Na\"{i}ve &  (0.005 , 0.083) & (1, 1, 1) & 1.0 & 0.045  \\
95\% Confident &  (0.005 , 0.082) & (1, 1.71, 7) & 0.466 & 0.074  \\
99\% Confident &  (0.009 , 0.077) & (2, 9.39, 19) & 0.0 & 0.361 \\
\bottomrule
\end{tabular}
\vspace{5pt}
\caption{Illustration of how IAPO policies differ from na\"{i}ve ones. ``Frequency'' is the range of how frequently each treatment is assigned. ``Active'' is the number of non-zero treatment propensities for each unit on average. ``Deterministic'' is the fraction of units receiving a deterministic treatment assignment. ``Overlap'' is the similarity to the observational policy, reported as $1-TV(\pi,\pi^o)$ where $TV$ is the total variation distance.}
\label{tab:wal_policies}
\end{table}

\subsection{Case Study: Refugee Relocation Simulation}
\label{sec:refugee_sim}

In the United States, refugee resettlement agencies are responsible for choosing where to settle incoming refugees. Their goal is to choose an assignment policy $\pi$ that places refugees in available locations in a welfare maximizing way, often measured by outcomes such as achieving employment within one year of settling. Assignment policies must not overload locations' capacities (due to limited resources available to settle refugees at each location). We perform a simulation study to evaluate the efficacy of IAPO at policy learning when varying the best possible performance gains from personalization.

Specifically, we simulate a population of refugees matching the descriptive statistics of the population studied by \cite{bansak_improving_2018}. We generate 33,782 samples with characteristics of (i) location assigned, (ii) case status (free/fixed assignment), (iii) age, (iv) nationality, (v) education level, and (vi) arrival date. Each sample was assigned characteristics independently at random with probabilities matching the descriptive statistics given in Table S1 of \cite{bansak_improving_2018}. Age and nationality information were reported at a coarsened level, so we drew those characteristics uniformly at random from coarsened buckets. We refer to characteristics (iii)-(vi) of sample $n$ as $X_n$ and index different characteristics using subscript $j$. We use case status to determine whether samples are used for model training (24,327 samples) or testing (9,455 samples).

We simulate an employment outcome for each sample $n$ independently according to a bernoulli distribution whose mean $\mu_{n,t}$ depends on the sample's characteristics $X_n$ and assigned location $t$. We standardize the characteristics $X_n$ to $X_n^\dagger$ by subtracting the average of each characteristic and dividing by that characteristic's standard deviation across all samples. Then, we generate a fixed causal effect for each covariate $\alpha_j$ and each covariate-location interaction $\beta^{(t)}_j$; these interactions are what enable gains from personalization. We determine $\mu_{n,t}$ using a logistic function whose input is determined by a linear interaction of $\alpha$ and $\beta^{(t)}$ with $X_n^\dagger$:
\begin{align*}
Y_{n,t}&\sim \mathrm{Ber}(\mu_{n,t})
&
\mu_{n,t}&= \frac{1}{1+e^{-X^\dagger_{n} \left(\alpha + \beta^{(t)} \right) + c_{tot}}}
&
\frac{\alpha_j}{c_{unit}}&\sim \mathrm{Unif}(-1,1)
&
\frac{\beta^{(t)}_j}{c_{intx}}&\sim  \mathrm{Unif}(-1,1)
\end{align*}
The parameters $c_{unit}$ and $c_{intx}$ allow us to control the magnitude of the causal effects we introduce while $c_{tot}$ allows us to change the average employment outcome. Increasing $c_{unit}$ increases the heterogeneity in refugees' baseline employment probabilities independent of location while increasing $c_{intx}$ increases the heterogeneity in the impact each location has on each refugee's employment probability. Table~\ref{tab:Refugee_Data} gives the parameters used to generate 3 datasets with varying magnitudes of covariate-location interaction effects. For each, we measure cross-unit heterogeneity as the standard deviation across samples employment probability (averaged over locations) and within-unit heterogeneity as the average standard deviation of samples employment probability across locations:
\begin{align*}
\text{Cross-Unit Het.} &= \sqrt{\frac{1}{N}\sum_{n=1}^N \left(\overline{\mu}_n - \overline{\mu}\right)^2}
&
\text{Within-Unit Het.} &= \frac{1}{N}\sum_{n=1}^N \sqrt{\frac{1}{K}\sum_{t=1}^K \left(\mu_{n,t} - \overline{\mu}_n\right)^2}\\
\overline{\mu}&= \frac{1}{N} \sum_{n=1}^N \overline{\mu}_n
&
\overline{\mu}_n &= \frac{1}{K} \sum_{t=1}^K \mu_{n,t}
\end{align*}
Our simulated refugee datasets maintain a consistent average outcome and cross-unit heterogeneity while varying the impact locations can have on a refugee's employment.

\begin{table}
\centering
\begin{tabular}{ccccccc}
\toprule
Interaction & $c_{tot}$ & $c_{unit}$ & $c_{intx}$ & Employed & Cross-Unit Het. & Within-Unit Het. \\
\midrule
None & 0.86  & 0.25 & 0.0 & 0.316 & 0.136 & 0.0 \\
Small & 0.86 & 0.25 & 0.075 & 0.318 & 0.135 & 0.037 \\
Medium & 0.86 & 0.25 & 0.15 & 0.323 & 0.132 & 0.074 \\
\bottomrule
\end{tabular}
\vspace{5pt}
\caption{Parameters used to generate eachof our three refugee simulation datasets with varying magnitudes of covariate-location interaction effects.}
\label{tab:Refugee_Data}
\end{table}

One challenge applying IAPO to this setting is the presence of resource constraints that limit the set of feasible refugee assignment policies. Each location $t$ can accept a maximum number of refugees $d_t$ given their available resources. We consider a policy feasible if, in expectation, it assigns fewer than $d_t$ refugees to location $t$, as represented by the constraint
$$\sum_{n=1}^N \pi_{n,t} \leq d_t \ \forall \ t.$$
Meeting the assignment constraints themselves can then be included as part of the randomization process or can be ignored if the constraints are approximate or somewhat flexible. For our simulations, we set each locations capacity $d_t$ as $5/4$ of the number of freely assigned refugees (our test samples) settled in location $t$. This choice allows proposed policies to slightly increase the number of refugees sent to any given location while requiring the majority of gains to come from identifying refugee-location synergy.

We integrate resource constraints into the IAPO pipeline as linear constraints to the convex optimization problems defining $\Pi_{\Lambda}$. Let $D =\left \{\pi\in(\Delta^K)^N\mid \sum_{n=1}^N \pi_n \leq d \right\}$ be the set of feasible policies according to our capacities $d$. Then, our Pareto frontier is
$$\Pi_{\text{Opt}} =\left\{ \pi \in D\biggm\vert \not\exists \ \pi'\ \in D \text{ s.t. } \tau(\pi') \geq\tau(\pi)\wedge\frac{\tau(\pi')}{s(\pi')} \geq \frac{\tau(\pi)}{s(\pi)}\wedge(\tau(\pi'),s(\pi')) \neq (\tau(\pi),s(\pi))\right\},$$
which can be relaxed to the family of convex optimization problems given by
$$\Pi_{\Lambda} = \bigcup_{\lambda \in [0,\tau_{\text{max}}]} \operatorname*{\arg\min}_{\{\pi \in (\Delta^K)^N\mid\tau(\pi) = \lambda, \sum_{n=1}^N \pi_n \leq d\}} s^2(\pi).$$
Any policy constraint which can be represented as a valid convex programming constraint in terms of the policy's propensities can be incorporated into IAPO in the same way as resource constraints.

Instead of characterizing the resource constrained $\Pi_{\Lambda}$, we approximate it using convex optimization software. Specifically, we implement the convex program defining $\Pi_{\Lambda}$ in CVXPY and solve it using the Splitting Conic Solver (SCS) while varying the value of $\lambda$ to solve for $\Pi_{\Lambda}$ point-wise. We use a convergence tolerance of $10^{-6}$ to achieve precise results; however, as our objective approaches zero for policies close to the observational policy, we only solve for policies with sufficiently high expected improvement $\lambda$ to ensure stability of the solver's estimate.

\begin{figure}
\centering
\begin{subfigure}[b]{0.32\textwidth}
\centering
\includegraphics[width=\textwidth]{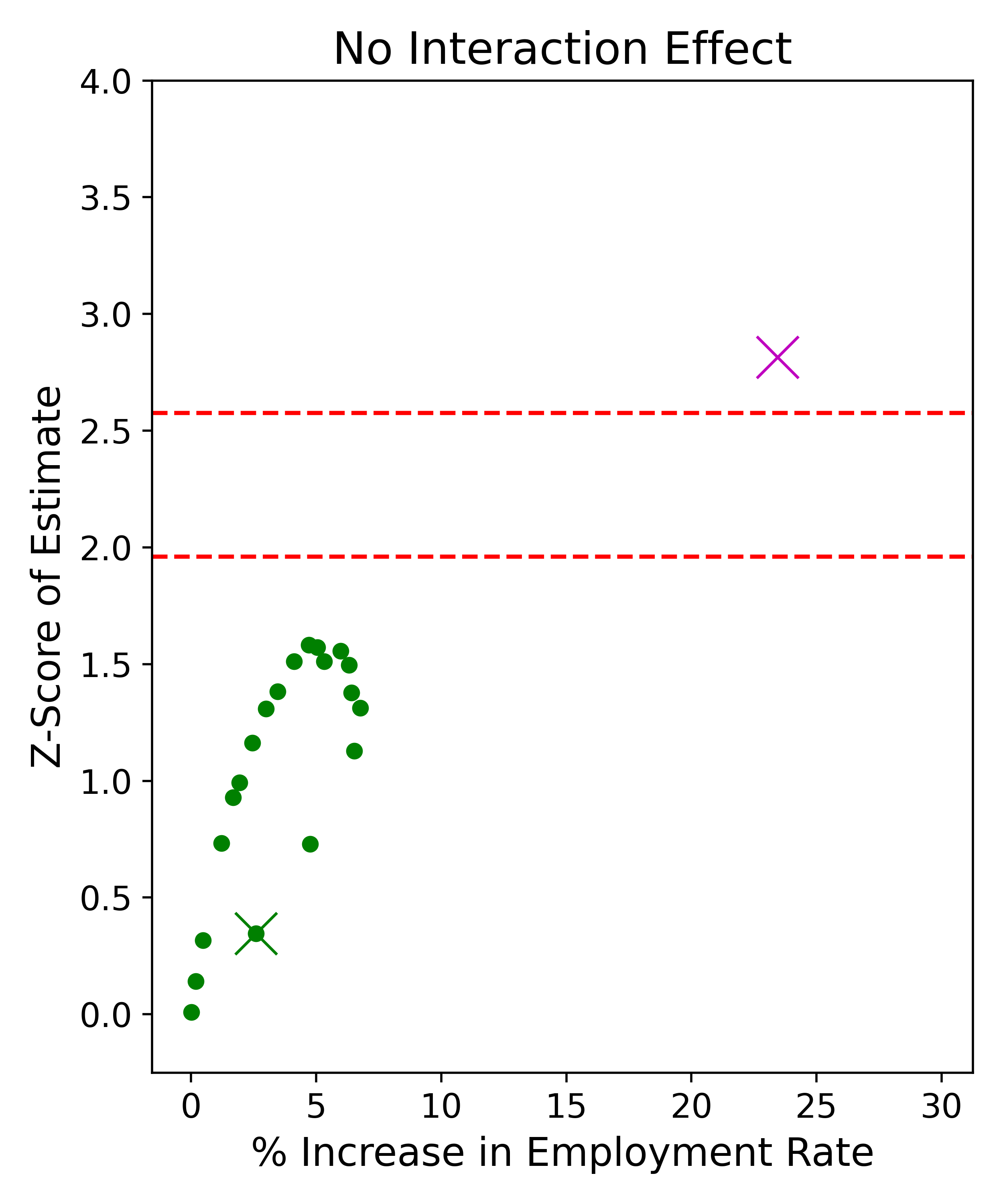}
\label{fig:nointx}
\end{subfigure}
\hfill
\begin{subfigure}[b]{0.32\textwidth}
\centering
\includegraphics[width=\textwidth]{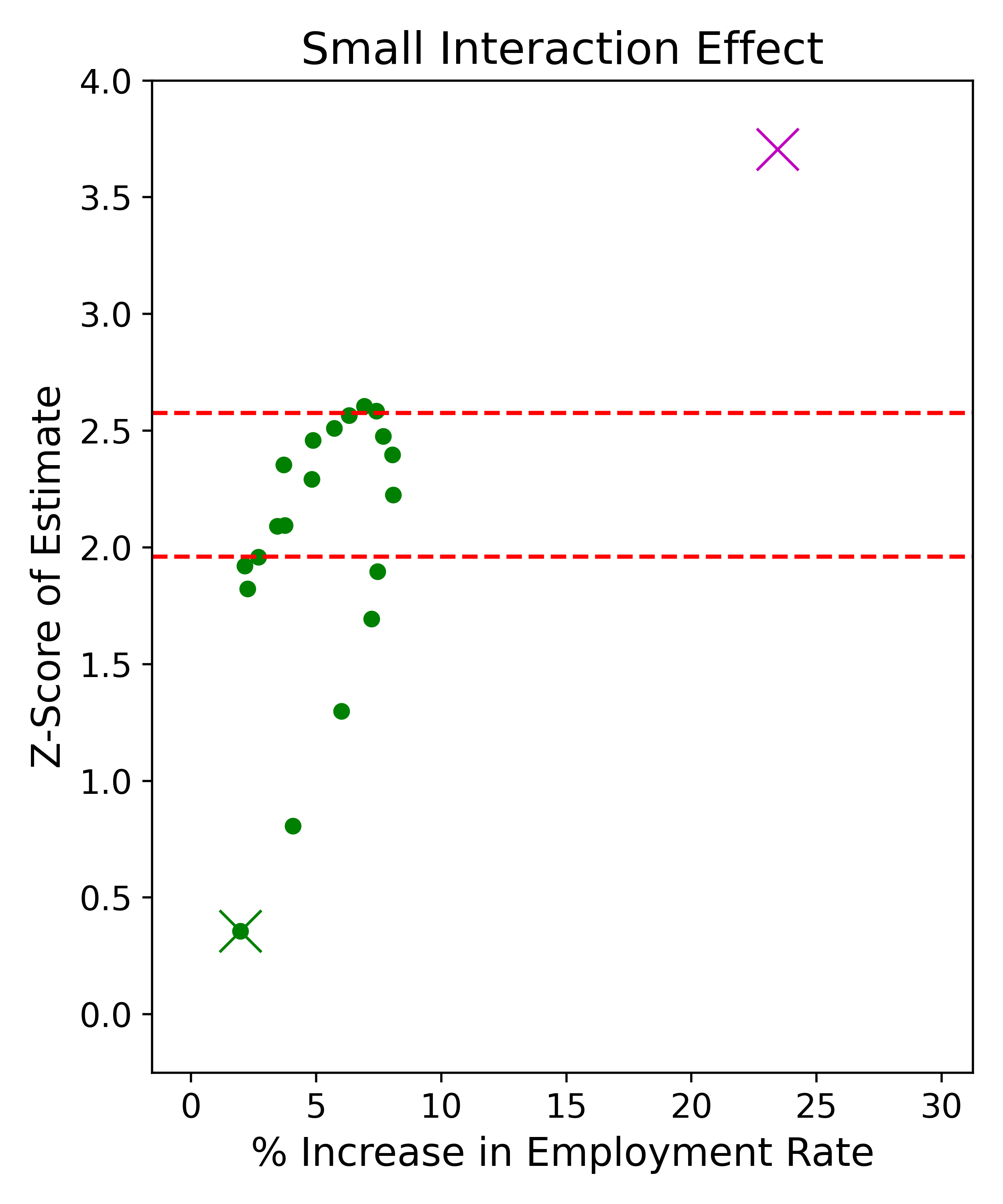}
\label{fig:smallintx}
\end{subfigure}
\hfill
\begin{subfigure}[b]{0.32\textwidth}
\centering
\includegraphics[width=\textwidth]{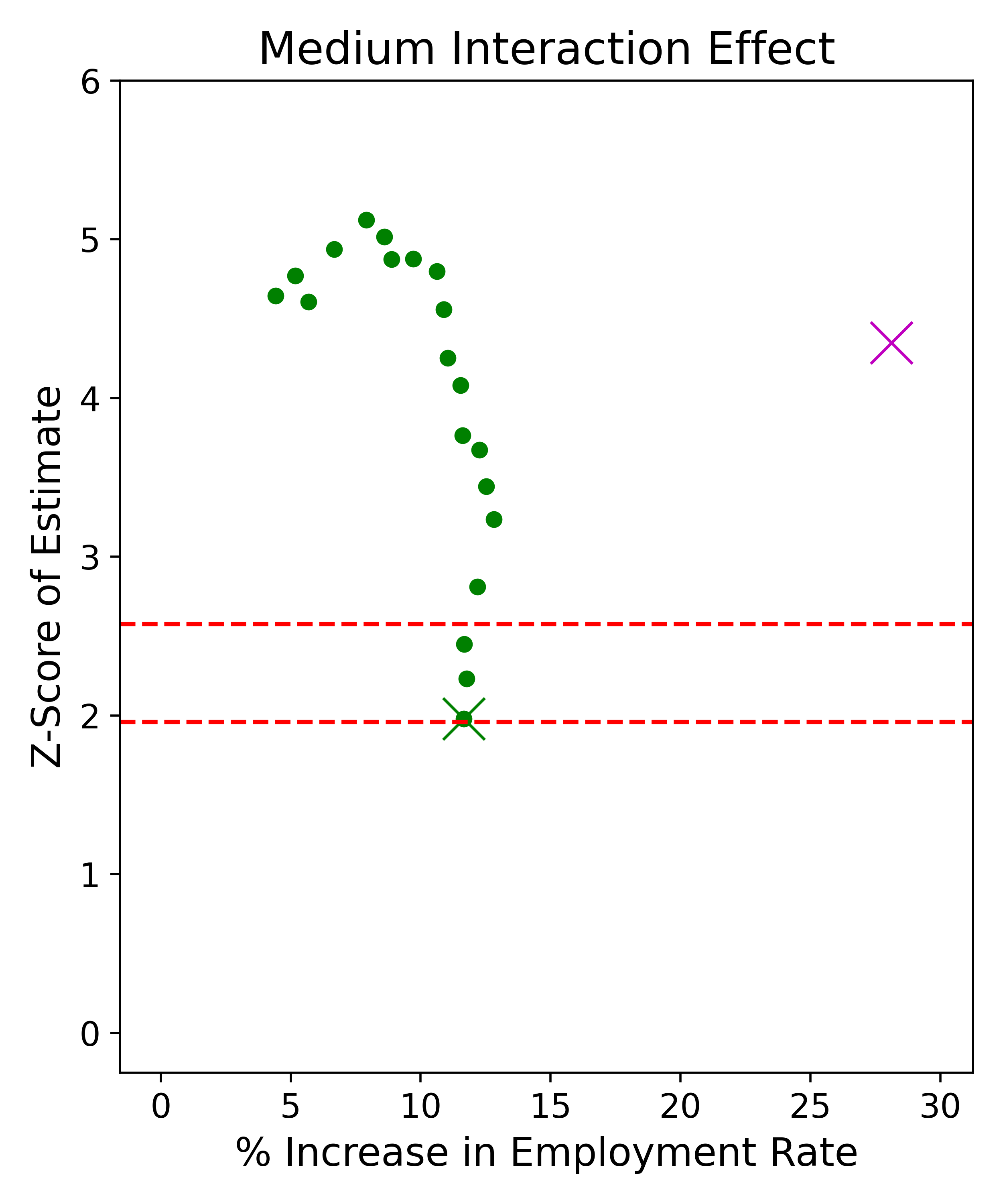}
\label{fig:medintx}
\end{subfigure}\\
\includegraphics[width=0.8\textwidth]{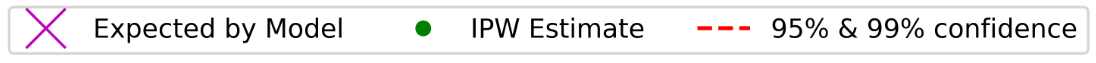}
\caption{Pareto frontier of policies computed using IAPO for each of our three refugee simulation datasets according to IPW estimation on a held out test set. The purple ``x'' marks the predicted performance of the expectation-maximizing policy, and the green ``x'' marks its actual performance.}
\label{fig:refugee}
\end{figure}

We train a set of honest random forest regressions on each of our three refugee datasets to estimate the Pareto frontier of policies for each one. Figure~\ref{fig:refugee} reports IPW performance estimates using held-out test sets, as well as the predicted and actual performance of the corresponding expectation-maximizing policies. For all three datasets, the winner's curse causes the model to believe it can obtain massive employment gains regardless of the actual potential to improve employment. In contrast, when useful within-unit heterogeneity exists, IAPO obtains gains with much higher confidence and at least as much improvement as the na\"{i}ve approach. In more detail, when policy improvement is impossible (``No Interaction Effect''), no policy along the frontier obtains significant policy improvement, which is expected. For the ``Small Interaction Effect'' dataset, the na\"{i}ve approach fails to obtain statistically significant policy improvement, whereas IAPO obtains policy improvements up to the 99\% confidence level while maintaining similar gains as the na\"{i}ve approach. For the ``Medium Interaction Effect'' dataset, the na\"{i}ve approach just barely obtains significant improvements at the 95\% confidence level; in contrast, IAPO readily obtains similar improvements at a much higher 99\% confidence level.

\section{Conclusion}

We have proposed a novel policy optimization strategy that aims to identify the full Pareto frontier of policies trading off expected performance with provably significant performance improvements. Focusing on the specific setting where policy evaluation is performed using IPW, we have mathematically characterized the Pareto frontier.  Our characterization reveals that focusing on statistical significance prioritizes elimination of bad treatment assignments over selecting the best treatment assignments. Intuitively, rather than try and identify optimal treatments, a decision-maker focuses on the problem of eliminating bad treatments, which can be simpler while yielding confident improvements.
Based on our theoretical analysis, we devise a policy optimization algorithm using a plug-in approach. Practitioners first use their choice of supervised machine learning algorithm to train a counterfactual prediction model on the observational data, and then use its predictions to compute the Pareto frontier of optimal policies.
Our simulations show that our plug-in approach can extract significant policy improvements, while other common policy optimization approaches cannot, especially when the underlying causal model is complex.

We leave a number of important directions for future work. First, while we have performed some preliminary analysis of how prediction errors affect estimation of the Pareto frontier, significantly more work is required to understand the impact. Our work also assumes the importance weights are known, whereas they may need to be estimated; understanding how IPW estimation error affects our procedure is another important direction.
One key benefit of our convex formulation is that additional constraints (e.g., budget constraints) can be incorporated; it may be interesting to explore the impact of these constraints on the characterization of the Pareto frontier. Finally, we have focused on IPW; exploring how to perform policy optimization in conjunction with other policy evaluation methodologies (e.g., randomized controlled trials, instrumental variables, etc.) deserves similar attention.

\bibliography{refs}

\begin{thebibliography}{58}
\providecommand{\natexlab}[1]{#1}
\providecommand{\url}[1]{\texttt{#1}}
\providecommand{\urlprefix}{URL }

\bibitem[{Ahani et~al.(2021)Ahani, Andersson, Martinello, Teytelboym, \protect\BIBand{} Trapp}]{ahani_placement_2021}
Ahani N, Andersson T, Martinello A, Teytelboym A, Trapp AC (2021) Placement {Optimization} in {Refugee} {Resettlement}. \emph{Operations Research} 69(5):1468--1486, ISSN 0030-364X, 1526-5463, \urlprefix\url{http://dx.doi.org/10.1287/opre.2020.2093}.

\bibitem[{Andrews et~al.(2024)Andrews, Kitagawa, \protect\BIBand{} McCloskey}]{andrews_inference_2024}
Andrews I, Kitagawa T, McCloskey A (2024) Inference on {Winners}. \emph{The Quarterly Journal of Economics} 139(1):305--358, ISSN 0033-5533, 1531-4650, \urlprefix\url{http://dx.doi.org/10.1093/qje/qjad043}.

\bibitem[{Athey et~al.(2019)Athey, Tibshirani, \protect\BIBand{} Wager}]{athey_generalized_2019}
Athey S, Tibshirani J, Wager S (2019) Generalized random forests. \emph{The Annals of Statistics} 47(2), ISSN 0090-5364, \urlprefix\url{http://dx.doi.org/10.1214/18-AOS1709}.

\bibitem[{Athey \protect\BIBand{} Wager(2021)}]{athey_policy_2021}
Athey S, Wager S (2021) Policy {Learning} {With} {Observational} {Data}. \emph{Econometrica} 89(1):133--161, ISSN 0012-9682, \urlprefix\url{http://dx.doi.org/10.3982/ECTA15732}.

\bibitem[{Banerjee et~al.(2025)Banerjee, Chandrasekhar, Dalpath, Duflo, Floretta, Jackson, Kannan, Loza, Sankar, Schrimpf, \protect\BIBand{} Shrestha}]{banerjee_selecting_2025}
Banerjee A, Chandrasekhar AG, Dalpath S, Duflo E, Floretta J, Jackson MO, Kannan H, Loza F, Sankar A, Schrimpf A, Shrestha M (2025) Selecting the {Most} {Effective} {Nudge}: {Evidence} {From} a {Large}‐{Scale} {Experiment} on {Immunization}. \emph{Econometrica} 93(4):1183--1223, ISSN 0012-9682, \urlprefix\url{http://dx.doi.org/10.3982/ECTA19739}.

\bibitem[{Bang \protect\BIBand{} Robins(2005)}]{bang2005doubly}
Bang H, Robins JM (2005) Doubly robust estimation in missing data and causal inference models. \emph{Biometrics} 61(4):962--973.

\bibitem[{Bansak et~al.(2018)Bansak, Ferwerda, Hainmueller, Dillon, Hangartner, Lawrence, \protect\BIBand{} Weinstein}]{bansak_improving_2018}
Bansak K, Ferwerda J, Hainmueller J, Dillon A, Hangartner D, Lawrence D, Weinstein J (2018) Improving refugee integration through data-driven algorithmic assignment. \emph{Science} 359(6373):325--329, ISSN 0036-8075, 1095-9203, \urlprefix\url{http://dx.doi.org/10.1126/science.aao4408}.

\bibitem[{Beygelzimer \protect\BIBand{} Langford(2008)}]{beygelzimer_offset_2008}
Beygelzimer A, Langford J (2008) The {Offset} {Tree} for {Learning} with {Partial} {Labels}. \urlprefix\url{http://dx.doi.org/10.48550/ARXIV.0812.4044}.

\bibitem[{Bottou et~al.(2013)Bottou, Peters, Qui{\~n}onero-Candela, Charles, Chickering, Portugaly, Ray, Simard, \protect\BIBand{} Snelson}]{bottou2013counterfactual}
Bottou L, Peters J, Qui{\~n}onero-Candela J, Charles DX, Chickering DM, Portugaly E, Ray D, Simard P, Snelson E (2013) Counterfactual reasoning and learning systems: The example of computational advertising. \emph{The Journal of Machine Learning Research} 14(1):3207--3260.

\bibitem[{Boutilier et~al.(2022)Boutilier, Jónasson, \protect\BIBand{} Yoeli}]{boutilier_improving_2022}
Boutilier JJ, Jónasson JO, Yoeli E (2022) Improving {Tuberculosis} {Treatment} {Adherence} {Support}: {The} {Case} for {Targeted} {Behavioral} {Interventions}. \emph{Manufacturing \& Service Operations Management} 24(6):2925--2943, ISSN 1523-4614, 1526-5498, \urlprefix\url{http://dx.doi.org/10.1287/msom.2021.1046}.

\bibitem[{Capen et~al.(1971)Capen, Clapp, \protect\BIBand{} Campbell}]{capen1971competitive}
Capen EC, Clapp RV, Campbell WM (1971) Competitive bidding in high-risk situations. \emph{Journal of petroleum technology} 23(06):641--653.

\bibitem[{Chapman et~al.(2010)Chapman, Li, Colby, \protect\BIBand{} Yoon}]{chapman_opting_2010}
Chapman GB, Li M, Colby H, Yoon H (2010) Opting {In} vs {Opting} {Out} of {Influenza} {Vaccination}. \emph{JAMA} 304(1):43, ISSN 0098-7484, \urlprefix\url{http://dx.doi.org/10.1001/jama.2010.892}.

\bibitem[{Chernozhukov et~al.(2025)Chernozhukov, Lee, Rosen, \protect\BIBand{} Sun}]{chernozhukov_policy_2025}
Chernozhukov V, Lee S, Rosen AM, Sun L (2025) Policy {Learning} with {Confidence}. \urlprefix\url{http://dx.doi.org/10.48550/arXiv.2502.10653}, arXiv:2502.10653.

\bibitem[{Dud{\'\i}k et~al.(2011)Dud{\'\i}k, Langford, \protect\BIBand{} Li}]{dudik2011doubly}
Dud{\'\i}k M, Langford J, Li L (2011) Doubly robust policy evaluation and learning. \emph{Proceedings of the 28th International Conference on International Conference on Machine Learning}, 1097--1104.

\bibitem[{Ernst et~al.(2005)Ernst, Geurts, \protect\BIBand{} Wehenkel}]{ernst2005tree}
Ernst D, Geurts P, Wehenkel L (2005) Tree-based batch mode reinforcement learning. \emph{Journal of Machine Learning Research} 6.

\bibitem[{Gupta et~al.(2024)Gupta, Huang, \protect\BIBand{} Rusmevichientong}]{gupta2024debiasing}
Gupta V, Huang M, Rusmevichientong P (2024) Debiasing in-sample policy performance for small-data, large-scale optimization. \emph{Operations Research} 72(2):848--870.

\bibitem[{Harrison \protect\BIBand{} March(1984)}]{harrison1984decision}
Harrison JR, March JG (1984) Decision making and postdecision surprises. \emph{Administrative Science Quarterly} 26--42.

\bibitem[{Kallus(2018)}]{kallus_balanced_2018}
Kallus N (2018) Balanced {Policy} {Evaluation} and {Learning}. \emph{Advances in {Neural} {Information} {Processing} {Systems}}, volume~31 (Curran Associates, Inc.), \urlprefix\url{https://proceedings.neurips.cc/paper/2018/hash/6616758da438b02b8d360ad83a5b3d77-Abstract.html}.

\bibitem[{Kallus \protect\BIBand{} Zhou(2018)}]{Kallus_Confounding}
Kallus N, Zhou A (2018) Confounding-robust policy improvement. Bengio S, Wallach H, Larochelle H, Grauman K, Cesa-Bianchi N, Garnett R, eds., \emph{Advances in Neural Information Processing Systems}, volume~31 (Curran Associates, Inc.), \urlprefix\url{https://proceedings.neurips.cc/paper_files/paper/2018/file/3a09a524440d44d7f19870070a5ad42f-Paper.pdf}.

\bibitem[{Kitagawa \protect\BIBand{} Tetenov(2018)}]{kitagawa_who_2018}
Kitagawa T, Tetenov A (2018) Who {Should} {Be} {Treated}? {Empirical} {Welfare} {Maximization} {Methods} for {Treatment} {Choice}. \emph{Econometrica} 86(2):591--616, ISSN 0012-9682, \urlprefix\url{http://dx.doi.org/10.3982/ECTA13288}.

\bibitem[{Kouvelis et~al.(2017)Kouvelis, Milner, \protect\BIBand{} Tian}]{kouvelis_clinical_2017}
Kouvelis P, Milner J, Tian Z (2017) Clinical {Trials} for {New} {Drug} {Development}: {Optimal} {Investment} and {Application}. \emph{Manufacturing \& Service Operations Management} 19(3):437--452, ISSN 1523-4614, 1526-5498, \urlprefix\url{http://dx.doi.org/10.1287/msom.2017.0616}.

\bibitem[{Kumar et~al.(2020)Kumar, Zhou, Tucker, \protect\BIBand{} Levine}]{kumar2020conservative}
Kumar A, Zhou A, Tucker G, Levine S (2020) Conservative q-learning for offline reinforcement learning. \emph{Advances in neural information processing systems} 33:1179--1191.

\bibitem[{Kuzborskij et~al.(2021)Kuzborskij, Vernade, Gyorgy, \protect\BIBand{} Szepesvari}]{kuzborskij_confident_2021}
Kuzborskij I, Vernade C, Gyorgy A, Szepesvari C (2021) Confident {Off}-{Policy} {Evaluation} and {Selection} through {Self}-{Normalized} {Importance} {Weighting}. \emph{Proceedings of {The} 24th {International} {Conference} on {Artificial} {Intelligence} and {Statistics}}, 640--648 (PMLR), \urlprefix\url{https://proceedings.mlr.press/v130/kuzborskij21a.html}.

\bibitem[{Ladhania et~al.(2023)Ladhania, Spiess, Ungar, \protect\BIBand{} Wu}]{ladhania_personalized_2023}
Ladhania R, Spiess J, Ungar L, Wu W (2023) Personalized {Assignment} to {One} of {Many} {Treatment} {Arms} via {Regularized} and {Clustered} {Joint} {Assignment} {Forests}. \urlprefix\url{http://dx.doi.org/10.48550/ARXIV.2311.00577}.

\bibitem[{Lai \protect\BIBand{} Robbins(1985)}]{lai1985asymptotically}
Lai TL, Robbins H (1985) Asymptotically efficient adaptive allocation rules. \emph{Advances in applied mathematics} 6(1):4--22.

\bibitem[{Lekwijit et~al.(2024)Lekwijit, Terwiesch, Asch, \protect\BIBand{} Volpp}]{lekwijit2024evaluating}
Lekwijit S, Terwiesch C, Asch DA, Volpp KG (2024) Evaluating the efficacy of connected healthcare: An empirical examination of patient engagement approaches and their impact on readmission. \emph{Management Science} 70(6):3417--3446.

\bibitem[{Levine et~al.(2020)Levine, Kumar, Tucker, \protect\BIBand{} Fu}]{levine2020offline}
Levine S, Kumar A, Tucker G, Fu J (2020) Offline reinforcement learning: Tutorial, review, and perspectives on open problems. \emph{arXiv preprint arXiv:2005.01643} .

\bibitem[{Li et~al.(2011)Li, Chu, Langford, \protect\BIBand{} Wang}]{li_unbiased_2011}
Li L, Chu W, Langford J, Wang X (2011) Unbiased offline evaluation of contextual-bandit-based news article recommendation algorithms. \emph{Proceedings of the fourth {ACM} international conference on {Web} search and data mining}, 297--306 (Hong Kong China: ACM), ISBN 9781450304931, \urlprefix\url{http://dx.doi.org/10.1145/1935826.1935878}.

\bibitem[{Lu \protect\BIBand{} Hardin(2021)}]{Lu_varPred}
Lu B, Hardin J (2021) A unified framework for random forest prediction error estimation. \emph{Journal of Machine Learning Research} 22(8):1--41, \urlprefix\url{http://jmlr.org/papers/v22/18-558.html}.

\bibitem[{Ma et~al.(2022)Ma, Zeng, \protect\BIBand{} Liu}]{Ma_adaptive_fusion}
Ma H, Zeng D, Liu Y (2022) Learning individualized treatment rules with many treatments: A supervised clustering approach using adaptive fusion. Koyejo S, Mohamed S, Agarwal A, Belgrave D, Cho K, Oh A, eds., \emph{Advances in Neural Information Processing Systems}, volume~35, 15956--15969 (Curran Associates, Inc.), \urlprefix\url{https://proceedings.neurips.cc/paper_files/paper/2022/file/663865ea167425c6c562cb0b6bcf76c7-Paper-Conference.pdf}.

\bibitem[{Milkman et~al.(2011)Milkman, Beshears, Choi, Laibson, \protect\BIBand{} Madrian}]{milkman_using_2011}
Milkman KL, Beshears J, Choi JJ, Laibson D, Madrian BC (2011) Using implementation intentions prompts to enhance influenza vaccination rates. \emph{Proceedings of the National Academy of Sciences} 108(26):10415--10420, ISSN 0027-8424, 1091-6490, \urlprefix\url{http://dx.doi.org/10.1073/pnas.1103170108}.

\bibitem[{Milkman et~al.(2022)Milkman, Gandhi, Patel, Graci, Gromet, Ho, Kay, Lee, Rothschild, Bogard, Brody, Chabris, Chang, Chapman, Dannals, Goldstein, Goren, Hershfield, Hirsch, Hmurovic, Horn, Karlan, Kristal, Lamberton, Meyer, Oakes, Schweitzer, Shermohammed, Talloen, Warren, Whillans, Yadav, Zlatev, Berman, Evans, Ladhania, Ludwig, Mazar, Mullainathan, Snider, Spiess, Tsukayama, Ungar, Van Den~Bulte, Volpp, \protect\BIBand{} Duckworth}]{milkman_680000-person_2022}
Milkman KL, Gandhi L, Patel MS, Graci HN, Gromet DM, Ho H, Kay JS, Lee TW, Rothschild J, Bogard JE, Brody I, Chabris CF, Chang E, Chapman GB, Dannals JE, Goldstein NJ, Goren A, Hershfield H, Hirsch A, Hmurovic J, Horn S, Karlan DS, Kristal AS, Lamberton C, Meyer MN, Oakes AH, Schweitzer ME, Shermohammed M, Talloen J, Warren C, Whillans A, Yadav KN, Zlatev JJ, Berman R, Evans CN, Ladhania R, Ludwig J, Mazar N, Mullainathan S, Snider CK, Spiess J, Tsukayama E, Ungar L, Van Den~Bulte C, Volpp KG, Duckworth AL (2022) A 680,000-person megastudy of nudges to encourage vaccination in pharmacies. \emph{Proceedings of the National Academy of Sciences} 119(6):e2115126119, ISSN 0027-8424, 1091-6490, \urlprefix\url{http://dx.doi.org/10.1073/pnas.2115126119}.

\bibitem[{Murphy(2003)}]{murphy2003optimal}
Murphy SA (2003) Optimal dynamic treatment regimes. \emph{Journal of the Royal Statistical Society Series B: Statistical Methodology} 65(2):331--355.

\bibitem[{Nachum et~al.(2019)Nachum, Dai, Kostrikov, Chow, Li, \protect\BIBand{} Schuurmans}]{nachum2019algaedice}
Nachum O, Dai B, Kostrikov I, Chow Y, Li L, Schuurmans D (2019) Algaedice: Policy gradient from arbitrary experience. \emph{arXiv preprint arXiv:1912.02074} .

\bibitem[{Precup et~al.(2001)Precup, Sutton, \protect\BIBand{} Dasgupta}]{precup2001off}
Precup D, Sutton RS, Dasgupta S (2001) Off-policy temporal-difference learning with function approximation. \emph{ICML}, 417--424.

\bibitem[{Qian \protect\BIBand{} Murphy(2011)}]{qian_performance_2011}
Qian M, Murphy SA (2011) Performance guarantees for individualized treatment rules. \emph{The Annals of Statistics} 39(2), ISSN 0090-5364, \urlprefix\url{http://dx.doi.org/10.1214/10-AOS864}.

\bibitem[{Regan et~al.(2017)Regan, Bloomfield, Peters, \protect\BIBand{} Effler}]{regan_randomized_2017}
Regan AK, Bloomfield L, Peters I, Effler PV (2017) Randomized {Controlled} {Trial} of {Text} {Message} {Reminders} for {Increasing} {Influenza} {Vaccination}. \emph{The Annals of Family Medicine} 15(6):507--514, ISSN 1544-1709, 1544-1717, \urlprefix\url{http://dx.doi.org/10.1370/afm.2120}.

\bibitem[{Robins et~al.(1994)Robins, Rotnitzky, \protect\BIBand{} Zhao}]{robins1994estimation}
Robins JM, Rotnitzky A, Zhao LP (1994) Estimation of regression coefficients when some regressors are not always observed. \emph{Journal of the American statistical Association} 89(427):846--866.

\bibitem[{Rosenbaum \protect\BIBand{} Rubin(1983)}]{rosenbaum_central_1983}
Rosenbaum PR, Rubin DB (1983) The central role of the propensity score in observational studies for causal effects. \emph{Biometrika} 70(1):41--55, ISSN 0006-3444, 1464-3510, \urlprefix\url{http://dx.doi.org/10.1093/biomet/70.1.41}.

\bibitem[{Rosenbaum \protect\BIBand{} Rubin(1984)}]{rosenbaum1984reducing}
Rosenbaum PR, Rubin DB (1984) Reducing bias in observational studies using subclassification on the propensity score. \emph{Journal of the American statistical Association} 79(387):516--524.

\bibitem[{Rubin(1974)}]{rubin_estimating_1974}
Rubin D (1974) Estimating causal effects of treatments in randomized and nonrandomized studies. \emph{Journal of Educational Psychology} 66(5):688--701, \urlprefix\url{http://dx.doi.org/https://doi.org/10.1037/h0037350}.

\bibitem[{Saito et~al.(2024)Saito, Yao, \protect\BIBand{} Joachims}]{saito_potec:_2024}
Saito Y, Yao J, Joachims T (2024) {POTEC}: {Off}-{Policy} {Learning} for {Large} {Action} {Spaces} via {Two}-{Stage} {Policy} {Decomposition}. \urlprefix\url{http://dx.doi.org/10.48550/ARXIV.2402.06151}.

\bibitem[{Sedrakyan \protect\BIBand{} Sedrakyan(2018)}]{sedrakyan2018useful}
Sedrakyan H, Sedrakyan N (2018) A useful inequality. \emph{Algebraic Inequalities}, 107--125 (Springer).

\bibitem[{Shaffer(1995)}]{shaffer_multiple_1995}
Shaffer JP (1995) Multiple {Hypothesis} {Testing}. \emph{Annual Review of Psychology} 46(1):561--584, ISSN 0066-4308, 1545-2085, \urlprefix\url{http://dx.doi.org/10.1146/annurev.ps.46.020195.003021}.

\bibitem[{Sheeran(2002)}]{sheeran_intentionbehavior_2002}
Sheeran P (2002) Intention—{Behavior} {Relations}: {A} {Conceptual} and {Empirical} {Review}. \emph{European Review of Social Psychology} 12(1):1--36, ISSN 1046-3283, 1479-277X, \urlprefix\url{http://dx.doi.org/10.1080/14792772143000003}.

\bibitem[{Smith \protect\BIBand{} Winkler(2006)}]{smith2006optimizer}
Smith JE, Winkler RL (2006) The optimizer’s curse: Skepticism and postdecision surprise in decision analysis. \emph{Management Science} 52(3):311--322.

\bibitem[{Spiess et~al.(2023)Spiess, Syrgkanis, \protect\BIBand{} Wang}]{spiess_finding_2023}
Spiess J, Syrgkanis V, Wang VY (2023) Finding {Subgroups} with {Significant} {Treatment} {Effects}. \urlprefix\url{http://dx.doi.org/10.48550/arXiv.2103.07066}, arXiv:2103.07066.

\bibitem[{Sutton et~al.(1998)Sutton, Barto et~al.}]{sutton1998reinforcement}
Sutton RS, Barto AG, et~al. (1998) \emph{Reinforcement learning: An introduction}, volume~1 (MIT press Cambridge).

\bibitem[{Swaminathan \protect\BIBand{} Joachims(2015{\natexlab{a}})}]{swaminathan_batch_2015}
Swaminathan A, Joachims T (2015{\natexlab{a}}) Batch learning from logged bandit feedback through counterfactual risk minimization. \emph{J. Mach. Learn. Res.} 16(1):1731--1755, ISSN 1532-4435.

\bibitem[{Swaminathan \protect\BIBand{} Joachims(2015{\natexlab{b}})}]{NIPS2015_39027dfa}
Swaminathan A, Joachims T (2015{\natexlab{b}}) The self-normalized estimator for counterfactual learning. Cortes C, Lawrence N, Lee D, Sugiyama M, Garnett R, eds., \emph{Advances in Neural Information Processing Systems}, volume~28 (Curran Associates, Inc.), \urlprefix\url{https://proceedings.neurips.cc/paper_files/paper/2015/file/39027dfad5138c9ca0c474d71db915c3-Paper.pdf}.

\bibitem[{Thaler(1988)}]{thaler1988anomalies}
Thaler RH (1988) Anomalies: The winner's curse. \emph{Journal of economic perspectives} 2(1):191--202.

\bibitem[{Tian et~al.(2023)Tian, Hazen, \protect\BIBand{} Li}]{tian_optimal_2023}
Tian Z, Hazen GB, Li H (2023) Optimal {Enrollment} in {Late}-{Stage} {New} {Drug} {Development} with {Learning} of {Drug}’s {Efficacy} for {Group}-{Sequential} {Clinical} {Trials}. \emph{Manufacturing \& Service Operations Management} 25(1):88--107, ISSN 1523-4614, 1526-5498, \urlprefix\url{http://dx.doi.org/10.1287/msom.2022.1162}.

\bibitem[{Watkins \protect\BIBand{} Dayan(1992)}]{watkins1992q}
Watkins CJ, Dayan P (1992) Q-learning. \emph{Machine learning} 8(3):279--292.

\bibitem[{Xu et~al.(2025)Xu, Thomadsen, \protect\BIBand{} Zhang}]{xu2025winner}
Xu S, Thomadsen R, Zhang D (2025) The winner's curse in data-driven decision making: Evidence and solutions. \emph{Available at SSRN 5930537} .

\bibitem[{Zhan et~al.(2024)Zhan, Ren, Athey, \protect\BIBand{} Zhou}]{zhan2024policy}
Zhan R, Ren Z, Athey S, Zhou Z (2024) Policy learning with adaptively collected data. \emph{Management Science} 70(8):5270--5297.

\bibitem[{Zhang et~al.(2012)Zhang, Tsiatis, Laber, \protect\BIBand{} Davidian}]{zhang2012robust}
Zhang B, Tsiatis AA, Laber EB, Davidian M (2012) A robust method for estimating optimal treatment regimes. \emph{Biometrics} 68(4):1010--1018.

\bibitem[{Zhou et~al.(2017)Zhou, Mayer-Hamblett, Khan, \protect\BIBand{} Kosorok}]{zhou_residual_2017}
Zhou X, Mayer-Hamblett N, Khan U, Kosorok MR (2017) Residual {Weighted} {Learning} for {Estimating} {Individualized} {Treatment} {Rules}. \emph{Journal of the American Statistical Association} 112(517):169--187, ISSN 0162-1459, 1537-274X, \urlprefix\url{http://dx.doi.org/10.1080/01621459.2015.1093947}.

\bibitem[{Zrnic \protect\BIBand{} Fithian(2024)}]{zrnic_flexible_2024}
Zrnic T, Fithian W (2024) A {Flexible} {Defense} {Against} the {Winner}'s {Curse}. \urlprefix\url{http://dx.doi.org/10.48550/ARXIV.2411.18569}.

\end{thebibliography}

\begin{APPENDIX}{}

\section{Proofs}
\label{apx:proofs}
Here we give proofs for all Propositions and Theorems in this article that are note given in the main text.

\bigskip

\begin{proof}{Proof of Proposition~\ref{prop:relax}.}
We will show $\pi'\in\Pi_{\text{Opt}} \implies \pi'\in\Pi_{\Lambda}$. 

Let $\pi'\in\Pi_{\text{Opt}}$. Thus $\tau(\pi') \geq 0$, otherwise $\pi'$ would induce a negative expected improvement and $z$-score and can be eliminated from the Pareto frontier by the observational policy $\pi^o$ which we know achieves $\tau(\pi') = 0$ with certainty. As no policy achieving an expected improvement of $\tau(\pi')$ achieves a lower expected $z$-score than $\pi'$, 
$$\pi' \in \underset{\left\{\pi\in\left(\Delta^K\right)^N\middle| \tau(\pi) = \tau(\pi')\right\}}{\mathrm{argmin}}s(\pi).$$
Therefore, if $\lambda = \tau(\pi')$,
$$\pi' \in \underset{\left\{\pi\in\left(\Delta^K\right)^N\middle| \tau(\pi) = \lambda\right\}}{\mathrm{argmin}}s^2(\pi),$$
as $s(\pi)$ is non-negative and $s^2(\pi)$ is thus a monotonic transformation of $s(\pi)$.
Thus $\pi' \in \Pi_\Lambda$.

\qed
\end{proof}

\bigskip

\begin{proof}{Proof of Proposition~\ref{prop:convex}.}
We will derive the form of $s^2(\pi)$ given in the proposition, which is convex in $\pi$ due to being a sum of squares of affine functions of $\pi$. From its definition we see $\tau(\pi)$ is affine in $\pi$.

We can decompose $s^2(\pi)$ across units via Bienaym\'{e}'s identity as both observed treatments $T^o_n$ and outcomes $Y^o_n$ are independent across units in the Rubin causal model,
\begin{align*}
s^2(\pi) &= \mathrm{Var}\left(\hat{\tau}(\pi)\right) \\
&=\mathrm{Var}\left(\frac{1}{N}\sum_{n=1}^N Y_n^o(W_n - 1)\right)  \\
&= \frac{1}{N^2} \sum_{n=1}^N \mathrm{Var}\left(Y_n^o\left(W_n-1\right)\right).
\end{align*}
We then apply the law of total variance to each unit,
\begin{align*}
\mathrm{Var}(Y_n^o(W_n-1)) &= \mathbb{E}\left[\mathrm{Var}\left(Y_n^o\left(W_n-1\right)|T_n^o\right)\right] + \mathrm{Var}\left(\mathbb{E}\left[Y_n^o\left(W_n-1\right)|T_n^o\right]\right) \\
&=\mathbb{E}\left[\sigma^2_{n,T_n^o}\left(\frac{\pi_{n,T_n^o}-\pi^o_{n,T_n^o}}{\pi^o_{n,T_n^o}}\right)^2\right] + \mathrm{Var}\left(\mu_{n,T_n^o} \left(\frac{\pi_{n,T_n^o}-\pi^o_{n,T_n^o}}{\pi^o_{n,T_n^o}}\right)\right) \\
&=\sum_{t=0}^K \frac{\sigma_{n,t}^2 \left(\pi_{n,t}-\pi^o_{n,t}\right)^2}{\pi^o_{n,t}} + \pi_{n,t}^o \left(\frac{\mu_{n,t} \left(\pi_{n,t}-\pi^o_{n,t}\right)}{\pi^o_{n,t}} - \sum_{v=0}^K \mu_{n,v} \left(\pi_{n,v}-\pi^o_{n,v}\right)\right)^2.
\end{align*}
Plugging these unit specific variances back into the above sum yields the version of $s(\pi)^2$ presented. 

\qed
\end{proof}

\bigskip

\begin{proof}{Proof of Proposition~\ref{prop:FOCs}.}
We derive first partials for $\tau(\pi)$ and $s^2(\pi)$ with respect to $\pi_{n,t}$ given in the proposition.

As $\tau(\pi)$ is linear in $\pi_{n,t}$ its derivative is simple,
\begin{align*}
\frac{\partial \tau(\pi)}{\partial\pi_{n,t}} &= \frac{\partial}{\partial\pi_{n,t}} \frac{1}{N}\sum_{n=1}^N\sum_{v=0}^K \mu_{n,v}\left(\pi_{n,v} - \pi^o_{n,v}\right) \\
&=\frac{\mu_{n,t}}{N}.
\end{align*}

We first expand the form of $s^2(\pi)$given in Proposition~\ref{prop:convex} to,
\begin{align*}
s^2(\pi) &= \frac{1}{N^2} \sum_{n=1}^N \sum_{t=0}^K \frac{\sigma_{n,t}^2 \left(\pi_{n,t}-\pi^o_{n,t}\right)^2}{\pi^o_{n,t}} + \pi_{n,t}^o \left(\frac{\mu_{n,t} \left(\pi_{n,t}-\pi^o_{n,t}\right)}{\pi^o_{n,t}} - \sum_{v=0}^K \mu_{n,v} \left(\pi_{n,v}-\pi^o_{n,v}\right)\right)^2\\
&= \frac{1}{N^2} \sum_{n=1}^N \left[\left(\sum_{t=0}^K \frac{\left(\mu_{n,t}^2+\sigma_{n,t}^2\right) \left(\pi_{n,t}-\pi^o_{n,t}\right)^2}{\pi^o_{n,t}}\right)-\left(\sum_{t=0}^K \mu_{n,t} \left(\pi_{n,t}-\pi^o_{n,t}\right)\right)^2\right],
\end{align*}
then calculate its derivative,
\begin{align*}
\frac{\partial s^2(\pi)}{\partial \pi_{n,t}} &= \frac{\partial}{\partial \pi_{n,t}} \frac{1}{N^2} \sum_{n=1}^N \left[\left(\sum_{v=0}^K \frac{\left(\mu_{n,v}^2+\sigma_{n,v}^2\right) \left(\pi_{n,v}-\pi^o_{n,v}\right)^2}{\pi^o_{n,v}}\right)-\left(\sum_{v=0}^K \mu_{n,v} \left(\pi_{n,v}-\pi^o_{n,v}\right)\right)^2\right] \\
&= \frac{2}{N^2} \left[\frac{\mu_{n,t}^2 + \sigma_{n,t}^2}{\pi_{n,t}^o} (\pi_{n,t} - \pi_{n,t}^o) - \mu_{n,t} \sum_{v=0}^K \mu_{n,v} \left(\pi_{n,v} - \pi_{n,v}^o\right)\right].
\end{align*}

\qed

\end{proof}

\bigskip

\begin{proof}{Proof of Proposition~\ref{prop:form}.}
We use the primal constraint of $\sum_{t=0}^K \pi_{n,t} = 1$ to show any optimal solution $(\pi^*,\zeta^*,\kappa^*)$ satisfies the proposition and than $Q_n$ as stated is positive.

We can rearrange the first order condition of $(\pi^*,\zeta^*,\beta^*,\kappa^*)$ as,
$$\pi^*_{n,t} =\pi_{n,t}^o\left( 1+\frac{\mu_{n,t}\left(\frac{N}{2}\zeta^*+\sum_{v=0}^K \mu_{n,v} \left(\pi_{n,v}^* - \pi_{n,v}^o\right)\right)+ \frac{N^2}{2}\left(\beta_n^* + \kappa_{n,t}^*\right)}{\mu_{n,t^2} + \sigma_{n,t}^2}\right).$$
For each unit we then sum the first order conditions across treatment, making use of $\sum_{t=0}^K \pi^*_{n,t} = 1$ and $\sum_{t=0}^K \pi_{n,t}^o = 1$,
{\begin{align*}
\sum_{t = 0}^K \pi^*_{n,t} &= \left(\sum_{t = 0}^K \pi^o_{n,t}\right) + \sum_{t=0}^K \frac{\pi_{n,t}^o\left[\mu_{n,t}\left(\frac{N}{2}\zeta^*+\sum_{v=0}^K \mu_{n,v} \left(\pi_{n,v}^* - \pi_{n,v}^o\right)\right)+ \frac{N^2}{2}\left(\beta_n^* + \kappa_{n,t}^*\right)\right]}{\mu_{n,t^2} + \sigma_{n,t}^2} \\
-\frac{\beta_n^*N^2}{2}\left( \sum_{t=0}^K \frac{\pi^o_{n,t}}{\mu_{n,t}^2 + \sigma_{n,t}^2}\right) &= \frac{N^2}{2} \left( \sum_{t=0}^K \frac{\pi^o_{n,t}\kappa_{n,t}^*}{\mu_{n,t}^2 + \sigma_{n,t}^2} \right) + \left(\frac{N}{2}\zeta^*+\sum_{v=0}^K \mu_{n,v} \left(\pi_{n,v}^* - \pi_{n,v}^o\right)\right)\left( \sum_{t=0}^K \frac{\pi^o_{n,t}\mu_{n,t}^*}{\mu_{n,t}^2 + \sigma_{n,t}^2} \right) \\
\beta_n^* &= - \tilde{\kappa}_n^* - \left(\frac{1}{N}\zeta^*+ \frac{2}{N^2}\sum_{v=0}^K \mu_{n,v} \left(\pi_{n,v}^* - \pi_{n,v}^o\right)\right)\tilde{\mu}_n,
\end{align*}
where,
\begin{align*}
\tilde{\kappa}_n^*&=\frac{\sum_{t=0}^K \frac{\pi^o_{n,t}}{\mu_{n,t}^2 + \sigma_{n,t}^2} \kappa_{n,t}^*}{\sum_{t=0}^K \frac{\pi^o_{n,t}}{\mu_{n,t}^2 + \sigma_{n,t}^2}} & &\text{and} & \tilde{\mu}_n&= \frac{\sum_{t=0}^K \frac{\pi^o_{n,t}}{\mu_{n,t}^2 + \sigma_{n,t}^2}\mu_{n,t}}{\sum_{t=0}^K \frac{\pi^o_{n,t}}{\mu_{n,t}^2 + \sigma_{n,t}^2}}.
\end{align*}}
Substituting this form of $\beta_n^*$ back into the first order condition gives,
\begin{align*}
\pi^*_{n,t} &= \pi_{n,t}^o\left( 1+\frac{\left(\frac{N}{2}\zeta^*+\sum_{v=0}^K \mu_{n,v} \left(\pi_{n,v}^* - \pi_{n,v}^o\right)\right) \left(\mu_{n,t} - \tilde{\mu}_n\right)+ \frac{N^2}{2}\left(\kappa_{n,t}^*- \tilde{\kappa}_n^*\right)}{\mu_{n,t^2} + \sigma_{n,t}^2}\right).
\end{align*}
Next we subtract multiply each equation by $\mu_{n,t}$ and sum again to find a substitution for $\sum_{v=0}^K\mu_{n,v} \left(\pi^*_{n,v} - \pi^o_{n,v}\right)$,
\begin{align*}
\sum_{t=0}^K \mu_{n,t} \pi_{n,t}^* &= \left(\sum_{t=0}^K  \mu_{n,t}\pi_{n,t}^o\right) + \sum_{t=0}^K \frac{\pi_{n,t}^o\mu_{n,t}}{\mu_{n,t}^2 + \sigma_{n,t}^2} \left[\left(\frac{N}{2}\zeta^*+\sum_{v=0}^K \mu_{n,v} \left(\pi_{n,v}^* - \pi_{n,v}^o\right)\right) \left(\mu_{n,t} - \tilde{\mu}_n\right)+ \frac{N^2}{2}\left(\kappa_{n,t}^*- \tilde{\kappa}_n^*\right)\right]
\end{align*}
which can be rearranged to,
\begin{align*}
\sum_{t=0}^K \mu_{n,t}\left(\pi^*_{n,t} - \pi^o_{n,t}\right)&= \frac{ \zeta^* \left(\sum_{t=0}^K \frac{\pi^o_{n,t}}{\mu_{n,t}^2 + \sigma_{n,t}^2} \mu_{n,t} \left(\mu_{n,t} - \tilde{\mu}_n\right)\right) + N\left( \sum_{t=0}^K \frac{\pi^o_{n,t}}{\mu_{n,t}^2 + \sigma_{n,t}^2} \mu_{n,t} \left(\kappa^*_{n,t} - \tilde{\kappa}^*_n\right)\right) }{2Q_n}N
\end{align*}
where $$Q_n = 1 - \sum_{t=0}^K \frac{\pi^o_{n,t}}{\mu_{n,t}^2 + \sigma_{n,t}^2} \mu_{n,t} \left(\mu_{n,t} - \tilde{\mu}_n\right).$$
Thus we can write the first order condition as,
\begin{align*}\pi_{n,t}^* &= \pi^o_{n,t} \left(1+\frac{Q_n^{-1}\left(\zeta^* + N\sum_{v=0}^K \frac{\pi^o_{n,v}}{\mu_{n,v}^2 + \sigma_{n,v}^2} \mu_{n,v} \left(\kappa^*_{n,v} - \tilde{\kappa}^*_n\right)\right) \left(\mu_{n,t} - \tilde{\mu}_n\right) + N\left(\kappa^*_{n,t} - \tilde{\kappa}^*_n \right)}{2\left(\mu_{n,t}^2 + \sigma_{n,t}^2\right) }N\right) \\
&= \pi^o_{n,t} \left(1+\frac{Q_n^{-1}\left(\zeta^* + N\sum_{v=0}^K \frac{\pi^o_{n,v}}{\mu_{n,v}^2 + \sigma_{n,v}^2} \kappa^*_{n,v} \left(\mu_{n,v} - \tilde{\mu}_n\right)\right) \left(\mu_{n,t} - \tilde{\mu}_n\right) + N\left(\kappa^*_{n,t} - \tilde{\kappa}^*_n \right)}{2\left(\mu_{n,t}^2 + \sigma_{n,t}^2\right) }N\right).
\end{align*}
\end{proof}

\qed

\bigskip

\begin{proof}{Proof of Theorem~\ref{thm:binary}.}
We will show that the first order condition has a unique solution $(\pi^*,\zeta^*,\kappa^*)$ for ever $\zeta^* \in \mathbb{R}$, then show that those solutions with $\zeta \geq 0$ characterize $\Pi_{\Lambda}$. Throughout we focus on $\pi_{n,1}^*$ as $\pi_{n,0}^* = 1 - \pi_{n,1}^*$.

First we apply $K=1$ to Proposition~\ref{prop:form} to derive the general first order condition of the problems in $\Pi_{\Lambda}$ for the single-treatment case. We first give the expanded forms of the subject specific terms, noting that $\pi_{n,0}^o = 1-\pi_{n,1}^o$,
\begin{align*}
\tilde{\mu}_n &= \frac{ \frac{1-\pi^o_{n,1}}{\mu_{n,0}^2 + \sigma_{n,0}^2}\mu_{n,0} + \frac{\pi^o_{n,1}}{\mu_{n,1}^2 + \sigma_{n,1}^2}\mu_{n,1}}{\frac{1-\pi^o_{n,1}}{\mu_{n,0}^2 + \sigma_{n,0}^2} + \frac{\pi^o_{n,1}}{\mu_{n,1}^2 + \sigma_{n,1}^2}}\\
\tilde{\kappa}^*_n &= \frac{\frac{1-\pi^o_{n,1}}{\mu_{n,0}^2 + \sigma_{n,0}^2} \kappa_{n,0}^* + \frac{\pi^o_{n,1}}{\mu_{n,1}^2 + \sigma_{n,1}^2}\kappa_{n,1}^*}{\frac{1-\pi^o_{n,1}}{\mu_{n,0}^2 + \sigma_{n,0}^2} + \frac{\pi^o_{n,1}}{\mu_{n,1}^2 + \sigma_{n,1}^2}} \\
Q_n&=1 - \frac{1-\pi^o_{n,1}}{\mu_{n,0}^2 + \sigma_{n,0}^2} \mu_{n,0}^2  - \frac{\pi^o_{n,1}}{\mu_{n,1}^2 + \sigma_{n,1}^2} \mu_{n,1}^2  + \frac{\left( \frac{1-\pi^o_{n,1}}{\mu_{n,0}^2 + \sigma_{n,0}^2}\mu_{n,0} + \frac{\pi^o_{n,1}}{\mu_{n,1}^2 + \sigma_{n,1}^2}\mu_{n,1}\right)^2}{\frac{1-\pi^o_{n,1}}{\mu_{n,0}^2 + \sigma_{n,0}^2} + \frac{\pi^o_{n,1}}{\mu_{n,1}^2 + \sigma_{n,1}^2}}\\
&= 1 - \frac{\frac{(1-\pi^o_{n,1})\pi^o_{n,1}}{(\mu_{n,0}^2 + \sigma_{n,0}^2)(\mu_{n,1}^2 + \sigma_{n,1}^2)}(\mu_{n,0}^2+\mu_{n,1}^2)-2 \frac{(1-\pi^o_{n,1})\pi^o_{n,1}}{(\mu_{n,0}^2 + \sigma_{n,0}^2)(\mu_{n,1}^2 + \sigma_{n,1}^2)} \mu_{n,0}\mu_{n,1}}{\frac{1-\pi^o_{n,1}}{\mu_{n,0}^2 + \sigma_{n,0}^2} + \frac{\pi^o_{n,1}}{\mu_{n,1}^2 + \sigma_{n,1}^2}} \\
&= 1 - \frac{(1-\pi_{n,1}^o)\pi_{n,1}^o}{(1-\pi_{n,1}^o)(\mu_{n,1}^2 + \sigma_{n,1}^2) + \pi_{n,1}^o(\mu_{n,0}^2 + \sigma_{n,0}^2)} \left(\mu_{n,1} - \mu_{n,0}\right)^2
\end{align*}
We then simplify the first order condition of $\pi^*_{n,1}$,
\begin{align*}
\pi_{n,1}^* &= \pi^o_{n,1} \left(1+\frac{Q_n^{-1}\left(\zeta^* + N\sum_{v=0}^K \frac{\pi^o_{n,v}}{\mu_{n,v}^2 + \sigma_{n,v}^2} \kappa^*_{n,v} \left(\mu_{n,v} - \tilde{\mu}_n\right)\right) \left(\mu_{n,1} - \tilde{\mu}_n\right) + N\left(\kappa^*_{n,1} - \tilde{\kappa}^*_n \right)}{2\left(\mu_{n,1}^2 + \sigma_{n,1}^2\right) }N\right)\\
&= \pi_{n,1}^o + \frac{N^2}{2} \left(\frac{(1-\pi^o_{n,1})\pi^o_{n,1}}{(1-\pi_{n,1}^o)(\mu_{n,1}^2 + \sigma_{n,1}^2) + \pi_{n,1}^o(\mu_{n,0}^2 + \sigma_{n,0}^2)}\right) \left(\kappa_{n,1}^* - \kappa^*_{n,0}\right)\\
&\hspace{10pt}+\frac{\left(\zeta^* + N\left(\frac{(1-\pi^o_{n,1})\pi^o_{n,1}}{(1-\pi_{n,1}^o)(\mu_{n,1}^2 + \sigma_{n,1}^2) + \pi_{n,1}^o(\mu_{n,0}^2 + \sigma_{n,0}^2)}\right)\left(\mu_{n,1}-\mu_{n,0}\right)\left(\kappa_{n,1}^*-\kappa_{n,0}^*\right) \right)\left(\mu_{n,1}-\mu_{n,0}\right)}{2\left(\frac{1-\pi_{n,1}^o}{\mu_{n,0}^2 + \sigma_{n,0}^2} + \frac{\pi_{n,1}^o}{\mu_{n,1}^2 + \sigma_{n,1}^2} - \frac{(1-\pi^o_{n,1})\pi^o_{n,1}}{(\mu_{n,0}^2 + \sigma_{n,0}^2)(\mu_{n,1}^2 + \sigma_{n,1}^2)}(\mu_{n,1} - \mu_{n,0})^2 \right) \left(\frac{\mu_{n,0}^2 + \sigma_{n,0}^2}{1-\pi_{n,1}^o}\right)\left(\frac{\mu_{n,1}^2 + \sigma_{n,1}^2}{\pi_{n,1}^o}\right)}N \\
&= \pi^o_{n,1}  + \frac{N\zeta^* (\mu_{n,1} - \mu_{n,0})}{2 \left(\frac{\mu^2_{n,0} + \sigma^2_{n,0}}{1 - \pi^o_{n,1}} + \frac{\mu^2_{n,1} + \sigma^2_{n,1}}{\pi_{n,1}^o}- \left(\mu_{n,1} - \mu_{n,0}\right)^2 \right)} \\
&\hspace{10pt}+ \frac{N^2}{2} \left(\frac{(1-\pi^o_{n,1})\pi^o_{n,1}}{(1-\pi_{n,1}^o)(\mu_{n,1}^2 + \sigma_{n,1}^2) + \pi_{n,1}^o(\mu_{n,0}^2 + \sigma_{n,0}^2)}\right) \left(1 + \frac{\left(\mu_{n,1} - \mu_{n,0}\right)^2}{\frac{\mu^2_{n,0} + \sigma^2_{n,0}}{1 - \pi^o_{n,1}} + \frac{\mu^2_{n,1} + \sigma^2_{n,1}}{\pi_{n,1}^o}- \left(\mu_{n,1} - \mu_{n,0}\right)^2 }\right)\\
&\hspace{40pt}\left(\kappa_{n,1}^* - \kappa_{n,0}^*\right).
\end{align*}
The term in front of $(\kappa_{n,1}^* - \kappa_{n,0}^*)$ is positive, thus by complimentary slackness, $\kappa_{n,1}^*$ is only positive when it needs to increase $\pi^*_{n,1}$ to 0 and $\kappa_{n,1}^*$ is only positive when it needs to decrease $\pi^*_{n,1}$ to 1 (as this is the same as increasing $\pi^*_{n,0}$ to 0). This allows us to rewrite the first order conditions as
\begin{align*}
\pi_{n,1}^* &= \min\left(\left[\pi^o_{n,1}+\frac{N\zeta^*\left(\mu_{n,1} - \mu_{n,0}\right)}{2\left(\frac{\mu^2_{n,0} + \sigma^2_{n,0}}{1 - \pi^o_{n,1}} + \frac{\mu^2_{n,1} + \sigma^2_{n,1}}{\pi_{n,1}^o}- \left(\mu_{n,1} - \mu_{n,0}\right)^2 \right)}  \right]^+,1\right)\\
\kappa_{n,1}^* &= \left[\frac{-\left(\pi^o_{n,1}+\frac{N\zeta^*\left(\mu_{n,1} - \mu_{n,0}\right)}{2\left(\frac{\mu^2_{n,0} + \sigma^2_{n,0}}{1 - \pi^o_{n,1}} + \frac{\mu^2_{n,1} + \sigma^2_{n,1}}{\pi_{n,1}^o}- \left(\mu_{n,1} - \mu_{n,0}\right)^2 \right)}\right)}{\frac{N^2}{2} \left(\frac{(1-\pi^o_{n,1})\pi^o_{n,1}}{(1-\pi_{n,1}^o)(\mu_{n,1}^2 + \sigma_{n,1}^2) + \pi_{n,1}^o(\mu_{n,0}^2 + \sigma_{n,0}^2)}\right) \left(1 + \frac{\left(\mu_{n,1} - \mu_{n,0}\right)^2}{\frac{\mu^2_{n,0} + \sigma^2_{n,0}}{1 - \pi^o_{n,1}} + \frac{\mu^2_{n,1} + \sigma^2_{n,1}}{\pi_{n,1}^o}- \left(\mu_{n,1} - \mu_{n,0}\right)^2 }\right)} \right]^+\\
\kappa_{n,0}^* &= \left[\frac{\left(\pi^o_{n,1}+\frac{N\zeta^*\left(\mu_{n,1} - \mu_{n,0}\right)}{2\left(\frac{\mu^2_{n,0} + \sigma^2_{n,0}}{1 - \pi^o_{n,1}} + \frac{\mu^2_{n,1} + \sigma^2_{n,1}}{\pi_{n,1}^o}- \left(\mu_{n,1} - \mu_{n,0}\right)^2 \right)}\right)-1}{\frac{N^2}{2} \left(\frac{(1-\pi^o_{n,1})\pi^o_{n,1}}{(1-\pi_{n,1}^o)(\mu_{n,1}^2 + \sigma_{n,1}^2) + \pi_{n,1}^o(\mu_{n,0}^2 + \sigma_{n,0}^2)}\right) \left(1 + \frac{\left(\mu_{n,1} - \mu_{n,0}\right)^2}{\frac{\mu^2_{n,0} + \sigma^2_{n,0}}{1 - \pi^o_{n,1}} + \frac{\mu^2_{n,1} + \sigma^2_{n,1}}{\pi_{n,1}^o}- \left(\mu_{n,1} - \mu_{n,0}\right)^2 }\right)}\right]^+
\end{align*}
Thus, for any $\zeta^*\in \mathbb{R}$ there is a unique $\pi^*$ and $\kappa^*$ which satisfy the first order condition for \textit{some} (not necessarily unique) feasible $\lambda$ in the original optimization problem. Moreover every $\lambda$ maps to at least one $\zeta^*$. We thus describe $\pi^*(\zeta)$, the policy induced by the first order conditions when $\zeta^* = \zeta$ as,
$$\pi^*_{n,1}(\zeta)= 
\min\left(\left[\pi^o_{n,1}+\frac{N\zeta\left(\mu_{n,1} - \mu_{n,0}\right)}{2\left(\frac{\mu^2_{n,0} + \sigma^2_{n,0}}{1 - \pi^o_{n,1}} + \frac{\mu^2_{n,1} + \sigma^2_{n,1}}{\pi_{n,1}^o}- \left(\mu_{n,1} - \mu_{n,0}\right)^2 \right)}  \right]^+,1\right).$$
Now we need to show the optimal policies for $\lambda \geq 0$ are exactly those policies detailed by $\pi^*(\zeta)$ for $\zeta \geq 0$. Note for all units, the effect of $\zeta$ on the optimal propensity is determined by the individual treatment effect of the unit, $\mu_{n,1} - \mu_{n,0}$. As the average treatment effect is determined by the sum of individual treatment effects, $\zeta > 0$ induces $\tau(\pi^*(\zeta)) > 0$, $\zeta < 0$ induces $\tau(\pi^*(\zeta)) < 0$, and $\zeta = 0$ induces $\tau(\pi^*(\zeta)) = 0$. Furthermore note that every $\zeta$ induces a policies for some feasible $\lambda$ as for $\zeta$ sufficiently far from $0$, $\pi^*$ either returns the expectation maximizing (for $\zeta > 0$) or expectation minimizing (for $\zeta< 0$) policy.
Thus,
$$\Pi_{\Lambda} = \bigcup_{\zeta \geq 0}\left\{\pi^*(\zeta) \right\},$$
as every $\zeta \geq 0$ generates an optimal policy for some $\lambda \in [0,\tau_{\text{max}}]$ and every feasible $\lambda\in [0,\tau_{\text{max}}]$ has an optimal policy satisfying the first order conditions for some $\zeta \geq 0$.

The form of $\pi_{n,t}^*(\zeta)$ in the theorem statement follows from simple algebra. Finally, we have
\begin{align*}
\tau(\zeta)&=
\frac{1}{N}\sum_{n=1}^N\sum_{t=0}^1\mu_{n,t}\psi_t\alpha_n=\frac{1}{N}\sum_{n=1}^N\tau_n\alpha_n,
\end{align*}
and
\begin{align*}
s^2(\zeta)
&=\frac{1}{N^2}\sum_{n=1}^N\sum_{t=0}^1\left[\frac{\sigma_{n,t}^2\alpha_n^2}{\pi_{n,t}^o}
+\pi_{n,t}^o\left(\frac{\mu_{n,t}\psi_t\alpha_n}{\pi_{n,t}^o}-\sum_{v=0}^1\mu_{n,v}\psi_v\alpha_n\right)^2
\right] \\
&=\frac{1}{N^2}\sum_{n=1}^N\alpha_n^2\sum_{t=0}^1\left[\frac{\sigma_{n,t}^2}{\pi_{n,t}^o}
+\pi_{n,t}^o\left(\frac{\mu_{n,t}\psi_t}{\pi_{n,t}^o}-\mu_{n,1}+\mu_{n,0}\right)^2\right] \\
&=\frac{1}{N^2}\sum_{n=1}^N\alpha_n^2\left[\sum_{t=0}^1\frac{\sigma_{n,t}^2}{\pi_{n,t}^o}
+\pi_{n,0}^o\left(\frac{(1-\pi_{n,0}^o)\mu_{n,0}}{\pi_{n,0}^o}+\mu_{n,1}\right)^2
+\pi_{n,1}^o\left(\frac{(1-\pi_{n,1}^o)\mu_{n,1}}{\pi_{n,1}^o}+\mu_{n,0}\right)^2\right] \\
&=\frac{1}{N^2}\sum_{n=1}^N\alpha_n^2\left[\sum_{t=0}^1\frac{\sigma_{n,t}^2}{\pi_{n,t}^o}
+\frac{(1-\pi_{n,0}^o)\mu_{n,0}^2}{\pi_{n,0}^o}
+\frac{(1-\pi_{n,1}^o)\mu_{n,1}^2}{\pi_{n,1}^o}
+2\mu_{n,0}\mu_{n,1}
\right] \\
&=\frac{1}{N^2}\sum_{n=1}^N\alpha_n^2\left[\sum_{t=0}^1\frac{\sigma_{n,t}^2}{\pi_{n,t}^o}
+\left(\sum_{t=0}^1\sqrt{\frac{1-\pi_{n,t}^o}{\pi_{n,t}^o}}\mu_{n,t}\right)^2\right] \\
&=\frac{1}{N^2}\sum_{n=1}^N\eta_n\alpha_n^2.
\end{align*}

\qed

\end{proof}

\bigskip

\begin{proof}{Proof of Proposition~\ref{prop:zeros}.}
We will show that if $(\pi',\zeta',\kappa')$ and $(\pi'',\zeta'',\kappa'')$ both solve the first order conditions, $\omega_n'' \geq \omega_n' > 0$ for
$$\omega_n^\cdot = \zeta^\cdot + N\sum_{t=0}^K \frac{\pi_{n,t}^o}{\mu_{n,t}^2 + \sigma_{n,t}^2} \kappa_{n,t}^\cdot \left(\mu_{n,t} - \tilde{\mu}_n\right),$$
and $\pi'_{n,t} = 0$ for some unit $n$ and treatment $t$, then $\pi''_{n,t} = 0$. Note that we can write the first order condition as,
$$\pi_{n,t}^*= \pi^o_{n,t} \left(1+\frac{Q_n^{-1}\omega_n^*\left(\mu_{n,t} - \tilde{\mu}_n\right) + N\left(\kappa^*_{n,t} - \tilde{\kappa}^*_n \right)}{2\left(\mu_{n,t}^2 + \sigma_{n,t}^2\right) }N\right).$$ For each unit $n$, we show our claim via an inductive proof over treatments in order of the first value of $\omega^*_n \geq 0$ for which $\pi^*_{n,t}$. For ease we relabel the treatments without loss of generality for each subject $n$, so that treatment $t$ determines the (weak) order of the minimal value of $\omega_n^*$ for which some solution to the first order condition has $\pi_{n,t}^* = 0$.

For the base case we consider of treatment $0$. Let $\hat{\omega}_n$ be the smallest value of $\omega_n^*$ such that $\pi_{n,0}^*=0$.We claim that
$$0 = \pi^o_{n,0} \left(1+\frac{Q_n^{-1}\hat{\omega}_n\left(\mu_{n,0} - \tilde{\mu}_n\right)}{2\left(\mu_{n,0}^2 + \sigma_{n,0}^2\right) }N\right).$$ The only other alternative would be that $\kappa^*_{n,0} - \tilde{\kappa}_n^* < 0$ at $\omega_n^* = \hat{\omega}_n$ as for $0 \leq \omega_n^* < \hat{\omega}_n$,
$$\pi^*_{n,t} = \pi^o_{n,t}\left(1+\frac{Q_n^{-1}\omega^*_n\left(\mu_{n,t} - \tilde{\mu}_n\right)}{2\left(\mu_{n,t}^2 + \sigma_{n,t}^2\right) }N\right) > 0,$$
by the definition of $\hat{\omega}_n$ complementary slackness, and the linearity of $\left(1+\frac{Q_n^{-1}\omega^*_n\left(\mu_{n,t} - \tilde{\mu}_n\right)}{2\left(\mu_{n,t}^2 + \sigma_{n,t}^2\right) }N\right)$ in $\omega_n^*$. However, this cannot be the case as for all $t$, $\kappa_{n,t}^* -\tilde{\kappa}_n^* \leq 0$ for the same reason. This forms a contradiction as $\tilde{\kappa}_n^*$ is a weighted average of $\kappa_{n,t}^*$ across $t$, but is somehow (weakly) greater than all its components as well as strictly greater than $\kappa_{n,0}^*$.

We can therefore say $\mu_{n,0} - \tilde{\mu}_n < 0$ in order to satisfy this equation. Therefore for all $\omega_n^* > \hat{\omega}_n$,
$$0 > \pi^o_{n,0} \left(1+\frac{Q_n^{-1}\omega_n^*\left(\mu_{n,0} - \tilde{\mu}_n\right) - N\tilde{\kappa_n^*}}{2\left(\mu_{n,0}^2 + \sigma_{n,0}^2\right) }N\right),$$
as $\tilde{\kappa_n^*} \geq 0$. This implies that $\kappa_{n,0}^* > 0$ and $\pi_{n,t}^* = 0$ which concludes the base case.

Now we address the inductive step where we assume our proposition holds for treatments $t \in \{0,\ldots,m-1\}$ in order to show it holds for treatment $m$. Redefine $\hat{\omega}_n$ to be the minimal value for which $\pi^*_{n,m}=0$ and let $\omega_n'' \geq \hat{\omega}_n >0$. Note that this means $\pi_{n,t}'' = 0$ for all $t \in \{0,\ldots,m-1\}$ by our inductive hypothesis.

Note that for any solution $(\pi^*,\zeta^*,\kappa^*)$ we can give a closed form for $\tilde{\kappa}^*_n$ in terms of the set of treatments with $0$ propensity, $\mathcal{Z}_n = \{t|\pi_{n,t}^* = 0\}$. Specifically for all $t\in\mathcal{Z}_n$ we can rearrange the first order condition in terms of $\kappa_{n,t}^*$ as,
\begin{align*}
0&= \pi^o_{n,t} \left(1+\frac{Q_n^{-1}\omega_n^*\left(\mu_{n,t} - \tilde{\mu}_n\right) + N\left(\kappa^*_{n,t} - \tilde{\kappa}^*_n \right)}{2\left(\mu_{n,t}^2 + \sigma_{n,t}^2\right) }N\right) \\
-\frac{2(\mu_{n,t}^2 + \sigma_{n,t}^2)}{N^2} &= \frac{\omega_{n}^* (\mu_{n,t} - \tilde{\mu}_n)}{NQ_n} + \kappa_{n,t}^* - \tilde{\kappa}^*_{n} \\
\kappa_{n,t}^* &= \tilde{\kappa}^*_{n} -\frac{2(\mu_{n,t}^2 + \sigma_{n,t}^2)}{N^2} - \frac{\omega_{n}^* (\mu_{n,t} - \tilde{\mu}_n)}{NQ_n}.
\end{align*}
We can then use a weighted sum of these conditions to derive $\tilde{\kappa}^*_{n}$ in terms of $\mathcal{Z}_n$,
\begin{align*}
\underbrace{\sum_{t\in \mathcal{Z}_n}\frac{\pi^o_{n,t}}{\mu_{n,t}^2 + \sigma_{n,t}^2} \kappa_{n,t}^*}_{\tilde{\kappa}^*_{n} \sum_{t=0}^K\frac{\pi^o_{n,t}}{\mu_{n,t}^2 + \sigma_{n,t}^2}} &= \tilde{\kappa_n^*} \sum_{t\in\mathcal{Z}_n} \frac{\pi^o_{n,t}}{\mu_{n,t}^2 + \sigma_{n,t}^2} -\frac{2}{N^2}\sum_{t\in\mathcal{Z}_n} \frac{\pi^o_{n,t}}{\mu_{n,t}^2 + \sigma_{n,t}^2} \left(\mu_{n,t}^2 + \sigma_{n,t}^2\right) - \frac{\omega_n^*}{NQ_n} \sum_{t\in\mathcal{Z}_n} \frac{\pi^o_{n,t}}{\mu_{n,t}^2 + \sigma_{n,t}^2} (\mu_{n,t} - \tilde{\mu}_n) \\
\tilde{\kappa_n^*} &= -\frac{\frac{2}{N^2}\sum_{t\in\mathcal{Z}_n} \pi_{n,t}^o + \frac{\omega_n^*}{NQ_n} \sum_{t\in\mathcal{Z}_n} \frac{\pi^o_{n,t}}{\mu_{n,t}^2 + \sigma_{n,t}^2}(\mu_{n,t} - \tilde{\mu}_n)}{\sum_{t\not\in \mathcal{Z}_n}\frac{\pi^o_{n,t}}{\mu_{n,t}^2 + \sigma_{n,t}^2} }
\end{align*}
Allowing us to rewrite the first order condition for all $t \in \{0,\ldots,K\}$,
\begin{align*}
\pi_{n,t}^* &= \pi^o_{n,t} + \frac{\frac{\pi^o_{n,t}}{\mu_{n,t}^2+\sigma_{n,t}^2}}{\sum_{v\not\in\mathcal{Z}_n}\frac{\pi^o_{n,v}}{\mu_{n,v}^2+\sigma_{n,v}^2}} \sum_{v\in\mathcal{Z}_n} \pi_{n,v}^o + \frac{N \pi^o_{n,t}\omega_{n}^* }{2Q_n(\mu_{n,t}^2+\sigma_{n,t}^2)}\left(\mu_{n,t} - \tilde{\mu}_n + \frac{\sum_{v\in \mathcal{Z}_n} \frac{\pi_{n,v}^o}{\mu_{n,v}^2 + \sigma_{n,v}^2} (\mu_{n,v}-\tilde{\mu}_n)}{\sum_{v\not\in \mathcal{Z}_n} \frac{\pi_{n,v}^o}{\mu_{n,v}^2 + \sigma_{n,v}^2}}\right) \\
&\hspace{25pt} + \frac{N^2 \pi^o_{n,t}}{2(\mu_{n,t}^2+\sigma_{n,t}^2)} \kappa_{n,t}^*\\
&=\pi^o_{n,t} + \frac{\frac{\pi^o_{n,t}}{\mu_{n,t}^2+\sigma_{n,t}^2}}{\sum_{v\not\in\mathcal{Z}_n}\frac{\pi^o_{n,v}}{\mu_{n,v}^2+\sigma_{n,v}^2}} \sum_{v\in\mathcal{Z}_n} \pi_{n,v}^o + \frac{N \pi^o_{n,t}\omega_{n}^* }{2Q_n(\mu_{n,t}^2+\sigma_{n,t}^2)}\left(\mu_{n,t} -  \frac{\sum_{v\not\in \mathcal{Z}_n} \frac{\pi_{n,v}^o}{\mu_{n,v}^2 + \sigma_{n,v}^2} \mu_{n,v}}{\sum_{v\not\in \mathcal{Z}_n} \frac{\pi_{n,v}^o}{\mu_{n,v}^2 + \sigma_{n,v}^2}}\right) + \frac{N^2 \pi^o_{n,t}}{2(\mu_{n,t}^2+\sigma_{n,t}^2)} \kappa_{n,t}^*.
\end{align*}

Thus for all $v \in \mathcal{Z}_n''$,  where $\mathcal{Z}_n'' = \{t|\pi_{n,t}'' = 0\}$,
$$0\geq \pi^o_{n,v} + \frac{\frac{\pi^o_{n,v}}{\mu_{n,v}^2+\sigma_{n,v}^2}}{\sum_{t \not\in\mathcal{Z}_n''}\frac{\pi^o_{n,t}}{\mu_{n,t}^2+\sigma_{n,t}^2}} \sum_{t\in\mathcal{Z}_n''} \pi_{n,t}^o + \frac{N \pi^o_{n,v}\omega_{n}'' }{2Q_n(\mu_{n,v}^2+\sigma_{n,v}^2)}\left(\mu_{n,v} -  \frac{\sum_{t\not\in \mathcal{Z}_n''} \frac{\pi_{n,t}^o}{\mu_{n,t}^2 + \sigma_{n,t}^2} \mu_{n,t}}{\sum_{t\not\in \mathcal{Z}_n''}\frac{\pi_{n,t}^o}{\mu_{n,t}^2 + \sigma_{n,t}^2}}\right).$$ For ease later in the proof we write $\mathcal{B} = \mathcal{Z}_n''\setminus\{0,\ldots,m-1\}$ Furthermore this applies 
Meanwhile, as $\hat{\omega}_n$ is the minimal $\omega^*_n$ for which $\pi^*_{n,m}$=0, we know for all $v \in \{0,\ldots,j\}$ for some $j\geq m$ (in case multiple treatments hit $0$ at the same time) and $\Delta_{n,v}\geq 0$,
\begin{align*}0 &= \pi^o_{n,v} + \frac{\frac{\pi^o_{n,v}}{\mu_{n,v}^2+\sigma_{n,v}^2}}{\sum_{t=j+1}^K\frac{\pi^o_{n,t}}{\mu_{n,t}^2+\sigma_{n,t}^2}} \sum_{t=0}^j \pi_{n,t}^o + \frac{N \pi^o_{n,v}\hat{\omega}_{n} }{2Q_n(\mu_{n,v}^2+\sigma_{n,v}^2)}\left(\mu_{n,v} -  \frac{\sum_{t=j+1}^K \frac{\pi_{n,t}^o}{\mu_{n,t}^2 + \sigma_{n,t}^2} \mu_{n,t}}{\sum_{t=j+1}^K \frac{\pi_{n,t}^o}{\mu_{n,t}^2 + \sigma_{n,t}^2}}\right)+ \Delta_{n,v}\\
&= \pi^o_{n,v} + \frac{\frac{\pi^o_{n,v}}{\mu_{n,v}^2+\sigma_{n,v}^2}}{\sum_{t=m}^K\frac{\pi^o_{n,t}}{\mu_{n,t}^2+\sigma_{n,t}^2}} \sum_{t=0}^{m-1} \pi_{n,t}^o + \frac{N \pi^o_{n,v}\hat{\omega}_{n} }{2Q_n(\mu_{n,v}^2+\sigma_{n,v}^2)}\left(\mu_{n,v} -  \frac{\sum_{t=m}^K \frac{\pi_{n,t}^o}{\mu_{n,t}^2 + \sigma_{n,t}^2} \mu_{n,t}}{\sum_{t=m}^K \frac{\pi_{n,t}^o}{\mu_{n,t}^2 + \sigma_{n,t}^2}}\right) + \Delta_{n,v} \\
&\hspace{20pt} + \frac{\frac{\pi^o_{n,v}}{\mu_{n,v}^2+\sigma_{n,v}^2}}{\sum_{t=m}^K \frac{\pi_{n,t}^o}{\mu_{n,t}^2 + \sigma_{n,t}^2}}\left( \sum_{t=m}^j \frac{\pi_{n,t}^o}{\mu_{n,t}^2 + \sigma_{n,t}^2}\left(\frac{\sum_{t'=0}^j \pi^o_{n,t'}}{\sum_{t'=j+1}^K \frac{\pi^o_{n,t'}}{\mu_{n,t'}^2 + \sigma_{n,t'}^2}} + \frac{N\hat{\omega}_n}{2Q_n}\left(\mu_{n,t} - \frac{\sum_{t' = j+1}^K \frac{\pi_{n,t'}^o}{\mu_{n,t'}^2 + \sigma_{n,t'}^2}\mu_{n,t'}}{\sum_{t' = j+1}^K \frac{\pi_{n,t'}^o}{\mu_{n,t'}^2 + \sigma_{n,t'}^2}}\right)\right)\right)\\
&\hspace{20pt} + \frac{\frac{\pi^o_{n,v}}{\mu_{n,v}^2+\sigma_{n,v}^2}}{\sum_{t=m}^K \frac{\pi_{n,t}^o}{\mu_{n,t}^2 + \sigma_{n,t}^2}}\left( \sum_{t=m}^j \pi^o_{n,t}\right)\\
&= \pi^o_{n,v} + \frac{\frac{\pi^o_{n,v}}{\mu_{n,v}^2+\sigma_{n,v}^2}}{\sum_{t=m}^K\frac{\pi^o_{n,t}}{\mu_{n,t}^2+\sigma_{n,t}^2}} \sum_{t=0}^{m-1} \pi_{n,t}^o + \frac{N \pi^o_{n,v}\hat{\omega}_{n} }{2Q_n(\mu_{n,v}^2+\sigma_{n,v}^2)}\left(\mu_{n,v} -  \frac{\sum_{t=m}^K \frac{\pi_{n,t}^o}{\mu_{n,t}^2 + \sigma_{n,t}^2} \mu_{n,t}}{\sum_{t=m}^K \frac{\pi_{n,t}^o}{\mu_{n,t}^2 + \sigma_{n,t}^2}}\right)\\
&\hspace{20pt}+ \Delta_{n,v} - \frac{\frac{\pi^o_{n,v}}{\mu_{n,v}^2+\sigma_{n,v}^2}}{\sum_{t=m}^K \frac{\pi_{n,t}^o}{\mu_{n,t}^2 + \sigma_{n,t}^2}} \sum_{t=m}^j  \Delta_{n,t}.
\end{align*}
It must be the case for some $v\in\{m,\ldots,j\}$ that $\Delta_{n,v} - \frac{\frac{\pi^o_{n,v}}{\mu_{n,v}^2+\sigma_{n,v}^2}}{\sum_{t=m}^K \frac{\pi_{n,t}^o}{\mu_{n,t}^2 + \sigma_{n,t}^2}} \sum_{t=m}^j  \Delta_{n,t} \geq0$ as the later term is at most a weighted average over $t \in \{m,\ldots,j\}$. We assume WLOG that this is true for $v=m$ and thus can state,
$$0 \geq \pi^o_{n,m} + \frac{\frac{\pi^o_{n,m}}{\mu_{n,m}^2+\sigma_{n,m}^2}}{\sum_{t=m}^K\frac{\pi^o_{n,t}}{\mu_{n,t}^2+\sigma_{n,t}^2}} \sum_{t=0}^{m-1} \pi_{n,t}^o + \frac{N \pi^o_{n,m}\hat{\omega}_{n} }{2Q_n(\mu_{n,m}^2+\sigma_{n,m}^2)}\left(\mu_{n,m} -  \frac{\sum_{t=m}^K \frac{\pi_{n,t}^o}{\mu_{n,t}^2 + \sigma_{n,t}^2} \mu_{n,t}}{\sum_{t=m}^K \frac{\pi_{n,t}^o}{\mu_{n,t}^2 + \sigma_{n,t}^2}}\right).$$
We now show that $\pi''_{n,m} = 0$,
\begin{align*}
\pi_{n,m}'' &= \pi^o_{n,m} + \frac{\frac{\pi^o_{n,m}}{\mu_{n,m}^2+\sigma_{n,m}^2}}{\sum_{t\not\in\mathcal{Z}''_n}\frac{\pi^o_{n,t}}{\mu_{n,t}^2+\sigma_{n,t}^2}} \sum_{t\in\mathcal{Z}''_n} \pi_{n,t}^o + \frac{N \pi^o_{n,m}\omega_{n}'' }{2Q_n(\mu_{n,m}^2+\sigma_{n,m}^2)}\left(\mu_{n,m} -  \frac{\sum_{t\not\in \mathcal{Z}_n''} \frac{\pi_{n,t}^o}{\mu_{n,t}^2 + \sigma_{n,t}^2} \mu_{n,t}}{\sum_{t\not\in \mathcal{Z}''_n} \frac{\pi_{n,t}^o}{\mu_{n,t}^2 + \sigma_{n,t}^2}}\right) + \frac{N^2 \pi^o_{n,m}}{2(\mu_{n,m}^2+\sigma_{n,m}^2)} \kappa_{n,m}'' \\
&= \underbrace{\pi^o_{n,m} + \frac{\frac{\pi^o_{n,m}}{\mu_{n,m}^2+\sigma_{n,m}^2}}{\sum_{t=m}^K\frac{\pi^o_{n,t}}{\mu_{n,t}^2+\sigma_{n,t}^2}} \sum_{t=0}^{m-1} \pi_{n,t}^o + \frac{N \pi^o_{n,m}\hat{\omega}_{n} }{2Q_n(\mu_{n,m}^2+\sigma_{n,m}^2)}\left(\mu_{n,m} -  \frac{\sum_{t=m}^K \frac{\pi_{n,t}^o}{\mu_{n,t}^2 + \sigma_{n,t}^2} \mu_{n,t}}{\sum_{t=m}^K \frac{\pi_{n,t}^o}{\mu_{n,t}^2 + \sigma_{n,t}^2}}\right)}_{\leq 0} \\
&\hspace{20pt}+ \frac{N(\omega_{n}'' - \hat{\omega}_n)\pi_{n,m}^o}{2Q_n(\mu_{n,m}^2+\sigma_{n,m}^2)}\underbrace{\left(\mu_{n,m}-\frac{\sum_{t=m}^K\frac{\pi_{n,t}^o}{\mu_{n,t}^2 + \sigma_{n,t}^2}\mu_{n,t}}{\sum_{t=m}^K\frac{\pi_{n,t}^o}{\mu_{n,t}^2 + \sigma_{n,t}^2}}\right)}_{\leq 0} \\
&\hspace{20pt}+ \frac{\frac{\pi_{n,m}^o}{\mu_{n,m}^2 + \sigma_{n,m}^2}}{\sum_{t=m}^K \frac{\pi_{n,t}^o}{\mu_{n,t}^2 + \sigma_{n,t}^2}}\sum_{t\in \mathcal{B}}\underbrace{\left[\pi_{n,t}^o + \frac{\frac{\pi_{n,t}^o}{\mu_{n,t}^2 + \sigma_{n,t}^2}}{\sum_{v\not\in\mathcal{Z}_n''} \frac{\pi_{n,v}^o}{\mu_{n,v}^2 + \sigma_{n,v}^2}}\sum_{v\in\mathcal{Z}_n''}\pi_{n,t}^o + \frac{N\omega_n'' \pi_{n,t}^o}{2Q_n (\mu_{n,t}^2 + \sigma_{n,t}^2)} \left(\mu_{n,t} - \frac{\sum_{v\not\in\mathcal{Z}_n''}\frac{\pi_{n,v}^o}{\mu_{n,v}^2 + \sigma_{n,v}^2}\mu_{n,v}}{\sum_{v\not\in\mathcal{Z}_n''}\frac{\pi_{n,v}^o}{\mu_{n,v}^2 + \sigma_{n,v}^2}}\right)\right]}_{\leq 0}\\
&\hspace{20pt} -\frac{\frac{\pi_{n,m}^o}{\mu_{n,m}^2 + \sigma_{n,m}^2}\sum_{t\in\mathcal{B}}\frac{\pi_{n,t}^o}{\mu_{n,t}^2 + \sigma_{n,t}^2}}{\sum_{t=m}^K \frac{\pi_{n,t}^o}{\mu_{n,t}^2 + \sigma_{n,t}^2}\sum_{t\not\in\mathcal{Z}_n''} \frac{\pi_{n,t}^o}{\mu_{n,t}^2 + \sigma_{n,t}^2}} \sum_{t\in\mathcal{B}} \pi_{n,t}^o + \frac{N^2 \pi^o_{n,m}}{2(\mu_{n,m}^2+\sigma_{n,m}^2)} \kappa_{n,m}''.
\end{align*}
As $\kappa_{n,m}''$ can only be positive if $\pi_{n,m}'' = 0$, and all other terms in the sum are at most $0$, it must be the case $\pi_{n,m}'' = 0$. 

\qed

\end{proof}

\bigskip

\begin{proof}{Proof of Theorem~\ref{thm:full_char}.}

We prove Theorem~\ref{thm:full_char} in the following way,
\begin{itemize}
\item[(1)] We show that $\pi^* \in \Pi_{\Lambda}$ iff $(\pi^*,\zeta^*,\kappa^*)$ satisfies the first order conditions and $\omega_n^* \geq 0$ for all $n \in [N]$.
\item[(2)] We show that for any $\omega_n \geq 0$ there is a unique related solution $\pi^*_n$ to the first order condition for each $n\in[N]$. This allows us to ensure $\mathcal{Z}_n$ is a well defined function of $\omega_n^*$
\item[(3)] We show that, for each $n\in[N]$, $\zeta^*$ can be written as a continuous, piecewise linear, increasing function of $\omega_n^*$ with $\omega_n^* = 0 \implies \zeta^* = 0$. This allows us to then write each $\omega_n^*$ as a continuous increasing function of $\zeta^*$, define $\mathcal{Z}_n(\zeta^*)$ as the related $\mathcal{Z}_n(\omega_n^*)$,
and thus form the the iterative characterization Theorem~\ref{thm:full_char} provides of $\pi^*$ in terms of $\zeta$.
\end{itemize}
Note that the proof of Proposition~\ref{prop:zeros} gives the form of non-zero policies given $\omega_n^*$ and the related set of zero propensity policies $\mathcal{Z}_n$ as
$$\pi_{n,t}^*=\pi^o_{n,t} + \frac{\frac{\pi^o_{n,t}}{\mu_{n,t}^2+\sigma_{n,t}^2}}{\sum_{v\not\in\mathcal{Z}_n}\frac{\pi^o_{n,v}}{\mu_{n,v}^2+\sigma_{n,v}^2}} \sum_{v\in\mathcal{Z}_n} \pi_{n,v}^o + \frac{N \pi^o_{n,t}\omega_{n}^* }{2Q_n(\mu_{n,t}^2+\sigma_{n,t}^2)}\left(\mu_{n,t} -  \frac{\sum_{v\not\in \mathcal{Z}_n} \frac{\pi_{n,v}^o}{\mu_{n,v}^2 + \sigma_{n,v}^2} \mu_{n,v}}{\sum_{v\not\in \mathcal{Z}_n} \frac{\pi_{n,v}^o}{\mu_{n,v}^2 + \sigma_{n,v}^2}}\right)$$

\medskip

Proving (1) reduces to showing that $\omega_n^* \geq 0$ for all $n\in[N]$ iff $\tau(\pi^*) \geq 0$. 

Let $t^{\max}_n = \underset{t\in \{0,\ldots,K\}}{\mathrm{argmax}} \mu_{n,t}$. If $\tau(\pi^*) \geq 0$ it implies $\pi_{n,t^{\max}_n}^* \geq \pi_{n,t^{\max}_n}^o$ for all $n \in [N]$. Otherwise we can form $\pi'$ by, for all $n$ with $\pi_{n,t^{\max}_n}^* < \pi_{n,t^{\max}_n}^o$, increasing $\pi_{n,t^{\max}_n}^*$ to $\pi_{n,t}^o$, and contracting all $\pi_{n,t}^* > \pi_{n,t}^o$ toward $\pi_{n,t}^o$ until a valid policy is formed. $\tau(\pi') \geq \tau(\pi^*)$ (as, for each unit, we increased the treatment propensity with the largest mean while decreasing others) and $s^2(\pi') < s^2(\pi^*)$ as we contracted toward $\pi^o$, the minimal value of the quadratic $s^2(\pi)$. As a result we can further contract all values of $\pi'$ toward $\pi^o$ to form a new policy $\pi''$ with $\tau(\pi'') = \tau(\pi^*)$ (as $\tau(\pi)$ is linear with $\tau(\pi^o) = 0$) and $s(\pi'') \leq s(\pi')$. This would form a contradiction as $\pi''$ invalidates $\pi^* \in \Pi_{\Lambda}$. Thus $\pi_{n,t^{\max}_n}^* \geq \pi_{n,t}^o$ for all $n \in [N]$. The form of Proposition~\ref{prop:form} then implies that for any $n \in [N]$, $\omega_n^* \geq 0$ if $\pi_{n,t^{\max}_n}^* \geq \pi_{n,t}^o$ as $\mu_{n,t^{\max}_n} - \tilde{\mu}_n > 0$ (unless $\mu_{n,t} = \tilde{\mu}$ for all $t\in\{0,\ldots,K\}$, which is a degenerate case in which for all $\pi^* \in \Pi_{\Lambda}$, $\pi^*_n = \pi^o_n$ and our characterization returns $\pi_n^o$ for any $\zeta$). 

Now we show if $\omega_n^* \geq 0$ for all $n \in [N]$ then $\tau(\pi^*) \geq 0$. Note that under Proposition~\ref{prop:form},
That $\omega_n^* = 0$ implies $\pi_{n}^* = \pi_{n}^o$. Otherwise it would be that $\pi_{n,t}^* > \pi_{n,t}^o$ for some $t$ which would contradict the first order condition at $\omega_n^* = 0$,
$$\pi_{n,t}^* = \pi_{n,t}^o\left(1 - \frac{N^2\tilde{\kappa}^*_n }{2(\mu_{n,t}^2 + \sigma_{n,t}^2)}\right),$$
as $\tilde{\kappa}^*_n \geq 0$ and $\kappa_{n,t}^* = 0$ as $\pi_{n,t}^* > 0$. Furthermore, as it will be useful in part (2), this allows us to say there exists a unique solution $\pi^*_{n} = \pi_{n}^o$ to the first order conditions when $\omega_n^* = 0$.
Moreover for $\omega_n^* > 0$,
\begin{align*}
\mu_{n,t} &> \tilde{\mu}_{n} + \frac{NQ_n}{\omega_n^*} \tilde{\kappa}_n^* & &\text{iff} & & \pi_{n,t}^* - \pi_{n,t}^o > 0.
\end{align*}
This allows us to show,
\begin{align*}
\tau(\pi^*) &= \frac{1}{N}\sum_{n=1}^N\sum_{t=0}^K \mu_{n,t} \left(\pi_{n,t}^* - \pi_{n,t}^o\right)\\
&= \frac{1}{N}\sum_{n|\omega_n^* > 0}\sum_{t=0}^K \mu_{n,t} \left(\pi_{n,t}^* - \pi_{n,t}^o\right) \\
&\geq \frac{1}{N}\sum_{n|\omega_n^* > 0}\left(\tilde{\mu}_n +\frac{NQ_n}{\omega_n^*}\tilde{\kappa}_n\right)\underbrace{\sum_{t=0}^K  \left(\pi_{n,t}^* - \pi_{n,t}^o\right)}_{=0}
\end{align*}
completing part (1).

\medskip

As we have already shown through Proposition~\ref{prop:zeros} and part (1) of this proof that the policy $\pi^*_{n}$ for any $\omega_n^* \geq 0$ is unique (unique zero sets + solution to independent linear equations) we must show that there exists a solution to the first order conditions for every $\omega_n > 0$ (as $\omega_n = 0$ is shown in part (1)) in order to prove (2). We do so independently for each $n$

We show the existence of a solution for every $\omega_{n} \geq 0$ following the iterative construction given in Theorem~\ref{thm:full_char}. That is, we first set $\kappa_n = 0$ and find solutions for all $\omega_n \in [0,\omega_n^0]$ where $\omega_n^0$ is the smallest value of $\omega_n^*$ for which no solution with $\pi_{n,t}^*  = 0 \ \forall \ t\in\{0,\ldots,K\}$ exists. We argue this policy remains at $0$, then incorporate its impact on $\kappa^*$ into the first order conditions before finding the next $\omega_n^*$ for which a policy hits $0$. By construction, all non-zero propensity treatments meet the first order conditions. Thus showing the existence of a solution for every $\omega_n \geq 0$ involves deriving the process of forming $\mathcal{Z}_n(\omega_n)$ (the set of treatments with $\pi^{*}_{n,t} = 0$ at $\omega_n^* = \omega_n$) and verifying the first order condition is met for all $t\in\mathcal{Z}_n(\omega_n)$.

Note that the form of non-zero policies from the proof of Proposition~\ref{prop:zeros} allows us to write.
\begin{align*}
\omega_n^* &= \frac{2Q_n\left(\left(\frac{\pi_{n,t}^*}{\pi_{n,t}^o}-1\right)\left(\mu_{n,t}^2 + \sigma_{n,t}^2\right) -\frac{\sum_{v \in \mathcal{Z}_n}\pi_{n,v}^o}{\sum_{v\not \in \mathcal{Z}_n} \frac{\pi_{n,v}^o}{\mu_{n,v}^2+\sigma_{n,v}^2}}\right)}{N\left(\mu_{n,t} - \frac{\sum_{v\not \in \mathcal{Z}_n} \frac{\pi_{n,v}^o}{\mu_{n,v}^2+\sigma_{n,v}^2} \mu_{n,v}}{\sum_{v\not \in \mathcal{Z}_n} \frac{\pi_{n,v}^o}{\mu_{n,v}^2+\sigma_{n,v}^2}}\right)} \\
&= \frac{2Q_n\left(\left(1-\frac{\pi_{n,t}^*}{\pi_{n,t}^o}\right)\left(\mu_{n,t}^2 + \sigma_{n,t}^2\right)\sum_{v\not \in \mathcal{Z}_n} \frac{\pi_{n,v}^o}{\mu_{n,v}^2+\sigma_{n,v}^2} +\sum_{v \in \mathcal{Z}_n}\pi_{n,v}^o\right)}{N\sum_{v\not \in \mathcal{Z}_n} \frac{\pi_{n,v}^o}{\mu_{n,v}^2+\sigma_{n,v}^2} (\mu_{n,v} - \mu_{n,t})}
\end{align*}
and thus form $\hat{\omega}_{n,t}$, the $\omega_n^*$ required for policy $t$ to become non-zero, for any (current) zero set $\mathcal{Z}_n$,
$$\hat{\omega}_{n,t} = \frac{2Q_n\left(\left(\mu_{n,t}^2 + \sigma_{n,t}^2\right)\sum_{v\not \in \mathcal{Z}_n} \frac{\pi_{n,v}^o}{\mu_{n,v}^2+\sigma_{n,v}^2} +\sum_{v \in \mathcal{Z}_n}\pi_{n,v}^o\right)}{N\sum_{v\not \in \mathcal{Z}_n} \frac{\pi_{n,v}^o}{\mu_{n,v}^2+\sigma_{n,v}^2} (\mu_{n,v} - \mu_{n,t})}.$$
We know that at $\omega_n^* = 0$, $\mathcal{Z}_{n}(\omega_n^*) = \{ \}$, which initializes us to find policies for $\omega_n^* > 0$. We assume $\mathcal{Z}_n = \{\}$, locate the minimal $\hat{\omega}_{n,t}>0$ such that a policy must be $0$ and update $\mathcal{Z}_n(\omega_n) = \{\}$ for $\omega_n \in (0,\min_{t\in\{0,\ldots,K\}}\hat{\omega}_{n,t})$ and $\mathcal{Z}_n(\min_{t\in\{0,\ldots,K\}}\hat{\omega}_{n,t}) = \underset{t\in\{0,\ldots,K\}}{\mathrm{argmin} \ \hat{\omega}_{n,t}}$. As we know the set of zero policies at $\hat{\omega}_{n,t}$ is unique, and one such policy has been found, we can restrict our search for the remaining non-zero treatments to $\omega_n^* > \hat{\omega}_{n,t}$. Thus we reform estimates for $\hat{\omega}_{n,t}$ with the updated zero set, then look to make our next update to $\mathcal{Z}_n$  above the prior value of $\omega_n^*$ for which at least one treatment hit $0$. The condition of all $\mu_{n,t} = \mu_{n,v} \ \forall \ t,v \not \in \mathcal{Z}_n$, which determines $\mathcal{Z}_n(\omega_n)$ as $\omega_n \to \infty$ is guaranteed to eventually be satisfied as there must be at least 1 positive propensity treatment, and if there is only 1 positive propensity treatment remaining the condition is trivially satisfied. This completes the process of forming $\mathcal{Z}_n(\omega_n)$ for all $\omega_n \geq 0$.

Provided there exists some $\kappa_{n,t} \geq 0$ such that $\pi_{n,t} = 0$ satisfies the first order condition for all $t \in \mathcal{Z}_n(\omega_n)$ for any $\omega_n \geq 0$, each of the solutions implied by $\mathcal{Z}_n(\omega_n)$ and $\omega_n$ exist. Let $\omega_{n,t}^0 = \min \{\omega \geq 0 | t \in \mathcal{Z}_n(\omega)\} $. From the proof of Proposition~\ref{prop:zeros} we know some $\kappa_{n,t} \geq 0$ satisfies the first order condition for $t\in \mathcal{Z}_n(\omega_n)$ with $\pi_{n,t} = 0$ if,
\begin{align*}
0 &\geq  \pi_{n,t}^o + \frac{\frac{\pi_{n,t}^o}{\mu_{n,t}^2 + \sigma_{n,t}^2}}{\sum_{v \not\in \mathcal{Z}_n(\omega_n)}\frac{\pi^o_{n,v}}{\mu_{n,v}^2+\sigma_{n,v}^2}}\sum_{v \in \mathcal{Z}_n(\omega_n)} \pi_{n,v}^o + \frac{N \pi^o_{n,t}\omega_{n} }{2Q_n(\mu_{n,t}^2+\sigma_{n,t}^2)}\left(\mu_{n,t} -  \frac{\sum_{v\not\in \mathcal{Z}_n(\omega_n)} \frac{\pi_{n,v}^o}{\mu_{n,v}^2 + \sigma_{n,v}^2} \mu_{n,v}}{\sum_{v\not\in \mathcal{Z}_n(\omega_n)} \frac{\pi_{n,v}^o}{\mu_{n,v}^2 + \sigma_{n,v}^2}}\right) 
\end{align*}
However, we proved this inequality is true when $\omega_n \geq \omega_{n,t}^0$ (as is true now) in the proof of Proposition~\ref{prop:zeros}. Thus we can be assured that a valid solution to the first order conditions exists for any $\omega_n \geq 0$. As uniqueness of $\mathcal{Z}_n(\omega_n)$ for $\omega_n \geq 0$ has already been shown, $\mathcal{Z}_n(\omega_n)$ is a well-defined set which together with $\omega_n^*$ describes a unique $\pi_n^*$ part of some $\pi^* \in \Pi_{\Lambda}$.

\medskip

Proving (3) involves solving for $\zeta^*$ in terms of $\omega_n^*$ and $\mathcal{Z}_n(\omega_n^*)$, then showing this function is piecewise linear, continuous, and increasing in $\omega_n^*$ with $\zeta^* = 0$ implied by $\omega_n^* = 0$. We know that the related $\zeta^*$ is unique from the proof of Proposition~\ref{prop:zeros} as $\kappa_{n,t}^*$ can be written as the following function of $\omega_n^*$ and $\mathcal{Z}_n(\omega_n^*)$,
\begin{align*}
\kappa_{n,t}^* &= -\frac{\omega_n^*}{NQ_n} \left[\mu_{n,t}-\tilde{\mu}_n + \frac{\sum_{v \in \mathcal{Z}_n(\omega_n^*)} \frac{\pi^o_{n,v}}{\mu_{n,v}^2+\sigma_{n,v}^2} \left(\mu_{n,v}-\tilde{\mu}_n  \right)}{\sum_{v \not\in \mathcal{Z}_n(\omega_n^*)}\frac{\pi^o_{n,v}}{\mu_{n,v}^2+\sigma_{n,v}^2}}\right] - \frac{2}{N^2} \left[\mu_{n,t}^2+\sigma_{n,t}^2 + \frac{\sum_{v \in \mathcal{Z}_n(\omega_n^*)} \pi_{n,v}^o}{\sum_{v \not\in \mathcal{Z}_n(\omega_n^*)}\frac{\pi^o_{n,v}}{\mu_{n,v}^2+\sigma_{n,v}^2}} \right]\\
&= - \frac{2(\mu_{n,t}^2 + \sigma_{n,t}^2)}{N^2\pi_{n,t}^o}\\
&\hspace{20pt}\left(\pi_{n,t}^o + \frac{\frac{\pi_{n,t}^o}{\mu_{n,t}^2 + \sigma_{n,t}^2}}{\sum_{v \not\in \mathcal{Z}_n(\omega_n^*)}\frac{\pi^o_{n,v}}{\mu_{n,v}^2+\sigma_{n,v}^2}}\sum_{v \in \mathcal{Z}_n(\omega_n^*)} \pi_{n,v}^o + \frac{N \pi^o_{n,t}\omega_{n}^* }{2Q_n(\mu_{n,t}^2+\sigma_{n,t}^2)}\left(\mu_{n,t} -  \frac{\sum_{v\not\in \mathcal{Z}_n(\omega_n^*)} \frac{\pi_{n,v}^o}{\mu_{n,v}^2 + \sigma_{n,v}^2} \mu_{n,v}}{\sum_{v\not\in \mathcal{Z}_n(\omega_n^*)} \frac{\pi_{n,v}^o}{\mu_{n,v}^2 + \sigma_{n,v}^2}}\right) \right),
\end{align*}
for all $t\in\mathcal{Z}_n(\omega_n^*)$ with $\kappa_{n,t}^* = 0$ for $t\not\in\mathcal{Z}_n(\omega_n^*)$.Thus,
\begin{align*}
\zeta^* &= \omega_n^* - N \sum_{t\in\mathcal{Z}_n(\omega_n^*)} \frac{\pi_{n,t}^o}{\mu_{n,t}^2 + \sigma_{n,t}^2} \kappa_{n,t}^* (\mu_{n,t} - \tilde{\mu}_{n})\\
\zeta^* &= \omega_n^* + \frac{2}{N}\sum_{t\in\mathcal{Z}_n(\omega_n^*)}\left(\pi_{n,t}^o + \frac{\frac{\pi_{n,t}^o}{\mu_{n,t}^2 + \sigma_{n,t}^2}}{\sum_{v \not\in \mathcal{Z}_n(\omega_n^*)}\frac{\pi^o_{n,v}}{\mu_{n,v}^2+\sigma_{n,v}^2}}\sum_{v \in \mathcal{Z}_n(\omega_n^*)} \pi_{n,v}^o + \frac{N \pi^o_{n,t}\omega_{n}^* }{2Q_n(\mu_{n,t}^2+\sigma_{n,t}^2)}\left(\mu_{n,t} -  \frac{\sum_{v\not\in \mathcal{Z}_n(\omega_n^*)} \frac{\pi_{n,v}^o}{\mu_{n,v}^2 + \sigma_{n,v}^2} \mu_{n,v}}{\sum_{v\not\in \mathcal{Z}_n(\omega_n^*)} \frac{\pi_{n,v}^o}{\mu_{n,v}^2 + \sigma_{n,v}^2}}\right) \right) \\
&\hspace{90pt}\left(\mu_{n,t} - \tilde{\mu}_n\right)\\
&= \frac{2}{N}\sum_{t\in\mathcal{Z}_n(\omega_n^*)}\pi_{n,t}^o\left(\mu_{n,t}- \frac{\sum_{v\not \in\mathcal{Z}_n(\omega_n^*) } \frac{\pi_{n,v}^o}{\mu_{n,v}^2+\sigma_{n,v}^2} \mu_{n,v}}{\sum_{v\not \in\mathcal{Z}_n(\omega_n^*) } \frac{\pi_{n,v}^o}{\mu_{n,v}^2+\sigma_{n,v}^2}}\right)\\
&\hspace{20pt}+ \Bigg(1+ Q_n^{-1}\Bigg(\sum_{t\in\mathcal{Z}_n(\omega_n^*)} \frac{\pi_{n,t}^o}{\mu_{n,t}^2+\sigma_{n,t}^2}\mu_{n,t} \left(\mu_{n,t}-\tilde{\mu}_n\right) \\
&\hspace{100pt}- \frac{\left(\sum_{t\not\in\mathcal{Z}_n(\omega_n^*)}\frac{\pi_{n,t}^o}{\mu_{n,t}^2+\sigma_{n,t}^2} \mu_{n,t}\right)\left(\sum_{t\in\mathcal{Z}_n(\omega_n^*)} \frac{\pi_{n,t}^o}{\mu_{n,t}^2+\sigma_{n,t}^2} \left(\mu_{n,t}-\tilde{\mu}_n\right)\right)}{\sum_{t\not\in\mathcal{Z}_n(\omega_n^*)} \frac{\pi_{n,t}^o}{\mu_{n,t}^2+\sigma_{n,t}^2} }\Bigg)\Bigg)\omega_{n}^* \\
&=\frac{2}{N}\sum_{t\in\mathcal{Z}_n(\omega_n^*)}\pi_{n,t}^o\left(\mu_{n,t} - \frac{\sum_{v\not \in\mathcal{Z}_n(\omega_n^*) } \frac{\pi_{n,v}^o}{\mu_{n,v}^2+\sigma_{n,v}^2} \mu_{n,v}}{\sum_{v\not \in\mathcal{Z}_n(\omega_n^*) } \frac{\pi_{n,v}^o}{\mu_{n,v}^2+\sigma_{n,v}^2}}\right) \\&\hspace{20pt}+\left(1 - \sum_{t\not\in\mathcal{Z}_n(\omega_n^*)}\frac{\pi^o_{n,t}}{\mu_{n,t}^2+\sigma_{n,t}^2} \left(\mu_{n,t} - \frac{\sum_{v\not\in\mathcal{Z}_n(\omega_n^*)}\frac{\pi^o_{n,v}}{\mu_{n,v}^2+\sigma_{n,v}^2}\mu_{n,v}}{\sum_{v\not\in\mathcal{Z}_n(\omega_n^*)}\frac{\pi^o_{n,v}}{\mu_{n,v}^2+\sigma_{n,v}^2}}\right)^2 \right) \frac{\omega_n^*}{Q_n},
\end{align*}
where,
\begin{align*}
&1 - \sum_{t\not\in\mathcal{Z}_n(\omega_n^*)}\frac{\pi^o_{n,t}}{\mu_{n,t}^2+\sigma_{n,t}^2} \left(\mu_{n,t} - \frac{\sum_{v\not\in\mathcal{Z}_n(\omega_n^*)}\frac{\pi^o_{n,v}}{\mu_{n,v}^2+\sigma_{n,v}^2}\mu_{n,v}}{\sum_{v\not\in\mathcal{Z}_n(\omega_n^*)}\frac{\pi^o_{n,v}}{\mu_{n,v}^2+\sigma_{n,v}^2}}\right)^2 \\
&=\left(1 - \sum_{v\not\in\mathcal{Z}_n(\omega_n^*)}\frac{\pi^o_{n,v}}{\mu_{n,v}^2+\sigma_{n,v}^2}\mu_{n,v}^2\right) + \frac{\left(\sum_{v\not\in\mathcal{Z}_n(\omega_n^*)}\frac{\pi^o_{n,v}}{\mu_{n,v}^2+\sigma_{n,v}^2}\mu_{n,v}\right)^2}{\sum_{v\not\in\mathcal{Z}_n(\omega_n^*)}\frac{\pi^o_{n,v}}{\mu_{n,v}^2+\sigma_{n,v}^2}}\ > 0.
\end{align*}
From this form we see that $\zeta^*$ is a piecewise linear function of $\omega_n^*$ which is locally increasing on each linear segment. Furthermore as $\mathcal{Z}_n(0) = \{\}$, $\omega_n^* = 0 \implies \zeta^* = 0$. Thus what remains to be shown is whether this function of $\omega_n^*$ is continuous, specifically at the set of knot points $\Omega_n^0 = \{\omega_n^* \geq 0|\lim_{\omega \uparrow \omega_n^*} \mathcal{Z}_n(\omega) \neq \lim_{\omega \downarrow \omega_n^*} \mathcal{Z}_n(\omega)\}$.

Let $\omega_n \in \Omega_n^0$. We will show $\zeta^*$ is continuous in $\omega_n^*$ by showing $\lim_{\omega_n^* \downarrow \omega_n}\zeta^* - \lim_{\omega_n^* \uparrow \omega_n} \zeta^* = 0$.Let $\mathcal{Z}_n^- = \lim_{\omega \uparrow \omega_n^*} \mathcal{Z}_n(\omega)$ and $\mathcal{B}_n = \lim_{\omega \downarrow \omega_n^*} \mathcal{Z}_n(\omega) \setminus \mathcal{Z}_n^-$. Note that for all $t\in\mathcal{B}_n$,
$$\omega_n = \frac{2Q_n\left(\left(\mu_{n,t}^2+\sigma_{n,t}^2\right) \sum_{v \not \in \mathcal{Z}_n^-} \frac{\pi_{n,v}^o}{\mu_{n,v}^2+\sigma_{n,v}^2} + \sum_{v\in\mathcal{Z}_n^-} \pi_{n,v}^o\right)}{N\sum_{v\not\in\mathcal{Z}_n^-} \frac{\pi_{n,v}^o}{\mu_{n,v}^2+\sigma_{n,v}^2}(\mu_{n,v} - \mu_{n,t})},$$
by construction .We have,
\begin{align*}
&\lim_{\omega_n^* \downarrow \omega_n}\zeta^* - \lim_{\omega_n^* \uparrow \omega_n} \zeta^* \\
&= \frac{2}{N} \left[\sum_{t\in\mathcal{B}_n} \pi_{n,t}^o \mu_{n,t} - \sum_{t\in\mathcal{Z}_n^- \cup \mathcal{B}_n}\pi_{n,t}^o \frac{\sum_{t\not\in\mathcal{Z}_n^-\cup \mathcal{B}_n} \frac{\pi_{n,t}^o}{\mu_{n,t}^2+\sigma_{n,t}^2}\mu_{n,t}}{\sum_{t\not\in\mathcal{Z}_n^-\cup \mathcal{B}_n} \frac{\pi_{n,t}^o}{\mu_{n,t}^2+\sigma_{n,t}^2}} +\sum_{t\in\mathcal{Z}_n^- }\pi_{n,t}^o\frac{\sum_{t\not\in\mathcal{Z}_n^-} \frac{\pi_{n,t}^o}{\mu_{n,t}^2+\sigma_{n,t}^2}\mu_{n,t}}{\sum_{t\not\in\mathcal{Z}_n^-} \frac{\pi_{n,t}^o}{\mu_{n,t}^2+\sigma_{n,t}^2}}\right] \\
&\hspace{10pt}+\frac{\omega_n}{Q_n} \left[\sum_{t\in\mathcal{B}_n} \frac{\pi^o_{n,t}}{\mu_{n,t}^2 + \sigma_{n,t}^2} (\mu_{n,t} - \tilde{\mu}_n)^2 + \frac{(\sum_{t\in\mathcal{Z}_n^- \cup\mathcal{B}_n} \frac{\pi_{n,t}^o}{\mu_{n,t}^2+\sigma_{n,t}^2} \left(\mu_{n,t}-\tilde{\mu}_n\right))^2}{\sum_{t\not\in\mathcal{Z}_n^- \cup\mathcal{B}_n} \frac{\pi_{n,t}^o}{\mu_{n,t}^2+\sigma_{n,t}^2} } - \frac{(\sum_{t\in\mathcal{Z}_n^- } \frac{\pi_{n,t}^o}{\mu_{n,t}^2+\sigma_{n,t}^2} \left(\mu_{n,t}-\tilde{\mu}_n\right))^2}{\sum_{t\not\in\mathcal{Z}_n^- } \frac{\pi_{n,t}^o}{\mu_{n,t}^2+\sigma_{n,t}^2} }\right] \\
&= \frac{2}{N}\sum_{t\in\mathcal{B}_n} \left(\pi_{n,t}^o + \frac{\frac{\pi_{n,t}^o}{\mu_{n,t}^2+\sigma_{n,t}^2}}{\sum_{v\not\in\mathcal{Z}_n^-} \frac{\pi_{n,v}^o}{\mu_{n,v}^2+\sigma_{n,v}^2} }\sum_{v\in\mathcal{Z}_n^-}\pi_{n,v}^o\right) \left(\mu_{n,t} - \frac{\sum_{v\not\in\mathcal{Z}_n^-\cup \mathcal{B}_n} \frac{\pi_{n,v}^o}{\mu_{n,v}^2+\sigma_{n,v}^2}\mu_{n,v}}{\sum_{v\not\in\mathcal{Z}_n^-\cup \mathcal{B}_n} \frac{\pi_{n,v}^o}{\mu_{n,v}^2+\sigma_{n,v}^2}} \right)  \\
&\hspace{10pt}+ \frac{\omega_n}{Q_n}\sum_{t\in\mathcal{B}_n} \frac{\frac{\pi_{n,t}^o}{\mu_{n,t}^2+\sigma_{n,t}^2}}{\sum_{v \not \in \mathcal{Z}_n^-}\frac{\pi_{n,v}^o}{\mu_{n,v}^2+\sigma_{n,v}^2}}\Bigg[(\mu_{n,t} - \tilde{\mu}_n)\sum_{v\not\in\mathcal{Z}_n^-}\frac{\pi_{n,v}^o}{\mu_{n,v}^2+\sigma_{n,v}^2}(\mu_{n,t} - \mu_{n,v})  \\&\hspace{140pt}+ \left(\mu_{n,t} - \frac{\sum_{v\not\in\mathcal{Z}_n^-\cup \mathcal{B}_n}\frac{\pi_{n,v}^o}{\mu_{n,v}^2+\sigma_{n,v}^2} \mu_{n,v}}{\sum_{v\not\in\mathcal{Z}_n^-\cup \mathcal{B}_n}\frac{\pi_{n,v}^o}{\mu_{n,v}^2+\sigma_{n,v}^2}}\right) \sum_{v \in \mathcal{Z}_n^- \cup \mathcal{B}} \frac{\pi_{n,v}^o}{\mu_{n,v}^2+\sigma_{n,v}^2}(\mu_{n,v} - \tilde{\mu}_n)\Bigg] \\
&= \frac{2}{N} \sum_{t\in\mathcal{B}_n} \left(\pi_{n,t}^o + \frac{\frac{\pi_{n,t}^o}{\mu_{n,t}^2+\sigma_{n,t}^2}}{\sum_{v \not \in \mathcal{Z}_n^-}\frac{\pi_{n,v}^o}{\mu_{n,v}^2+\sigma_{n,v}^2}} \sum_{v\in\mathcal{Z}_n^-} \pi_{n,v}^o\right) \\
&\hspace{50pt} \left(\tilde{\mu}_n - \frac{\sum_{v\not\in\mathcal{Z}_n^-\cup \mathcal{B}_n} \frac{\pi_{n,v}^o}{\mu_{n,v}^2+\sigma_{n,v}^2}\mu_{n,v}}{\sum_{v\not\in\mathcal{Z}_n^-\cup \mathcal{B}_n} \frac{\pi_{n,v}^o}{\mu_{n,v}^2+\sigma_{n,v}^2}} - \frac{\sum_{v \not\in \mathcal{Z}_n^-\cup \mathcal{B}_n}\frac{\pi_{n,v}^o}{\mu_{n,v}^2+\sigma_{n,v}^2}(\mu_{n,v} - \mu_{n,t})}{\sum_{v \not\in \mathcal{Z}_n^- }\frac{\pi_{n,v}^o}{\mu_{n,v}^2+\sigma_{n,v}^2}(\mu_{n,v} - \mu_{n,t}) } \left(\frac{\sum_{v \in \mathcal{Z}_n^- \cup \mathcal{B}} \frac{\pi_{n,v}^o}{\mu_{n,v}^2+\sigma_{n,v}^2}(\mu_{n,v} - \tilde{\mu}_n)}{\sum_{v \not\in \mathcal{Z}_n^- \cup \mathcal{B}_n}\frac{\pi_{n,v}^o}{\mu_{n,v}^2+\sigma_{n,v}^2}}\right)\right)\\
&=\frac{2}{N} \left(\frac{\sum_{v \in \mathcal{Z}_n^- \cup \mathcal{B}} \frac{\pi_{n,v}^o}{\mu_{n,v}^2+\sigma_{n,v}^2}(\mu_{n,v} - \tilde{\mu}_n)}{\sum_{v \not\in \mathcal{Z}_n^- \cup \mathcal{B}_n}\frac{\pi_{n,v}^o}{\mu_{n,v}^2+\sigma_{n,v}^2}}\right) \sum_{t\in \mathcal{B}_n} \left(\pi_{n,t}^o + \frac{\frac{\pi_{n,t}^o}{\mu_{n,t}^2+\sigma_{n,t}^2}}{\sum_{v \not \in \mathcal{Z}_n^-}\frac{\pi_{n,v}^o}{\mu_{n,v}^2+\sigma_{n,v}^2}} \sum_{v\in\mathcal{Z}_n^-} \pi_{n,v}^o\right) \left(\frac{\sum_{v \in \mathcal{B}_n}\frac{\pi_{n,v}^o}{\mu_{n,v}^2+\sigma_{n,v}^2}(\mu_{n,v} - \mu_{n,t})}{\sum_{v \not\in \mathcal{Z}_n^- }\frac{\pi_{n,v}^o}{\mu_{n,v}^2+\sigma_{n,v}^2}(\mu_{n,v} - \mu_{n,t}) }\right) \\
&=\frac{\omega_n}{Q_n} \left(\frac{\sum_{v \in \mathcal{Z}_n^- \cup \mathcal{B}} \frac{\pi_{n,v}^o}{\mu_{n,v}^2+\sigma_{n,v}^2}(\mu_{n,v} - \tilde{\mu}_n)}{\sum_{v \not\in \mathcal{Z}_n^- \cup \mathcal{B}_n}\frac{\pi_{n,v}^o}{\mu_{n,v}^2+\sigma_{n,v}^2}}\right) \underbrace{\sum_{t\in\mathcal{B}_n} \sum_{v\in\mathcal{B}_n} \frac{\pi_{n,t}^o}{\mu_{n,t}^2+\sigma_{n,t}^2}\frac{\pi_{n,v}^o}{\mu_{n,v}^2+\sigma_{n,v}^2}(\mu_{n,v} - \mu_{n,t})}_{=0}
\end{align*}
Thus $\zeta^*$ is continuous in $\omega_n^*\geq 0$ meaning it is also increasing and thus we can redefine $\mathcal{Z}_n$ in terms of $\zeta^*$ and write $\omega_n^*$ as the function of $\zeta^*$ given.
\qed
\end{proof}

\bigskip

\begin{proof}{Proof of Proposition~\ref{prop:tau_deriv}.}
We will show $\frac{d\tau(\pi^*(\zeta))}{d \zeta} > 0$ for $\zeta \in [0,\zeta_{\text{max}})$. Note that by Theorem~\ref{thm:full_char} we can write,
\begin{align*}
\frac{d\tau(\pi^*(\zeta))}{d \zeta} &= \sum_{n=1}^N \frac{\partial\tau(\pi^*(\zeta))}{\partial \omega_n^*(\zeta)} \frac{d\omega_n^*(\zeta)}{d \zeta} \\
&= \frac{1}{N}\sum_{n=1}^N \left(\frac{d\omega_n^*(\zeta)}{d \zeta}\right) \sum_{t\not \in \mathcal{Z}_n(\zeta)} \mu_{n,t} \left(\mu_{n,t} - \frac{\sum_{v\not \in \mathcal{Z}_n(\zeta)} \frac{\pi_{n,v}^o}{\mu_{n,v}^2+\sigma_{n,v}^2} \mu_{n,v}}{\sum_{v\not \in \mathcal{Z}_n(\zeta)} \frac{\pi_{n,v}^o}{\mu_{n,v}^2+\sigma_{n,v}^2}}\right) \frac{N\pi_{n,t}^o}{2Q_n(\mu_{n,t}^2 + \sigma_{n,t}^2)}\\
&= \sum_{n=1}^N \frac{1}{2Q_n}\left(\frac{d\omega_n^*(\zeta)}{d \zeta}\right) \sum_{t\not \in \mathcal{Z}_n(\zeta)} \frac{\pi_{n,t}^o}{\mu_{n,t}^2 + \sigma_{n,t}^2} \left(\mu_{n,t} - \frac{\sum_{v\not \in \mathcal{Z}_n(\zeta)} \frac{\pi_{n,v}^o}{\mu_{n,v}^2+\sigma_{n,v}^2} \mu_{n,v}}{\sum_{v\not \in \mathcal{Z}_n(\zeta)} \frac{\pi_{n,v}^o}{\mu_{n,v}^2+\sigma_{n,v}^2}}\right)^2,
\end{align*}
where $\frac{d\omega_n^*(\zeta)}{d \zeta} > 0$ and $\mu_{n,t} - \frac{\sum_{v\not \in \mathcal{Z}_n(\zeta)} \frac{\pi_{n,v}^o}{\mu_{n,v}^2+\sigma_{n,v}^2} \mu_{n,v}}{\sum_{v\not \in \mathcal{Z}_n(\zeta)} \frac{\pi_{n,v}^o}{\mu_{n,v}^2+\sigma_{n,v}^2}} \neq 0$ for some $n \in [N], t \in \{0,\ldots,K\}$ as $\zeta < \zeta_{\text{max}}$ (thus for some subject there is at least two nonzero propensity treatment with different expected outcomes by construction).

\qed
\end{proof}

\bigskip

\begin{proof}{Proof of Proposition~\ref{prop:z_deriv}.}
We will show $\frac{d\mathbb{E}[Z(\pi^*(\zeta))]}{d\zeta} = 0$ for $\zeta \in (0,\zeta_{\text{min}})$ and $\frac{d\mathbb{E}[Z(\pi^*(\zeta))]}{d\zeta} < 0$ for $\zeta \in (\zeta_{\text{min}},\zeta_{\text{max}})$. Note that,
\begin{align*}
\frac{d\mathbb{E}[Z(\pi^*(\zeta))]}{d\zeta} = \frac{d}{d\zeta} \left(\frac{\tau(\pi^*(\zeta))}{\sqrt{s^2(\pi^*(\zeta))}}\right) = \frac{1}{s(\pi^*(\zeta))} \left(\frac{d\tau(\pi^*(\zeta))}{d\zeta} - \frac{\tau(\pi^*(\zeta))}{2s^2(\pi^*(\zeta))}\frac{ds^2(\pi^*(\zeta))}{d\zeta}\right)
\end{align*}
while,
\begin{align*}
\frac{ds^2(\pi^*(\zeta))}{d\zeta} &= \sum_{n=1}^N\sum_{k=0}^T \frac{\partial s^2(\pi^*(\zeta))}{\partial \pi_{n,t}^*(\zeta)}\frac{d\pi_{n,t}^*(\zeta)}{d\omega_n^*(\zeta)}\frac{d\omega_n^*(\zeta)}{d\zeta}\\
&= \frac{1}{N}\sum_{n=1}^N \frac{1}{Q_n} \left(\frac{d\omega_n^*(\zeta)}{d\zeta}\right) \sum_{t\not \in \mathcal{Z}_n(\zeta)}\Bigg[ \left(\pi_{n,t}^*(\zeta) - \pi_{n,t}^o\right)\left(\mu_{n,t} - \frac{\sum_{v\not \in \mathcal{Z}_n(\zeta)}\frac{\pi_{n,v}^o}{\mu_{n,v}^2 + \sigma_{n,v}^2}\mu_{n,v}}{\sum_{v\not \in \mathcal{Z}_n(\zeta)} \frac{\pi_{n,v}^o}{\mu_{n,v}^2 + \sigma_{n,v}^2}}\right) \\
&\hspace{75pt} -\frac{\pi^o_{n,t}}{\mu_{n,t}^2+\sigma_{n,t}^2}\left(\mu_{n,t} - \frac{\sum_{v\not \in \mathcal{Z}_n(\zeta)}\frac{\pi_{n,v}^o}{\mu_{n,v}^2 + \sigma_{n,v}^2}\mu_{n,v}}{\sum_{v\not \in \mathcal{Z}_n(\zeta)} \frac{\pi_{n,v}^o}{\mu_{n,v}^2 + \sigma_{n,v}^2}}\right)^2 \left(\sum_{v=0}^K \mu_{n,v}(\pi^*_{n,t}(\zeta) - \pi_{n,t}^o)\right)\Bigg]
\end{align*}

Thus,
\begin{align*}
\frac{d\mathbb{E}[Z(\pi^*(\zeta))]}{d\zeta} &= \frac{1}{s(\pi^*(\zeta))}\Bigg[ \sum_{n=1}^N \frac{1}{2Q_n}\left(\frac{d\omega_n^*(\zeta)}{d \zeta}\right)\left(1 + \frac{\tau(\pi^*(\zeta))\sum_{v=0}^K \mu_{n,v}(\pi_{n,t}^*(\zeta) - \pi_{n,t}^o)}{Ns^2(\pi^*(\zeta))}\right)\\&\hspace{70pt}
\sum_{t\not \in \mathcal{Z}_n(\zeta)} \frac{\pi_{n,t}^o}{\mu_{n,t}^2 + \sigma_{n,t}^2} \left(\mu_{n,t} - \frac{\sum_{v\not \in \mathcal{Z}_n(\zeta)} \frac{\pi_{n,v}^o}{\mu_{n,v}^2+\sigma_{n,v}^2} \mu_{n,v}}{\sum_{v\not \in \mathcal{Z}_n(\zeta)} \frac{\pi_{n,v}^o}{\mu_{n,v}^2+\sigma_{n,v}^2}}\right)^2\\
&\hspace{50pt}- \sum_{n=1}^N \frac{1}{2Q_n}\left(\frac{d\omega_n^*(\zeta)}{d \zeta}\right)\left(\frac{\tau(\pi^*(\zeta))}{Ns^2(\pi^*(\zeta))}\right)\\
&\hspace{70pt} \sum_{t\not \in \mathcal{Z}_n(\zeta)} (\pi_{n,t}^*(\zeta) - \pi_{n,t}^o)\left(\mu_{n,t} - \frac{\sum_{v\not \in \mathcal{Z}_n(\zeta)} \frac{\pi_{n,v}^o}{\mu_{n,v}^2+\sigma_{n,v}^2} \mu_{n,v}}{\sum_{v\not \in \mathcal{Z}_n(\zeta)} \frac{\pi_{n,v}^o}{\mu_{n,v}^2+\sigma_{n,v}^2}}\right) \Bigg]\\
&= \frac{1}{s(\pi^*(\zeta))}\Bigg[\sum_{n=1}^N \frac{1}{2Q_n}\left(\frac{d\omega_n^*(\zeta)}{d \zeta}\right)\Bigg(1 + \frac{\tau(\pi^*(\zeta))}{Ns^2(\pi^*(\zeta))}\Bigg[\sum_{v\in\mathcal{Z}_n(\zeta)} \pi_{n,v}^o\left(\frac{\sum_{j\not \in \mathcal{Z}_n(\zeta)} \frac{\pi_{n,j}^o}{\mu_{n,j}^2+\sigma_{n,j}^2}\mu_{n,j}}{\sum_{j\not \in \mathcal{Z}_n(\zeta)} \frac{\pi_{n,j}^o}{\mu_{n,j}^2+\sigma_{n,j}^2}}-\mu_{n,v}\right)\\& \hspace{100pt} + \frac{N\omega_n^*(\zeta)}{2Q_n} \left(\sum_{v\not \in \mathcal{Z}_n(\zeta)} \frac{\pi_{n,v}^o}{\mu_{n,v}^2+\sigma_{n,v}^2}\left(\mu_{n,v}-\frac{\sum_{j\not \in \mathcal{Z}_n(\zeta)} \frac{\pi_{n,j}^o}{\mu_{n,j}^2+\sigma_{n,j}^2}\mu_{n,j}}{\sum_{j\not \in \mathcal{Z}_n(\zeta)} \frac{\pi_{n,j}^o}{\mu_{n,j}^2+\sigma_{n,j}^2}}\right)^2 -1 \right)\Bigg]\Bigg)\\
&\hspace{70pt}\sum_{t\not \in \mathcal{Z}_n(\zeta)} \frac{\pi_{n,t}^o}{\mu_{n,t}^2 + \sigma_{n,t}^2} \left(\mu_{n,t} - \frac{\sum_{v\not \in \mathcal{Z}_n(\zeta)} \frac{\pi_{n,v}^o}{\mu_{n,v}^2+\sigma_{n,v}^2} \mu_{n,v}}{\sum_{v\not \in \mathcal{Z}_n(\zeta)} \frac{\pi_{n,v}^o}{\mu_{n,v}^2+\sigma_{n,v}^2}} \right)^2\Bigg]\\
&= \frac{1}{s(\pi^*(\zeta))}\left[\sum_{n=1}^N \frac{1}{2Q_n}\left(\frac{d\omega_n^*(\zeta)}{d \zeta}\right)\Bigg(1 - \frac{\zeta\tau(\pi^*(\zeta))}{2s^2(\pi^*(\zeta))} \Bigg)\sum_{t\not \in \mathcal{Z}_n(\zeta)} \frac{\pi_{n,t}^o}{\mu_{n,t}^2 + \sigma_{n,t}^2} \left(\mu_{n,t} - \frac{\sum_{v\not \in \mathcal{Z}_n(\zeta)} \frac{\pi_{n,v}^o}{\mu_{n,v}^2+\sigma_{n,v}^2} \mu_{n,v}}{\sum_{v\not \in \mathcal{Z}_n(\zeta)} \frac{\pi_{n,v}^o}{\mu_{n,v}^2+\sigma_{n,v}^2}} \right)^2\right].
\end{align*}
We will complete the proof by showing $\left(1 - \frac{\zeta\tau(\pi^*(\zeta))}{2s^2(\pi^*(\zeta))} \right) = 0$ for $\zeta \in (0,\zeta_{\text{min}})$ and $\left(1 - \frac{\zeta\tau(\pi^*(\zeta))}{2s^2(\pi^*(\zeta))} \right) < 0$ for $\zeta \in (\zeta_{\text{min}},\zeta_{\text{max}})$. Note that the dual variable interpretation of $\zeta$ allows us to say,
\begin{align*}
\frac{ds^2(\pi^*(\zeta))}{d\tau(\pi^*(\zeta))} &= \zeta \\
\int_{x=0}^{x=\zeta} ds^2(\pi^*(x))&=\int_{x=0}^{x=\zeta} xd\tau(\pi^*(x)) \\
s^2(\pi^*(\zeta)) &= \zeta\tau(\pi^*(\zeta)) - \int_{0}^\zeta \tau(\pi^*(x))dx.
\end{align*}
If $\zeta \in (0,\zeta_{\text{min}})$ then $\tau(\pi^*(x))$ is linear in $x$ (as $\tau$ is linear in $\pi^*$ and $\pi^*$ is linear in $x$ below $\zeta_{\text{min}}$), so $\int_{0}^{\zeta}\tau(\pi^*(x))dx = \frac{\zeta}{2}\tau(\pi^*(\zeta))$. Thus for $\zeta\in (0,\zeta_{\text{min}})$,
\begin{align*}
\frac{s^2(\pi^*(\zeta))}{\tau(\pi^*(\zeta))} &= \frac{\zeta}{2} \\
\implies \left(1- \frac{\zeta \tau(\pi^*(\zeta))}{2s^2(\pi^*(\zeta))}\right) &= 0.
\end{align*}
On the other hand if $\zeta \in (\zeta_{\text{min}},\zeta_{\text{max}})$, then $\tau(\pi^*(x))$ is a concave piecewise-linear function of $x$. This is because, from the proof of Proposition~\ref{prop:tau_deriv} (evaluating $\frac{d\omega_n^*(\zeta)}{d\zeta}$), we have
\begin{align*}
\frac{d\tau(\pi^*(\zeta))}{d\zeta} & = \frac{1}{2} \sum_{n=1}^N \frac{ \sum_{t\not \in \mathcal{Z}_n(\zeta)} \frac{\pi_{n,t}^o}{\mu_{n,t}^2 + \sigma_{n,t}^2} \left(\mu_{n,t} - \frac{\sum_{v\not \in \mathcal{Z}_n(\zeta)} \frac{\pi_{n,v}^o}{\mu_{n,v}^2+\sigma_{n,v}^2} \mu_{n,v}}{\sum_{v\not \in \mathcal{Z}_n(\zeta)} \frac{\pi_{n,v}^o}{\mu_{n,v}^2+\sigma_{n,v}^2}}\right)^2}{1-\sum_{t\not \in \mathcal{Z}_n(\zeta)} \frac{\pi_{n,t}^o}{\mu_{n,t}^2 + \sigma_{n,t}^2} \left(\mu_{n,t} - \frac{\sum_{v\not \in \mathcal{Z}_n(\zeta)} \frac{\pi_{n,v}^o}{\mu_{n,v}^2+\sigma_{n,v}^2} \mu_{n,v}}{\sum_{v\not \in \mathcal{Z}_n(\zeta)} \frac{\pi_{n,v}^o}{\mu_{n,v}^2+\sigma_{n,v}^2}}\right)^2},
\end{align*}
which is increasing in $\sum_{t\not \in \mathcal{Z}_n(\zeta)} \frac{\pi_{n,t}^o}{\mu_{n,t}^2 + \sigma_{n,t}^2} \left(\mu_{n,t} - \frac{\sum_{v\not \in \mathcal{Z}_n(\zeta)} \frac{\pi_{n,v}^o}{\mu_{n,v}^2+\sigma_{n,v}^2} \mu_{n,v}}{\sum_{v\not \in \mathcal{Z}_n(\zeta)} \frac{\pi_{n,v}^o}{\mu_{n,v}^2+\sigma_{n,v}^2}}\right)^2$ for each $n$. However if $\zeta'$ is a knot in the piecewise linear evolution of $\pi_n^*(\zeta)$ such that  $\mathcal{Z}_n^-= \lim_{\zeta \uparrow \zeta'} \mathcal{Z}_n(\zeta)$ and $\mathcal{B}_n = \lim_{\zeta \downarrow \zeta'} \mathcal{Z}_n(\zeta)\setminus \mathcal{Z}_n^-$ we have,
\begin{align*}
&\sum_{t\not \in \mathcal{Z}_n^- \cup \mathcal{B}_n} \frac{\pi_{n,t}^o}{\mu_{n,t}^2 + \sigma_{n,t}^2} \left(\mu_{n,t} - \frac{\sum_{v\not \in \mathcal{Z}_n^-\cup\mathcal{B}_n} \frac{\pi_{n,v}^o}{\mu_{n,v}^2+\sigma_{n,v}^2} \mu_{n,v}}{\sum_{v\not \in \mathcal{Z}_n^- \cup\mathcal{B}_n} \frac{\pi_{n,v}^o}{\mu_{n,v}^2+\sigma_{n,v}^2}}\right)^2 - \sum_{t\not \in \mathcal{Z}_n^-} \frac{\pi_{n,t}^o}{\mu_{n,t}^2 + \sigma_{n,t}^2} \left(\mu_{n,t} - \frac{\sum_{v\not \in \mathcal{Z}_n^-} \frac{\pi_{n,v}^o}{\mu_{n,v}^2+\sigma_{n,v}^2} \mu_{n,v}}{\sum_{v\not \in \mathcal{Z}_n^-} \frac{\pi_{n,v}^o}{\mu_{n,v}^2+\sigma_{n,v}^2}}\right)^2 \\
&= -  \sum_{t \in \mathcal{B}_n} \frac{\pi_{n,t}^o}{\mu_{n,t}^2 + \sigma_{n,t}^2} \left(\mu_{n,t} - \frac{\sum_{v\not \in \mathcal{Z}_n^-} \frac{\pi_{n,v}^o}{\mu_{n,v}^2+\sigma_{n,v}^2} \mu_{n,v}}{\sum_{v\not \in \mathcal{Z}_n^- } \frac{\pi_{n,v}^o}{\mu_{n,v}^2+\sigma_{n,v}^2}}\right)^2\\
&\hspace{20pt} + \sum_{t\not \in \mathcal{Z}_n^-\cup\mathcal{B}_n}\frac{\pi_{n,t}^o}{\mu_{n,t}^2+\sigma_{n,t}^2} \left[\left(\mu_{n,t} - \frac{\sum_{v\not \in \mathcal{Z}_n^-\cup\mathcal{B}_n} \frac{\pi_{n,v}^o}{\mu_{n,v}^2+\sigma_{n,v}^2} \mu_{n,v}}{\sum_{v\not \in \mathcal{Z}_n^- \cup\mathcal{B}_n} \frac{\pi_{n,v}^o}{\mu_{n,v}^2+\sigma_{n,v}^2}}\right)^2 - \left(\mu_{n,t} - \frac{\sum_{v\not \in \mathcal{Z}_n^-} \frac{\pi_{n,v}^o}{\mu_{n,v}^2+\sigma_{n,v}^2} \mu_{n,v}}{\sum_{v\not \in \mathcal{Z}_n^-} \frac{\pi_{n,v}^o}{\mu_{n,v}^2+\sigma_{n,v}^2}}\right)^2 \right] \\
&= -  \sum_{t \in \mathcal{B}_n} \frac{\pi_{n,t}^o}{\mu_{n,t}^2 + \sigma_{n,t}^2} \left(\mu_{n,t} - \frac{\sum_{v\not \in \mathcal{Z}_n^-} \frac{\pi_{n,v}^o}{\mu_{n,v}^2+\sigma_{n,v}^2} \mu_{n,v}}{\sum_{v\not \in \mathcal{Z}_n^- } \frac{\pi_{n,v}^o}{\mu_{n,v}^2+\sigma_{n,v}^2}}\right)^2\\
&\hspace{20pt} +\sum_{t\not \in \mathcal{Z}_n^-\cup\mathcal{B}_n}\frac{\pi_{n,t}^o}{\mu_{n,t}^2+\sigma_{n,t}^2}\left(\frac{\sum_{v\not \in \mathcal{Z}_n^-\cup\mathcal{B}_n} \frac{\pi_{n,v}^o}{\mu_{n,v}^2+\sigma_{n,v}^2} \mu_{n,v}}{\sum_{v\not \in \mathcal{Z}_n^- \cup\mathcal{B}_n} \frac{\pi_{n,v}^o}{\mu_{n,v}^2+\sigma_{n,v}^2}}-\frac{\sum_{v\not \in \mathcal{Z}_n^-} \frac{\pi_{n,v}^o}{\mu_{n,v}^2+\sigma_{n,v}^2} \mu_{n,v}}{\sum_{v\not \in \mathcal{Z}_n^-} \frac{\pi_{n,v}^o}{\mu_{n,v}^2+\sigma_{n,v}^2}}\right)\left(\frac{\sum_{v\not \in \mathcal{Z}_n^-} \frac{\pi_{n,v}^o}{\mu_{n,v}^2+\sigma_{n,v}^2} \mu_{n,v}}{\sum_{v\not \in \mathcal{Z}_n^-} \frac{\pi_{n,v}^o}{\mu_{n,v}^2+\sigma_{n,v}^2} }-\mu_{n,t}\right)\\
&= -  \sum_{t \in \mathcal{B}_n} \frac{\pi_{n,t}^o}{\mu_{n,t}^2 + \sigma_{n,t}^2} \left(\mu_{n,t} - \frac{\sum_{v\not \in \mathcal{Z}_n^-} \frac{\pi_{n,v}^o}{\mu_{n,v}^2+\sigma_{n,v}^2} \mu_{n,v}}{\sum_{v\not \in \mathcal{Z}_n^- } \frac{\pi_{n,v}^o}{\mu_{n,v}^2+\sigma_{n,v}^2}}\right)^2\\
&\hspace{20pt}- \left(\frac{\sum_{v\not \in \mathcal{Z}_n^-\cup\mathcal{B}_n} \frac{\pi_{n,v}^o}{\mu_{n,v}^2+\sigma_{n,v}^2} \mu_{n,v}}{\sum_{v\not \in \mathcal{Z}_n^- \cup\mathcal{B}_n} \frac{\pi_{n,v}^o}{\mu_{n,v}^2+\sigma_{n,v}^2}}-\frac{\sum_{v\not \in \mathcal{Z}_n^-} \frac{\pi_{n,v}^o}{\mu_{n,v}^2+\sigma_{n,v}^2} \mu_{n,v}}{\sum_{v\not \in \mathcal{Z}_n^-} \frac{\pi_{n,v}^o}{\mu_{n,v}^2+\sigma_{n,v}^2}}\right)^2 \sum_{t\not \in \mathcal{Z}_n^- \cup \mathcal{B}_n} \frac{\pi_{n,t}^o}{\mu_{n,t}^2 + \sigma_{n,t}^2}.
\end{align*}        
Thus $\sum_{t\not \in \mathcal{Z}_n(\zeta)} \frac{\pi_{n,t}^o}{\mu_{n,t}^2 + \sigma_{n,t}^2} \left(\mu_{n,t} - \frac{\sum_{v\not \in \mathcal{Z}_n(\zeta)} \frac{\pi_{n,v}^o}{\mu_{n,v}^2+\sigma_{n,v}^2} \mu_{n,v}}{\sum_{v\not \in \mathcal{Z}_n(\zeta)} \frac{\pi_{n,v}^o}{\mu_{n,v}^2+\sigma_{n,v}^2}}\right)^2$ is decreasing in $\zeta$ at each knot point. 

Therefore for $\zeta \in (\zeta_{\text{min}},\zeta_{\text{max}})$ we have (by Jensen's inequality) that $\int_{0}^{\zeta}\tau(\pi^*(x))dx \geq \zeta\tau\left(\pi^*\left(\frac{\zeta}{2}\right)\right) > \frac{\zeta}{2}\tau(\pi^*(\zeta))$, so
\begin{align*}
\frac{s^2(\pi^*(\zeta))}{\tau(\pi^*(\zeta))} &< \frac{\zeta}{2} \\
\implies \left(1- \frac{\zeta \tau(\pi^*(\zeta))}{2s^2(\pi^*(\zeta))}\right) &< 0.
\end{align*}

\qed

\end{proof}

\bigskip

\begin{proof}{Proof of Theorem~\ref{thm:ParetoEquivalance}.}
From Proposition~\ref{prop:relax} and Theorem~\ref{thm:full_char} we know,
$$\Pi_{\text{Opt}}\subseteq \Pi_{\Lambda} = \left\{\pi^*(\zeta)|\zeta \in [0,\zeta_{\text{max}}]\right\},$$
where for $\zeta > \zeta_{\text{max}}$, $\pi^*(\zeta) = \pi^*(\zeta_{\text{max}})$ is redundant. Further, from Proposition~\ref{prop:tau_deriv} and ~\ref{prop:z_deriv} we know $\pi*(\zeta) \not \in \Pi_{\text{Opt}}$ for $\zeta \in [0,\zeta_{\text{min}})$ as $\tau(\pi^*(\zeta)) < \tau(\pi^*(\zeta_{\text{min}}))$ while $\mathbb{E}[Z(\pi^*(\zeta))] = \mathbb{E}[Z(\pi^*(\zeta_{\text{min}}))]$ (aside from $\zeta = 0$ which is excluded for returning $\pi^o$ and thus generating an undefined Z-score). Therefore,
$$\Pi_{\text{Opt}}\subseteq \left\{\pi^*(\zeta)|\zeta \in [\zeta_{\text{min}},\zeta_{\text{max}}]\right\},$$
and what remains is to show $\pi^*(\zeta) \in \Pi_{\text{Opt}}$ for $\zeta \in [\zeta_{\text{min}},\zeta_{\text{max}}].$

Let $\zeta \in [\zeta_{\text{min}},\zeta_{\text{max}}]$, and let $\pi \in \left(\Delta^K\right)^N$ such that $\tau(\pi) \geq \tau(\pi^*(\zeta))$ and $\frac{\tau(\pi)}{s(\pi)} \geq \frac{\tau(\pi^*(\zeta))}{s(\pi^*(\zeta))}$. Suppose to the contrary that $(\tau(\pi),s(\pi)) \neq (\tau(\pi^*(\zeta)), s(\pi^*(\zeta)))$. It must then be that $\tau(\pi)>\tau(\pi^*(\zeta))$ as $\tau(\pi)=\tau(\pi^*(\zeta))$ implies $s(\pi) < s(\pi^*(\zeta))$ which contradicts the construction of $\pi^*$ (minimal $s^2$ for a fixed $\tau(\pi)$). Let $\zeta' \in (\zeta,\zeta_{\text{max}}]$ be such that $\tau(\pi) = \tau(\pi^*(\zeta'))$ (which must exist by the intermediate value theorem). Thus, by definition of $\pi^*$, $s(\pi^*(\zeta')) \leq s(\pi)$. Therefore,
$$\frac{\tau(\pi^*(\zeta))}{s(\pi^*(\zeta))}\leq\frac{\tau(\pi)}{s(\pi)}\leq \frac{\tau(\pi^*(\zeta'))}{s(\pi^*(\zeta'))}.$$
However this violates Proposition~\ref{prop:z_deriv}, that the expected $z$-score of $\pi^*(\zeta)$ is decreasing in $(\zeta_{\text{min}},\zeta_{\text{max}}).$ This implies that $(\tau(\pi),s(\pi)) = (\tau(\pi^*(\zeta)), s(\pi^*(\zeta)))$, meaning,
$$\left\{\pi'\ \in \left(\Delta^K\right)^N \middle| \tau(\pi') \geq\tau(\pi),\frac{\tau(\pi')}{s(\pi')} \geq \frac{\tau(\pi)}{s(\pi)}, (\tau(\pi'),s(\pi')) \neq (\tau(\pi),s(\pi))\right\} = \varnothing,$$
so $\pi^*(\zeta) \in \Pi_{\mathrm{Opt}}$.

\qed

\end{proof}

\begin{proof}{Proof of Proposition~\ref{prop:order_0}.}
Note that using Theorem~\ref{thm:full_char} we can write,
\begin{align*}
\pi_{n,t}^*(\zeta)- \pi_{n,t}^o &= \frac{\pi_{n,t}^o}{\mu_{n,t}^2 +\sigma_{n,t}^2} \left[ \frac{\sum_{v\in\mathcal{Z}_n(\zeta)} \pi_{n,v}^o}{\sum_{v\not \in \mathcal{Z}_n(\zeta)} \frac{\pi_{n,v}^o}{\mu_{n,v}^2 +\sigma_{n,v}^2}} +  \frac{N\omega_n^*(\zeta)}{2Q_n}\left(\mu_{n,t} - \frac{\sum_{v\not \in \mathcal{Z}_n(\zeta)} \frac{\pi_{n,v}^o}{\mu_{n,v}^2 +\sigma_{n,v}^2}\mu_{n,v}}{\sum_{v\not \in \mathcal{Z}_n(\zeta)} \frac{\pi_{n,v}^o}{\mu_{n,v}^2 +\sigma_{n,v}^2}}\right)\right] \\
&= \frac{\pi_{n,t}^o}{\mu_{n,t}^2 +\sigma_{n,t}^2}\left[\frac{N\omega_n^*(\zeta)}{2Q_n}\underbrace{\left(\mu_{n,t}-\mu_{n,t'}\right)}_{>0} + \frac{\mu_{n,t'}^2 +\sigma_{n,t'}^2}{\pi_{n,t'}^o}\underbrace{\left(\pi_{n,t'}^*(\zeta) - \pi_{n,t'}^o\right)}_{>0} \right]\\
& >0
\end{align*}
\qed
\end{proof}

\begin{proof}{Proof of Proposition~\ref{prop:order_breaks}.}
From Theorem~\ref{thm:full_char} we know that $\frac{\pi_{n,t'}^*(\zeta)}{\pi_{n,t'}^o} > \frac{\pi_{n,t}^*(\zeta)}{\pi_{n,t}^o}$ implies,
\begin{align*}
&\left(\frac{1}{\mu_{n,t'}^2+\sigma_{n,t'}^2}\right)\left[ \frac{\sum_{v\in\mathcal{Z}_n(\zeta)} \pi_{n,v}^o}{\sum_{v\not \in \mathcal{Z}_n(\zeta)} \frac{\pi_{n,v}^o}{\mu_{n,v}^2 +\sigma_{n,v}^2}} +  \frac{N\omega_n^*(\zeta)}{2Q_n}\left(\mu_{n,t'} - \frac{\sum_{v\not \in \mathcal{Z}_n(\zeta)} \frac{\pi_{n,v}^o}{\mu_{n,v}^2 +\sigma_{n,v}^2}\mu_{n,v}}{\sum_{v\not \in \mathcal{Z}_n(\zeta)} \frac{\pi_{n,v}^o}{\mu_{n,v}^2 +\sigma_{n,v}^2}}\right)\right]\\
&> \left(\frac{1}{\mu_{n,t}^2+\sigma_{n,t}^2}\right)\left[ \frac{\sum_{v\in\mathcal{Z}_n(\zeta)} \pi_{n,v}^o}{\sum_{v\not \in \mathcal{Z}_n(\zeta)} \frac{\pi_{n,v}^o}{\mu_{n,v}^2 +\sigma_{n,v}^2}} +  \frac{N\omega_n^*(\zeta)}{2Q_n}\left(\mu_{n,t} - \frac{\sum_{v\not \in \mathcal{Z}_n(\zeta)} \frac{\pi_{n,v}^o}{\mu_{n,v}^2 +\sigma_{n,v}^2}\mu_{n,v}}{\sum_{v\not \in \mathcal{Z}_n(\zeta)} \frac{\pi_{n,v}^o}{\mu_{n,v}^2 +\sigma_{n,v}^2}}\right)\right]\\
\implies& \left(\frac{\sum_{v\in\mathcal{Z}_n(\zeta)} \pi_{n,v}^o}{\sum_{v\not \in \mathcal{Z}_n(\zeta)} \frac{\pi_{n,v}^o}{\mu_{n,v}^2 +\sigma_{n,v}^2}} + \frac{N\omega_n^*(\zeta)}{2Q_n}\left(\mu_{n,t'}-\frac{\sum_{v\not \in \mathcal{Z}_n(\zeta)} \frac{\pi_{n,v}^o}{\mu_{n,v}^2 +\sigma_{n,v}^2}\mu_{n,v}}{\sum_{v\not \in \mathcal{Z}_n(\zeta)} \frac{\pi_{n,v}^o}{\mu_{n,v}^2 +\sigma_{n,v}^2}} \right)\right)\left(\frac{1}{\mu_{n,t'}^2+\sigma_{n,t'}^2}-\frac{1}{\mu_{n,t}^2+\sigma_{n,t}^2}\right)\\
&> \frac{N\omega_n^*(\zeta)}{2Q_n} \left(\mu_{n,t}- \mu_{n,t'}\right)\left(\frac{1}{\mu_{n,t}^2+\sigma_{n,t}^2}\right) \\
\implies& \left(\frac{\pi_{n,t'}^*(\zeta)}{\pi_{n,t'}^o} -1\right)\left(\mu_{n,t}^2+\sigma_{n,t}^2 - \mu_{n,t'}^2-\sigma_{n,t'}^2\right) > \frac{N\omega_n^*(\zeta)}{2Q_n} \underbrace{\left(\mu_{n,t}- \mu_{n,t'}\right)}_{>0}.
\end{align*}
As the right hand side is positive for $\zeta > 0$ it must either be the case that,
\begin{itemize}
\item[(i)] $\frac{\pi_{n,t'}^*(\zeta)}{\pi_{n,t'}^o} >1$ and $\mu_{n,t}^2+\sigma_{n,t}^2 > \mu_{n,t'}^2+\sigma_{n,t'}^2$, or
\item[(ii)]$\frac{\pi_{n,t'}^*(\zeta)}{\pi_{n,t'}^o} <1$ and $\mu_{n,t}^2+\sigma_{n,t}^2 < \mu_{n,t'}^2+\sigma_{n,t'}^2$.
\end{itemize}
\qed
\end{proof}

\begin{proof}{Proof of Proposition~\ref{prop:bestZ}.}
Note that we have
$$\max_{\pi \in \left(\Delta^K\right)^N} \mathbb{E}\left[Z(\pi)\right] = \lim_{\zeta \downarrow 0} \mathbb{E}\left[Z(\pi^*(\zeta))\right]$$
by Proposition~\ref{prop:z_deriv}. Thus what remains is evaluating this limit. Note that near $\zeta = 0$, $\mathcal{Z}_n(\zeta) = \{\}$ for all $n$. Thus, plugging in the characterization of $\pi^*(\zeta)$ from Theorem~\ref{thm:full_char} we have
\begin{align*}
\lim_{\zeta \downarrow 0} \mathbb{E}\left[Z(\pi^*(\zeta))\right] & = \lim_{\zeta \downarrow 0} \frac{\tau(\pi^*(\zeta))}{s(\pi^*(\zeta))} \\
&= \lim_{\zeta\downarrow 0} \frac{\frac{1}{N}\sum_{n=1}^N  \frac{N\omega_n^*(\zeta)}{2Q_n}\sum_{t = 0}^K \frac{\pi_{n,t}^o}{\mu_{n,t}^2+\sigma_{n,t}^2} \left(\mu_{n,t} - \tilde{\mu}_n\right)^2}{\frac{1}{N}\sqrt{\sum_{n=1}^N\left(\frac{N\omega_n^*(\zeta)}{2Q_n}\right)^2\left[\sum_{t=0}^K \frac{\pi_{n,t}^o}{\mu_{n,t}^2+\sigma_{n,t}^2}\left(\mu_{n,t} - \tilde{\mu}_n\right)^2 - \left(\sum_{v = 0}^K \frac{\pi_{n,v}^o}{\mu_{n,v}^2+\sigma_{n,v}^2} \left(\mu_{n,v} - \tilde{\mu}_n\right)^2\right)^2\right]}} \\
&= \lim_{\zeta\downarrow 0} \frac{\zeta\sum_{n=1}^N  \frac{\sum_{t = 0}^K \frac{\pi_{n,t}^o}{\mu_{n,t}^2+\sigma_{n,t}^2} \left(\mu_{n,t} - \tilde{\mu}_n\right)^2}{1-\sum_{t = 0}^K \frac{\pi_{n,t}^o}{\mu_{n,t}^2+\sigma_{n,t}^2} \left(\mu_{n,t} - \tilde{\mu}_n\right)^2}}{\zeta\sqrt{\sum_{n=1}^N\frac{\sum_{t = 0}^K \frac{\pi_{n,t}^o}{\mu_{n,t}^2+\sigma_{n,t}^2} \left(\mu_{n,t} - \tilde{\mu}_n\right)^2\left(1-\sum_{t = 0}^K \frac{\pi_{n,t}^o}{\mu_{n,t}^2+\sigma_{n,t}^2} \left(\mu_{n,t} - \tilde{\mu}_n\right)^2\right)}{\left(1-\sum_{t = 0}^K \frac{\pi_{n,t}^o}{\mu_{n,t}^2+\sigma_{n,t}^2} \left(\mu_{n,t} - \tilde{\mu}_n\right)^2\right)^2}}} \\
&= \sqrt{\sum_{n=1}^N  \frac{\sum_{t = 0}^K \frac{\pi_{n,t}^o}{\mu_{n,t}^2+\sigma_{n,t}^2} \left(\mu_{n,t} - \tilde{\mu}_n\right)^2}{1-\sum_{t = 0}^K \frac{\pi_{n,t}^o}{\mu_{n,t}^2+\sigma_{n,t}^2} \left(\mu_{n,t} - \tilde{\mu}_n\right)^2}}.
\end{align*}
By rewriting $1 = \sum_{t=0}^K \pi_{n,t}^o$ we can further simplify this expression as,
\begin{align*}
\lim_{\zeta \downarrow 0} \mathbb{E}\left[Z(\pi^*(\zeta))\right] &= \sqrt{\sum_{n=1}^N  \frac{\sum_{t = 0}^K \frac{\pi_{n,t}^o}{\mu_{n,t}^2+\sigma_{n,t}^2} \left(\mu_{n,t} - \tilde{\mu}_n\right)^2}{\sum_{t = 0}^K \frac{\pi_{n,t}^o}{\mu_{n,t}^2+\sigma_{n,t}^2} \left( \mu_{n,t}^2 + \sigma_{n,t}^2 - \left(\mu_{n,t} - \tilde{\mu}_n\right)^2\right)}} \\
&= \sqrt{\sum_{n=1}^N  \frac{\sum_{t = 0}^K \frac{\pi_{n,t}^o}{\mu_{n,t}^2+\sigma_{n,t}^2} \mu_{n,t}^2 - \tilde{\mu}_n^2 \sum_{t=0}^K\frac{\pi_{n,t}^o}{\mu_{n,t}^2+\sigma_{n,t}^2}}{\sum_{t = 0}^K \frac{\pi_{n,t}^o}{\mu_{n,t}^2+\sigma_{n,t}^2} \sigma_{n,t}^2  + \tilde{\mu}_n^2\sum_{t=0}^K\frac{\pi_{n,t}^o}{\mu_{n,t}^2+\sigma_{n,t}^2}}}.
\end{align*}
By normalizing the numerator and denominator by $\sum_{t = 0}^K \frac{\pi_{n,t}^o}{\mu_{n,t}^2+\sigma_{n,t}^2}$ we get a form equivalent to the limit given in the proposition,
$$\lim_{\zeta \downarrow 0} \mathbb{E}\left[Z(\pi^*(\zeta))\right] = \sqrt{\sum_{n=1}^N \frac{\mathbb{E}\left[\mu_{n,T_n^\dagger}^2\right] - \mathbb{E}\left[\mu_{n,T_n^\dagger}\right]^2}{\mathbb{E}\left[\sigma_{n,T_n^\dagger}^2\right] + \mathbb{E}\left[\mu_{n,T_n^\dagger}\right]^2}}.$$
\qed
\end{proof}

\begin{proof}{Proof of Proposition~\ref{prop:unbiased_est}}
Using the definition of $\hat{\tau}$ we can write,
\begin{align*}
\mathbb{E}\left[\hat{\tau}(\pi(\mathcal{T},X^o))|\mathcal{T},X^o\right] &= \frac{1}{N}\sum_{n=1}^N \mathbb{E}\left[Y_n^o \left(\frac{\pi_{n,T_n^o}(\mathcal{T},X^o)}{\pi^o_{n,T_n^o}}-1\right)\middle|\mathcal{T},X^o\right].
\end{align*}
We can then use the law of total probability to separate the expectation based on the value of $T_n^o$ and simplify using the assumptions of the Rubin Causal Model,
\begin{align*}
\mathbb{E}\left[\hat{\tau}(\pi(\mathcal{T},X^o))\middle|\mathcal{T},X^o\right] &= \frac{1}{N}\sum_{n=1}^N \mathbb{E}\left[\mathbb{E}\left[Y_n^o \left(\frac{\pi_{n,T_n^o}(\mathcal{T},X^o)}{\pi^o_{n,T_n^o}}-1\right)\middle| T_n^o\right]\middle|\mathcal{T},X^o\right]\\
&= \frac{1}{N}\sum_{n=1}^N \sum_{t=0}^K \pi_{n,t}^o\mathbb{E}\left[Y_n^o \left(\frac{\pi_{n,T_n^o}(\mathcal{T},X^o)}{\pi^o_{n,T_n^o}}-1\right)\middle| T_n^o = t,\mathcal{T},X^o\right] \\
&= \frac{1}{N}\sum_{n=1}^N \sum_{t=0}^K \pi_{n,t}^o\mathbb{E}\left[Y_{n,t} \left(\frac{\pi_{n,t}(\mathcal{T},X^o)}{\pi^o_{n,t}}-1\right)\middle|\mathcal{T},X^o\right]\\
&= \frac{1}{N}\sum_{n=1}^N \sum_{t=0}^K \mu_{n,t}\left(\pi_{n,t}(\mathcal{T},X^o))-\pi^o_{n,t}\right)\\
&= \tau(\pi(\mathcal{T},X^o)).
\end{align*}
\qed
\end{proof}

\section{Stylized Simulations}
\label{sec:stylized}

We design a simulation environment in which each unit has a hidden type which determines whether a treatment has a positive, negative, or negligible effect. These types are connected to the unit's covariates via a randomly generated classification problem. We train a set of honest random forest regressors (one for each treatments) to provide counterfactual outcome estimates. We study how the estimated Pareto frontier varies with predictive model error as controlled by (1) the size of the training dataset, and (2) the complexity of the causal relationship between covariates and counterfactual outcomes as well as compare to alternate approaches to policy optimization. 

Upon generation, each unit $n$ is assigned a type $\theta_n$ from one of 5 possible types which determines their counterfactual outcome distributions for each of 25 possible treatments. If unit $n$ is assigned treatment $t$, their outcome is drawn from a normal distribution whose mean and variance are determined by their type $Y_{n,t}\sim \mathcal{N}(\mu_{\theta_n,t},\sigma^2_{\theta_n,t})$. We generate the counterfactual mean and variance of the outcome in response to each treatment for all types $\theta$ and treatments $t$ as,
\begin{align*}
\mu_{\theta,t} &\overset{iid}{\sim} \begin{cases}
\mathcal{N}(1,10^{-4}) & \text{w.p. }\ 0.1 \\
\mathcal{N}(-1,10^{-4}) & \text{w.p. }\ 0.1 \\
\mathcal{N}(0,10^{-4}) & \text{w.p. }\ 0.8
\end{cases} & \sigma_{\theta,t}^2 = 1 + 3\mu_{\theta,t}^2.
\end{align*}
Each treatment can be thought of as good, bad, or neutral for each type, with small perturbations added to ensure all treatments do not impose the same conditional average treatment effect for any type. We generate units in covariate--type pairs using a random classification problem generated by scikit learn's \texttt{make\_classification} method. We generate units with 10 covariates which could be informative of type and 10 noninformative covariates. 

To generate observational data for both training and testing, we generate treatments uniformly at randomly, assigning them a counterfactual outcome from the distribution of their given type--treatment. This allows us to generate an observational dataset $\mathcal{O} = \{(\pi^o,X^o,T^o,Y^o)\}$ of 2500 samples against which we run out of sample performance tests. It also allows us to create training datasets $\mathcal{T}$ of various sizes to explore how our approach depends on the quality of its model. As input to estimate the Pareto frontier of policies, we train one honest random forest for each treatment observed in the training dataset $\mathcal{T}$ \citep{athey_generalized_2019} as implemented by EconML's \texttt{RegressionForest} method. We then extract counterfactual variance predictions according to the bag of bootstraps method detailed in \cite{Lu_varPred}.

\begin{figure}
\centering
\begin{subfigure}[b]{0.49\textwidth}
\centering
\includegraphics[width=\textwidth]{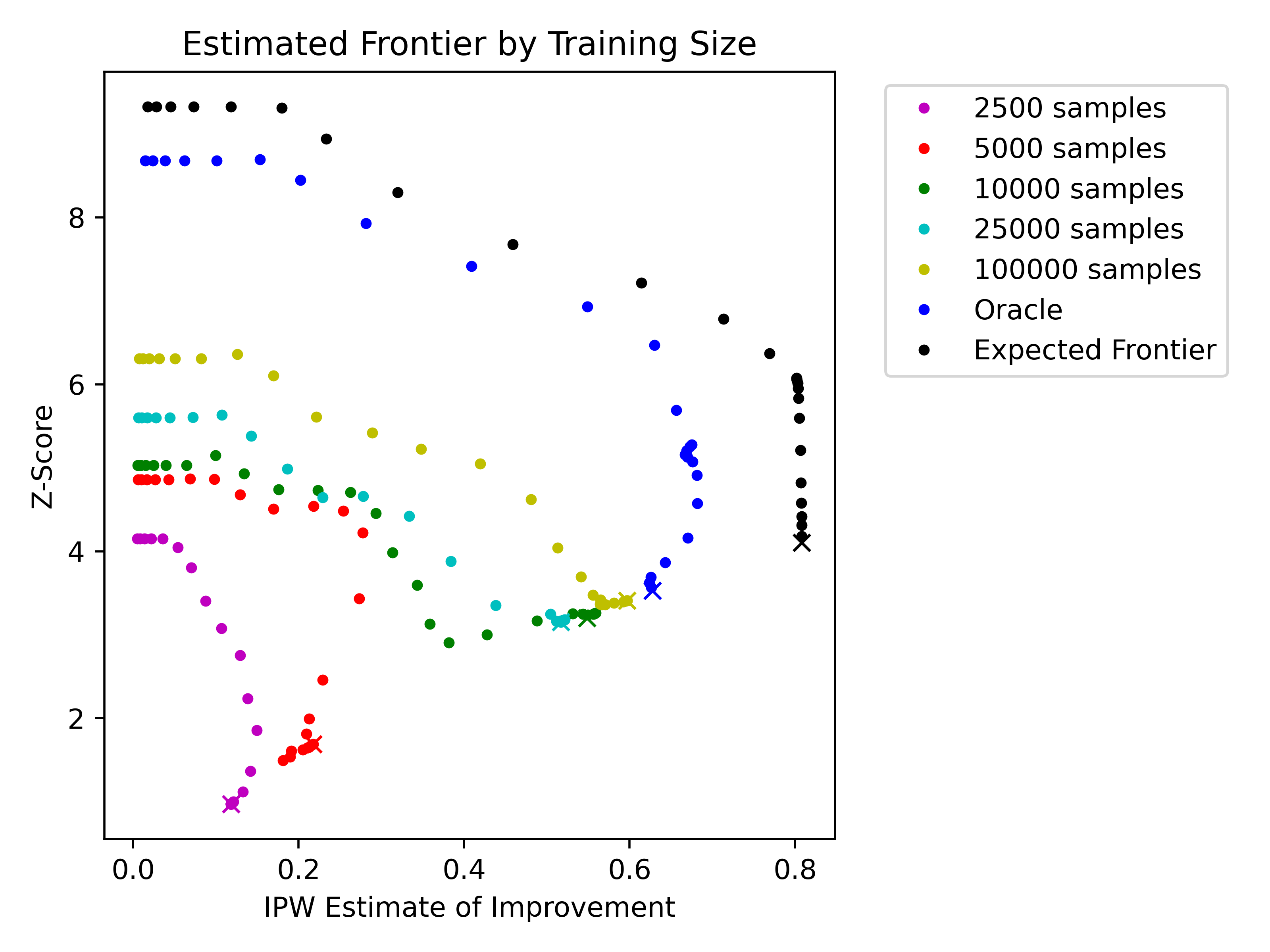}
\label{fig:frontier_samps}
\end{subfigure}
\hfill
\begin{subfigure}[b]{0.49\textwidth}
\centering
\includegraphics[width=\textwidth]{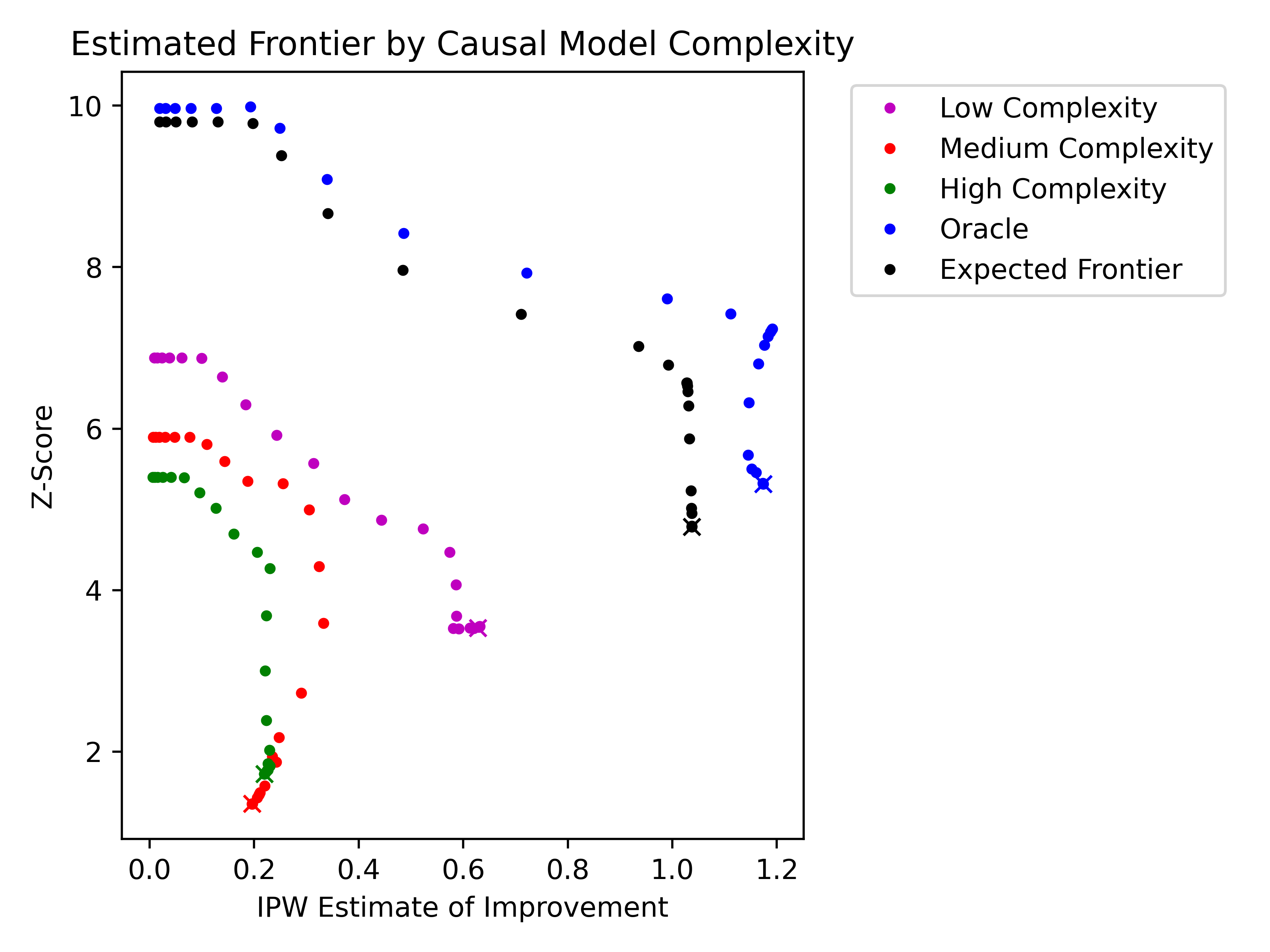}
\label{fig:frontier_complex}
\end{subfigure}
\caption{Points along the Pareto frontier in our stylized simulations, while varying training set size (left) and complexity of the true model (right). The $\times$'s mark the corresponding estimated expectation-maximizing policy.}
\label{fig:frontier}
\end{figure}

Figure~\ref{fig:frontier} shows the estimated Pareto frontier as we vary the error of the prediction model. The left holds the complexity of the causal model fixed by generating 3 clusters of covariates per type and explores how additional training samples improve the quality of the estimated frontier. The right fixes the number of training samples at 50,000 and varies the complexity of the causal model by generating either 1 (Low), 3 (Medium), or 9 (High) covariate clusters per type. In addition, we report the performance of the true frontier (i.e., where our predictive model is generated by an oracle) as well as its expected performance ex-ante generation of the test dataset.

As expected, performance along the Pareto frontier increases with training set size and decreases with the complexity of the true model; both lead to lower predictive accuracy. Prediction model error impacts performance along the frontier heterogeneously. The variance introduced to the IPW estimator by the realization of the test dataset's treatments and outcomes can curve the estimated Pareto Frontier so that increasing $\zeta$ can either increase the $z$-score or decrease the estimated improvement of the policy. We observe this most often for relatively high $\zeta$, where evaluation is the most variable. Furthermore, higher error prediction models generate frontiers that rapidly lose significance as the dual variable $\zeta$ increases. The less accurate the prediction model, the lower $\zeta$ for which $\hat{\pi}^*(\zeta)$ may begin to eliminate good treatments for some units, leading to this deterioration. 

Figure~\ref{fig:comp} compares the performance of a policy on the estimated Pareto frontier against policies generated by other common policy optimization methods. We compare IAPO to (1) the estimated expectation maximizing policy according to the honest random forests (the same predictive model we use with IAPO) \citep{athey_generalized_2019}, (2) a forest of policy trees designed for direct policy optimization \citep{athey_policy_2021} using EconML's \texttt{DRPolicyForest} method, and (3) a regularized weight-based approach to policy optimization called Norm-POEM \citep{NIPS2015_39027dfa} implemented in PyTorch.  IAPO selects the policy $\zeta = 0.128$, chosen following Section~\ref{sec:select} with $\lambda = 0.4$ and $Z_{\text{min}} = 2.5$. We chose $\lambda$ by consulting Figure~\ref{fig:frontier} and observing that policies built off of predictive models with higher error struggled to reach this improvement. Norm-POEM requires two hyper-parameters: $M$ (a variance at which to clip importance weights) and $\gamma$ (the regularization penalty). We run Norm-POEM for all combinations of $M\in\{1,\infty\}$ (with no weight inflation or no clipping) and 10 values of $\gamma$ spread logarithmically from 1 to 100 (the range in which the policy is responsive to $\gamma$ given these inputs), then report the most significant and best expected result over all runs. We use smaller sample sizes in these simulations (10,000 training samples in the right figure) to focus on the setting where extracting a policy that is statistically significantly better than the observational policy is challenging.

\begin{figure}
\centering
\begin{subfigure}[b]{0.49\textwidth}
\centering
\includegraphics[width=\textwidth]{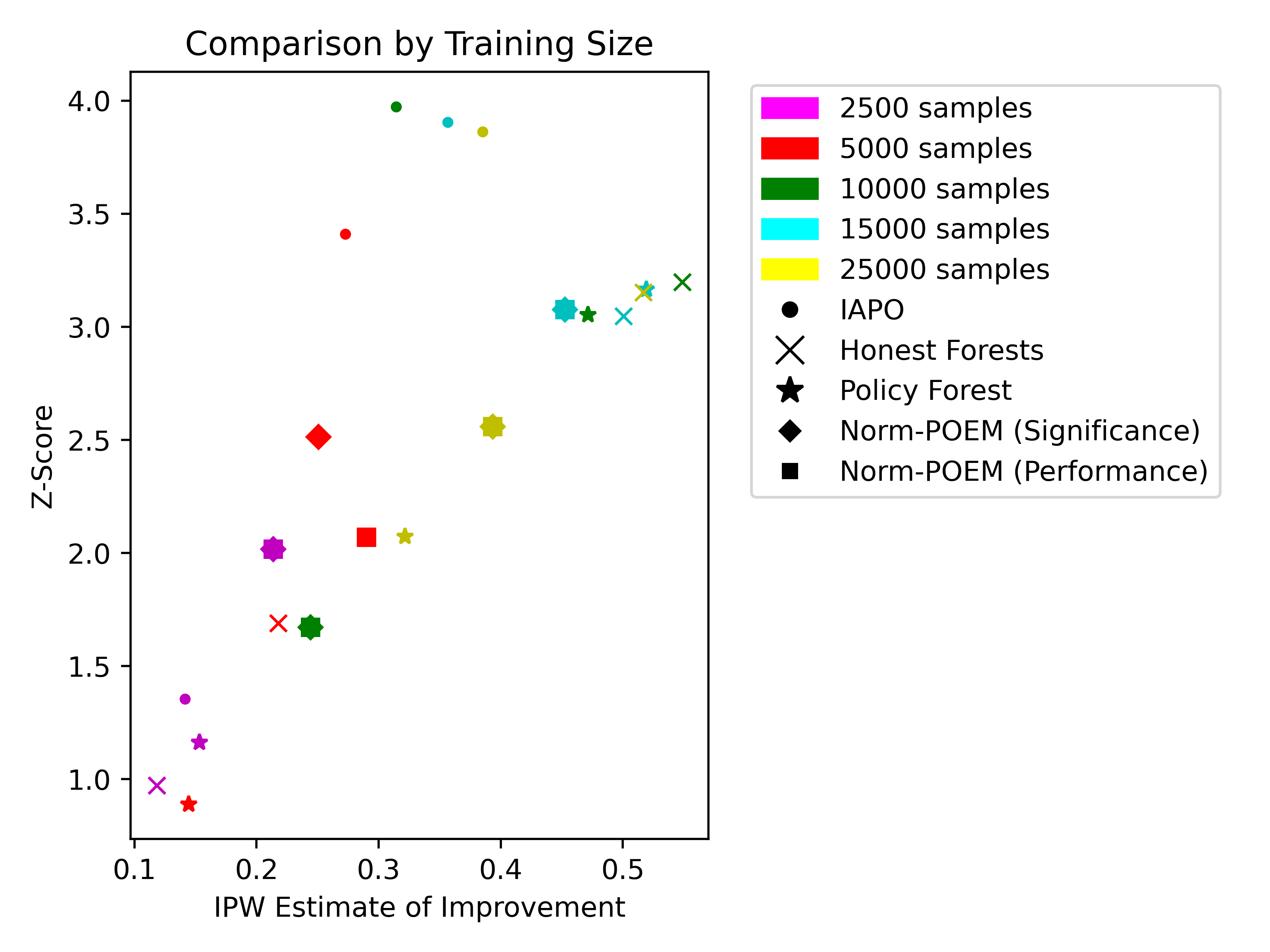}
\label{fig:comp_samps}
\end{subfigure}
\hfill
\begin{subfigure}[b]{0.49\textwidth}
\centering
\includegraphics[width=\textwidth]{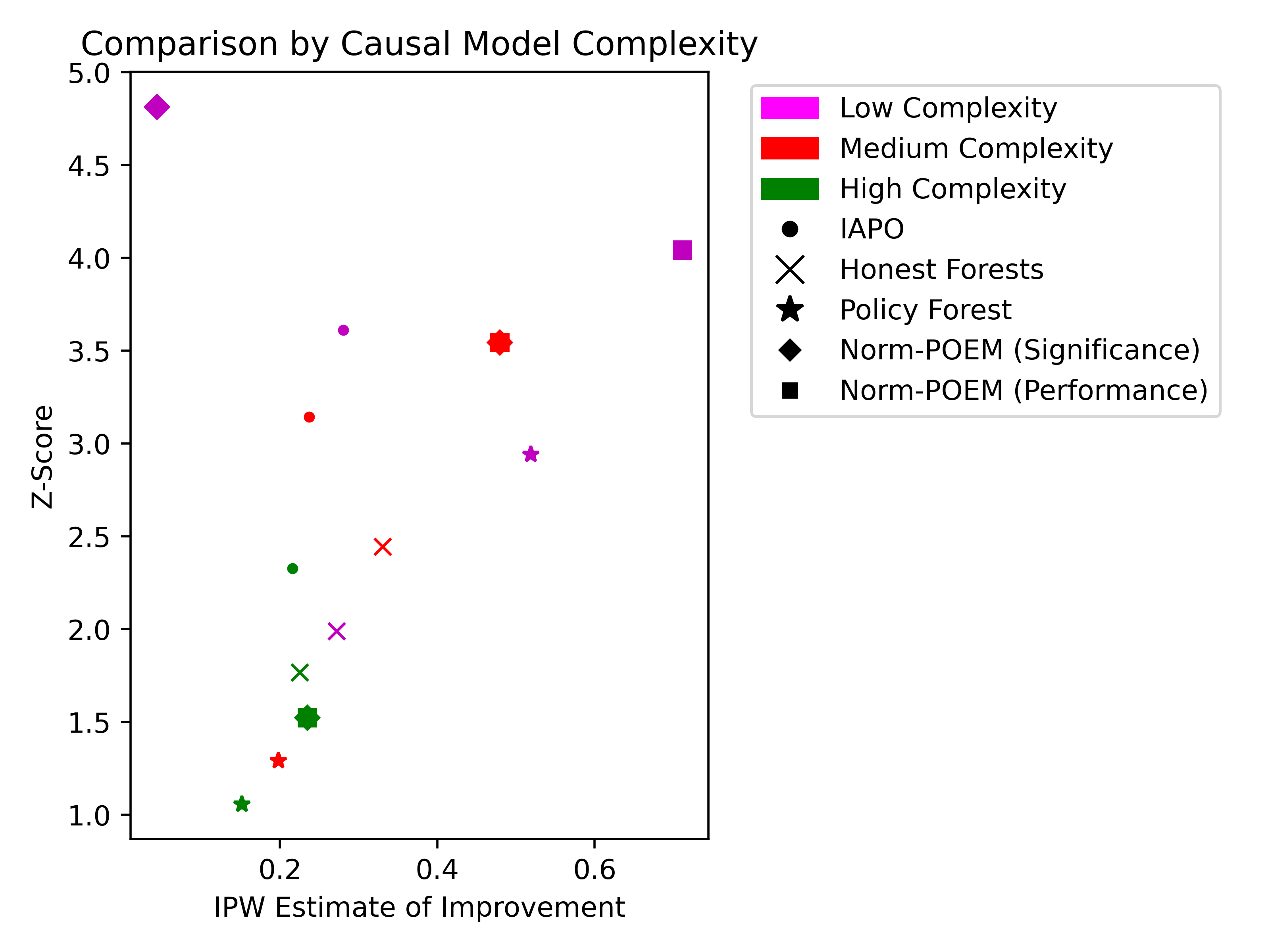}
\label{fig:comp_comp}
\end{subfigure}
\caption{Comparison of IAPO against policies generated by other policy optimization methods, while varying training set size (left) and complexity of the true model (right).}
\label{fig:comp}
\end{figure}

Given sufficient training samples and a simple enough causal model, most policy optimization methods can extract significant policy improvements. However, IAPO can do so consistently with fewer training samples and higher model complexities. An exception to this is Norm-POEM, although its performance is sporadic, failing a significance test at a 95\% confidence level when given 10,000 samples. Furthermore, from Figure~\ref{fig:frontier}, we know significant policy improvements can be obtained using IAPO with only 2500 training samples and a smaller value of $\zeta$. 

The right side of Figure~\ref{fig:comp} shows that the complexity of the causal model impacts the performance of Policy Forest and Norm-POEM substantially more than approaches based on honest random forests (such as IAPO). This finding may be because Policy Forest and Norm-POEM directly optimize policies over some policy class on the training dataset, which delivers favorable regret bounds \citep{beygelzimer_offset_2008}, but relies  on a near-optimal policy existing within the policy class. Complex policy classes are difficult to compute, which is why Policy Forest and Norm-Poem use decision trees and exponential models, respectively. The optimal policies for complex causal models may not be easy to identify for these policy classes, leading to sub-optimal performance.

\end{APPENDIX}

\end{document}